\documentclass[11pt]{article}

\usepackage[preprint]{acl}
\usepackage{times}
\usepackage{latexsym}
\usepackage{booktabs}
\usepackage{multirow}
\usepackage{siunitx}
\usepackage{tabularx}
\usepackage{CJKutf8}
\usepackage{placeins}
\usepackage{amssymb}
\usepackage{makecell}
\usepackage{hyperref}

\usepackage[most]{tcolorbox}
\tcbuselibrary{skins, breakable}
\usepackage{fvextra}

\newcommand{\ptag}[1]{%
  \tcbox[on line,
    colback=black,colframe=black,coltext=white,
    boxsep=0pt,left=2pt,right=2pt,top=1pt,bottom=1pt
  ]{\scriptsize\bfseries #1}%
}

\DefineVerbatimEnvironment{Code}{Verbatim}{
  fontsize=\scriptsize,
  breaklines=true,
  breakanywhere=false,     % 关键：避免单词内部断行
  breaksymbolleft={},
  breaksymbolright={},
  breaksymbol={}
}

\newtcolorbox{ListingBoxTaggedFloat}[2]{%
  enhanced,
  colback=white, colframe=black, boxrule=0.8pt, arc=2pt,
  left=5pt,right=5pt,top=4pt,bottom=4pt, boxsep=0pt,
  colbacktitle=black, coltitle=white,
  fonttitle=\bfseries\footnotesize,
  title={\ptag{#1}\hspace{0.6em}#2},
  titlerule=0pt             % 关键：去掉标题下那条横线
}

\newtcolorbox{ListingBoxTaggedBreak}[2]{%
  enhanced, breakable,
  colback=white, colframe=black, boxrule=0.8pt, arc=2pt,
  left=5pt,right=5pt,top=4pt,bottom=4pt, boxsep=0pt,
  colbacktitle=black, coltitle=white,
  fonttitle=\bfseries\footnotesize,
  title={\ptag{#1}\hspace{0.6em}#2},
  titlerule=0pt
}

\usepackage[T1]{fontenc}
\usepackage[utf8]{inputenc}

\usepackage{microtype}

\usepackage{inconsolata}

\usepackage{graphicx}
\usepackage{enumitem}

\title{Memory-Driven Role-Playing: Evaluation and Enhancement of Persona Knowledge Utilization in LLMs}

\author{
  \textbf{Kai Wang$^{\dag}$}, 
  \textbf{Haoyang You$^{\dag}$},
  \textbf{Yang Zhang$^{\ddag}$},
  \textbf{Zhongjie Wang$^{\dag}$\thanks{Corresponding author.}},
\\
  $^{\dag}$Harbin Institute of Technology, China\\
  $^{\ddag}$Macquarie University, Australia
\\
 $^{\dag}$kai\_wang@hit.edu.cn, 
yebai467@gmail.com, rainy@hit.edu.cn\\
 $^{\ddag}$yang.zhang@mq.edu.au\\
}

\begin{document}
\maketitle
\begin{abstract}
A core challenge for faithful LLM role-playing is sustaining consistent characterization throughout long, open-ended dialogues, as models frequently fail to recall and accurately apply their designated persona knowledge without explicit cues.
To tackle this, we propose the Memory-Driven Role-Playing paradigm. Inspired by Stanislavski’s ``emotional memory'' acting theory, this paradigm frames persona knowledge as the LLM’s internal memory store, requiring retrieval and application based solely on dialogue context, thereby providing a rigorous test of depth and autonomous use of knowledge. 
Centered on this paradigm, we contribute: 
(1) MREval, a fine-grained evaluation framework assessing four memory-driven abilities—Anchoring, Recalling, Bounding, and Enacting; 
(2) MRPrompt, a prompting architecture that guides structured memory retrieval and response generation; 
and (3) MRBench, a bilingual (Chinese/English) benchmark for fine-grained diagnosis. 
The novel paradigm provides a comprehensive diagnostic for four-staged role-playing abilities across 12 LLMs. Crucially, experiments show that MRPrompt allows small models (e.g., Qwen3-8B) to match the performance of much larger closed-source LLMs (e.g., Qwen3-Max and GLM-4.7), and confirms that upstream memory gains directly enhance downstream response quality, validating the staged theoretical foundation.
\end{abstract}

\section{Introduction}
\label{sec:introduction}

The role-playing capabilities of large language models (LLMs) are attracting significant interest, enabling applications that range from interactive game characters and personalized virtual companions to simulated assistants with defined personas~\citep{ran-etal-2025-bookworld,chen2025catch,qi2025kokorochat,tu-etal-2023-characterchat}. 
In these scenarios, success hinges on the LLM's ability to remain strictly \emph{in character}. This entails generating responses that are coherent, human-like, and faithful to the designated persona, without reverting to generic patterns or unrelated characters~\citep{wang-etal-2024-rolellm,yu-etal-2025-rpgbench,zhou2025characterbench,ruangtanusak2025talk}. 
Thus, the key is the sustained and faithful application of predefined persona knowledge within dynamic, open-ended dialogues~\citep{elboudouri-etal-2025-rpeval,ji-etal-2025-enhancing}.

\begin{figure}[t]
  \centering
  \includegraphics[width=1.0\linewidth]{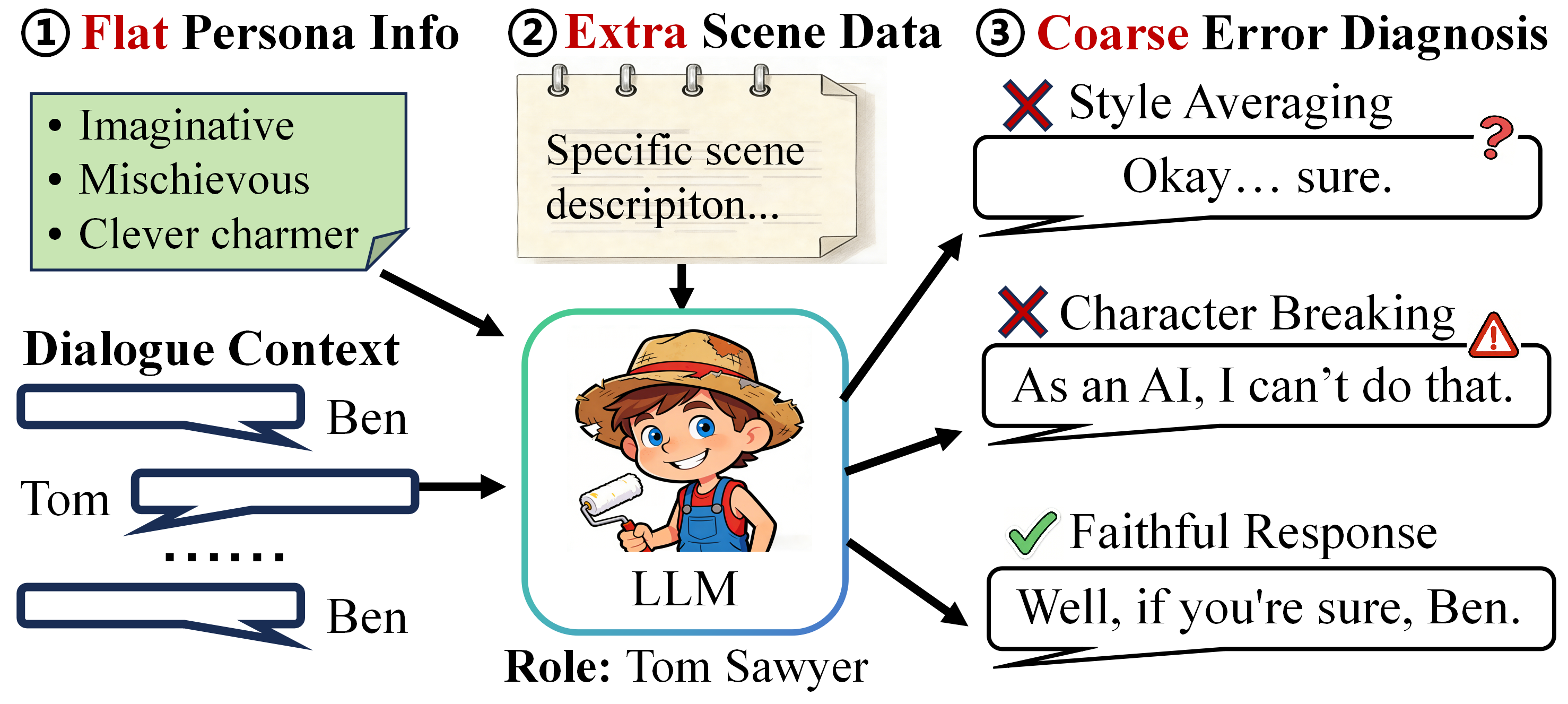}
  \vspace{-7mm}
  \caption{Three Issues in LLM Role-Playing Paradigm}
  \label{fig:motivation}
\vspace{-5mm}
\end{figure}

However, as illustrated in Figure~\ref{fig:motivation}, existing role-playing paradigms encounter three major issues in utilizing persona knowledge effectively:
(1) \textbf{Flat Persona Info}: Personas are often representationally flat, listing traits without contextual expression. This lack of guidance leads LLMs to average across persona facets into generic replies and to drift locally out of character~\citep{zhang-etal-2018-personalizing,li2023chatharuhi,shin-etal-2025-spotting,yu-etal-2025-beyond}.
(2) \textbf{Extra Scene Data}: Response generation relies on explicit, extra dialogue-scene descriptions, which simplifies reasoning but inflates success; models fail to generalize to real interactions lacking such cues~\citep{he-etal-2025-crab,zhang-etal-2025-roleplot}, creating misleading benchmarks. 
(3) \textbf{Coarse Error Diagnosis}: Holistic scoring aggregates performance into a single metric, obscuring failure modes and hindering attribution of issues (e.g., persona misalignment vs. context misunderstanding)~\citep{wang-etal-2025-characterbox,ran-etal-2025-bookworld,tang-etal-2025-rolebreak}.

To address these issues, we draw inspiration from a foundational performance theory: \emph{Stanislavski's system of emotional memory}~\citep{stanislavski1989actor},
which holds that authentic embodiment emerges when an actor recalls experiential memories through sensory details, rather than performing emotions.
Analogously, we argue that faithful LLM role-playing requires a contextual memory recall process. 
We operationalize this as a new \textbf{Memory-Driven Role-Playing (MDRP)} task,
which requires the model to (i) treat persona knowledge as a long-term memory store, and (ii) retrieve from it using only dialogue context (short-term memory).
MDRP thus serves as a targeted probe: it tests whether personas are encoded with sufficient depth for specific recall, and whether that recall can occur autonomously, without extra scene prompts.

To both evaluate and enhance LLMs under the MDRP task, we make a series of interconnected technical contributions:

First, we introduce a fine-grained evaluation framework \textbf{MREval} that decomposes faithful role-playing into four measurable memory-driven abilities: (a) Anchoring: accurate retention of persona knowledge; (b) Recalling: retrieving relevant facets given dialogue cues; (c) Bounding: adhering to the knowledge's constraints; and (d) Enacting: generating natural responses faithful to the recalled knowledge. 
By quantifying each ability with two metrics, it pinpoints breakdowns to specific stages of memory access and application, exposing weaknesses that traditional holistic metrics miss.

Second, we introduce the \textbf{MRPrompt} prompting architecture, implementing the contextual memory recall process required by MDRP.
It consists of: (i) Narrative Schema, which structures persona knowledge into a hierarchical and queryable format (e.g., global summary, core traits, and situational facets); and (ii) Magic-If Protocol, which, inspired by Stanislavski's acting technique, guides the LLM to perform targeted retrieval from this schema and to generate situatively coherent responses.

Finally, we construct \textbf{MRBench}, a bilingual MDRP benchmark derived from 10 English and 6 Chinese novels. 
It enables fine-grained diagnosis of the four memory abilities through systematic control over persona memory and dialogue context.
For scalable yet reliable evaluation, we implement an LLM-as-Judge procedure, whose scores are calibrated to human ratings via an annotation study.

We conduct a comprehensive evaluation of 12 representative LLMs on the MDRP task using MRBench. This benchmark not only offers a standardized view of memory-driven role-playing but, through its stage-wise design, also diagnoses failures by localizing them to specific memory stages. 
Beyond diagnostics, a crucial finding is that MRPrompt \textbf{empowers smaller open models to compete with larger closed-source ones}. 
For instance, Qwen3-8B augmented with MRPrompt attains an Avg. Score of 8.12, on par with the much larger GLM-4.7 Base (8.11) and surpassing Qwen3-Max Base (8.08), demonstrating that performance gains can be achieved without scaling the model backbone. 
This result, supported by granular analyses showing a consistent pipeline effect from memory to enactment, validates the staged nature of memory-driven role-playing.
\section{Related Work}
\label{sec:related_work}

\paragraph{Role-Playing Tasks and Benchmarks.}
Most role-playing setups condition generation on character profiles from canonical sources
and evaluate whether outputs remain \textit{in character}.
Early systems ground role-play in extracted dialogues or curated descriptions~\citep{li2023chatharuhi,shao2023characterllm}, while recent benchmarks broaden role pools and protocols or measure text-based persona consistency~\citep{wang-etal-2024-rolellm,liu2024roleagent,tu-etal-2024-charactereval,zhou2025characterbench}.
However, benchmarks often report aggregate response-level scores, offering limited attribution of failures.
In contrast, the proposed MDRP frames role-playing as cue-driven persona memory use under dialogue context, instantiated by MRBench for controlled comparison and paired with a stage-aware evaluation protocol for diagnosis.

\paragraph{Evaluation Protocols and Diagnostic Metrics.}
Role-play evaluation has moved from coarse judgments to more structured protocols for persona fidelity/consistency \citep{li2023chatharuhi,tu-etal-2024-charactereval}, but still largely relies on overall response-level ratings, sometimes with stress tests or judge-centric analyses \citep{zhou2025personaeval,tang-etal-2025-rolebreak,zhao-etal-2025-beware}.
Finer-grained probing methods (e.g., InCharacter) and segment-level OOC detection reveal that holistic scores can mask localized failures \citep{wang-etal-2024-incharacter,shin-etal-2025-spotting}, yet they do not attribute errors to distinct stages of persona-memory utilization.
We therefore propose MREval, a stage-aware protocol that decomposes MDRP into four sub-abilities with per-stage diagnostic metrics.

\begin{figure*}[t]
  \centering
  \includegraphics[width=\textwidth]{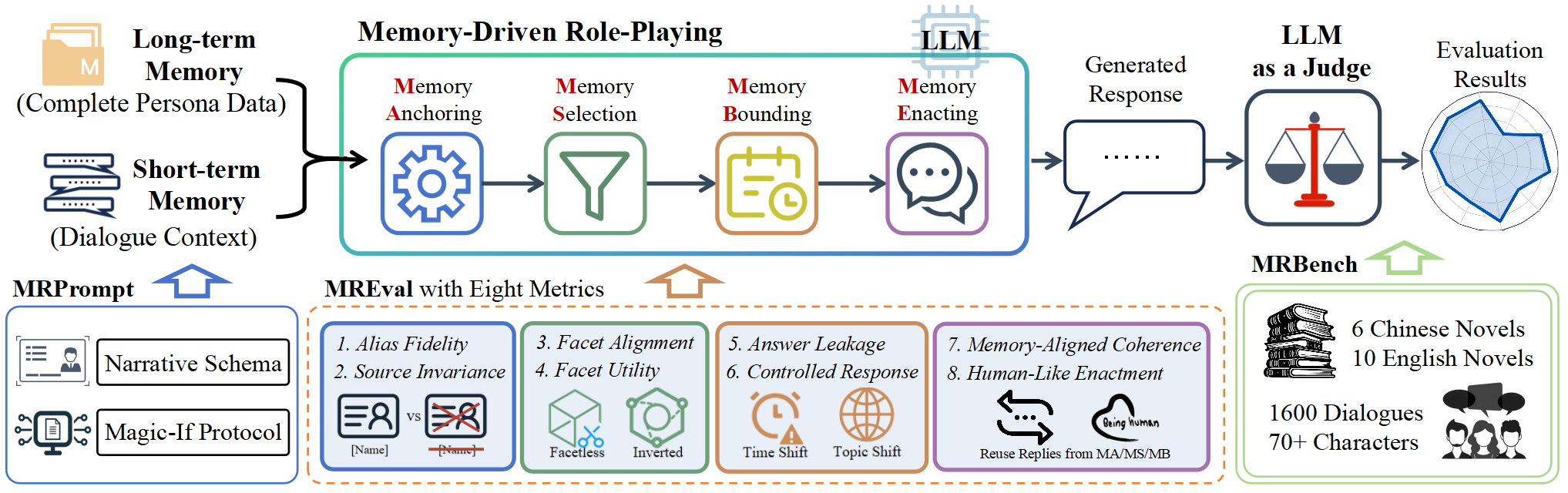} % pipeline.png
  \vspace{-5mm}
  \caption{\textbf{Overview}.
Given two parts of memory, an LLM performs memory-driven role-playing via four stages to generate an in-character response.
\textbf{MRPrompt} provides structured persona memory and a memory utilization protocol.
\textbf{MREval} scores eight stage-aligned metrics on the bilingual benchmark \textbf{MRBench}, by using an LLM-as-a-judge to assign per-metric scores.}
  \label{fig:overview}
\vspace{-5mm}
\end{figure*}

\paragraph{Role-Playing Methods and Memory-Oriented Mechanisms.}
Role-playing controllability is improved via richer persona representations, alignment/adaptation, and prompt-level controllers \citep{yu-etal-2025-beyond,lu2024superpositions,he-etal-2025-crab,wang-etal-2025-coser,duan-etal-2025-orpp,ruangtanusak2025talk,tang2025thinking}, as well as training-based specialization and persona refinement \citep{yu-etal-2024-neeko,yang2025hycora,yao2025dprf,fang2025charm}.
Memory-oriented mechanisms further introduce explicit retrieval or long-context organization for sustained role-play
%(e.g., RoleRAG, MOOM)\You{Do we need to delete these two examples?} 
~\citep{wang2025rolerag,chen-etal-2025-moom,huang2024emotionalrag,zhang2025teenempath}.
Unlike method-centric work, ours is benchmark-centric and diagnostic: we provide MRBench+MREval for stage-wise diagnosis and a prompt-only MRPrompt that standardizes structured persona memory and the reasoning guidance protocol.

\section{Methodology}

\begin{figure*}[t]
  \centering
  \includegraphics[width=\textwidth]{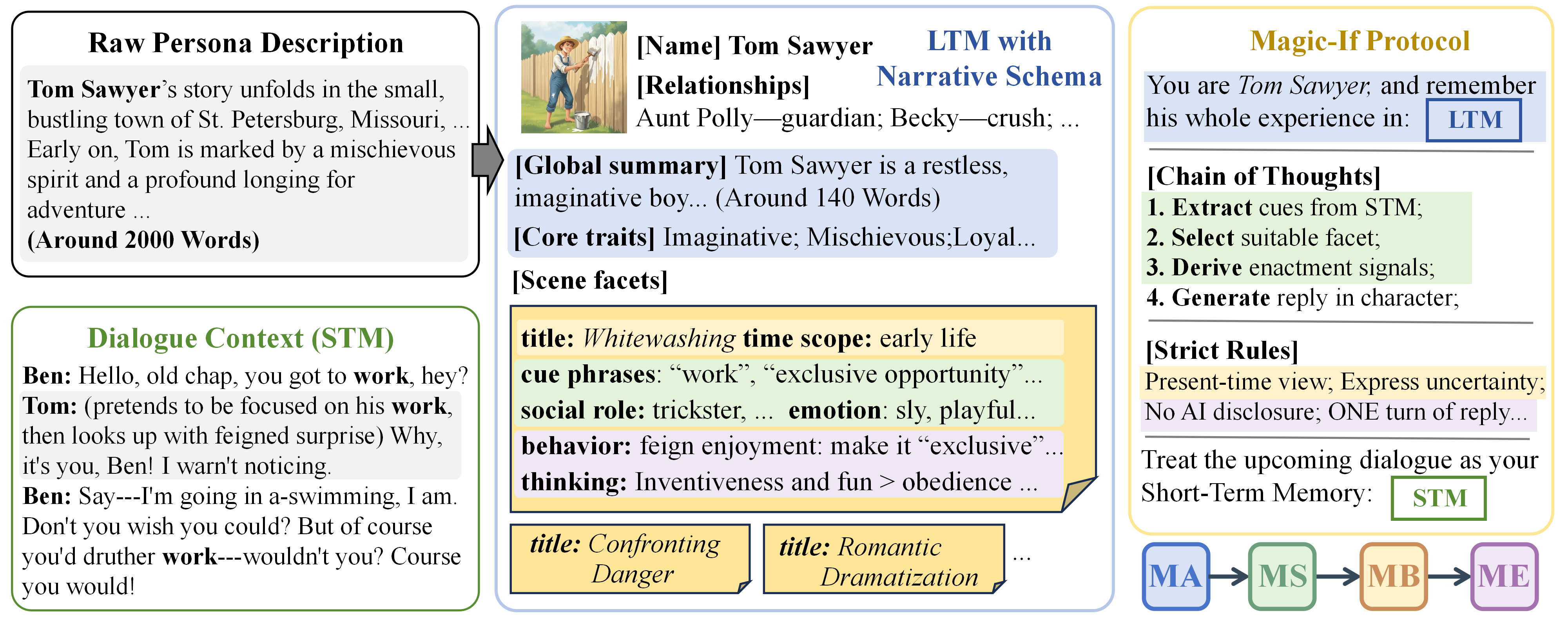}
  \caption{\textbf{MRPrompt}.
The raw persona description is structured as LTM via Narrative Schema and provided together with STM for role-playing.
Magic-If Protocol guides an LLM to generate responses following four stages.}
  \label{fig:mrprompt}
  \vspace{-5mm}
\end{figure*}

\subsection{Memory-Driven Role-Playing Task}
\label{sec:mcrp}

Given a specified character, we first formalize the Memory-Driven Role-Playing (MDRP) task as:
$\hat{y} \sim p_{\theta}(\, \cdot \mid \mathcal{M}_L, \mathcal{M}_S)$, 
where $p_{\theta}$ is the LLM's distribution parameterized by $\theta$. 
The in-character response $\hat{y}$ is generated by conditioning on the two memory inputs instantiated as follows:
\begin{itemize}[noitemsep,topsep=0pt,parsep=0pt,partopsep=0pt,leftmargin=10pt]
    \item \textbf{Long-Term Memory} (LTM) $\mathcal{M}_L$ is a finite set of persona facets: $\mathcal{M}_L = \{\phi_1, \phi_2, ..., \phi_N\}$. Each facet $\phi_i$ represents a coherent unit of persona knowledge, such as a core character trait together with its context-dependent expressions.
    \item \textbf{Short-Term Memory} (STM) $\mathcal{M}_S$ is an ordered sequence of the $K$ most recent dialogue turns: $\mathcal{M}_S = [u_1, u_2, ..., u_K]$. Each turn $u_i$ is a tuple $(r_i, c_i)$ consisting of the speaker role $r_i$ and the utterance content $c_i$, and the last turn $u_K$ is always the interlocutor's latest utterance (i.e., the model responds next).
\end{itemize}

This formulation redefines role-playing as a problem of \emph{contextual memory retrieval and application}, wherein the model must utilize the STM context $\mathcal{M}_S$ to select and apply the relevant knowledge from the LTM persona store $\mathcal{M}_L$.
Building on this foundation, we develop subsequent diagnostic and prompting methods. 
An end-to-end overview of the MDRP framework, including its core components MRPrompt, MREval, and MRBench, is provided in Figure~\ref{fig:overview} and elaborated in the following sections.

\iffalse
Under the MDRP paradigm, we contribute: (i) \textbf{MRPrompt}, which structures a flat persona into queryable \emph{Long-Term Memory} (LTM; $\mathcal{M}_L$) with core traits and contextualized facets and provides an LTM--STM interface prompt to guide cue-driven retrieval from \emph{Short-Term Memory} (STM; $\mathcal{M}_S$); (ii) \textbf{MREval}, a diagnostic evaluation framework that decomposes role-playing into four abilities, each measured by two metrics; and (iii) \textbf{MRBench}, a bilingual benchmark that instantiates MDRP with controlled variants of $\mathcal{M}_L$ and $\mathcal{M}_S$.
Figure~\ref{fig:overview} provides an end-to-end overview of MDRP together with MRPrompt, MREval, and MRBench, while Figure~\ref{fig:mrprompt} illustrates a concrete MRPrompt instantiation of facet-structured LTM and the Magic-If protocol for cue-driven, boundary-aware enactment.
\fi

\subsection{MREval: Evaluation Framework}
\label{sec:mdrp_ability}

% You: 仅将命名进行了修改
To enable the diagnostic assessment of LLMs within the Memory-Driven Role-Playing (MDRP) paradigm, we introduce the MREval framework. 
MREval decomposes the process of in-character response generation into \textbf{four sequential, measurable abilities}, corresponding to key stages in the utilization of persona knowledge from a human memory perspective~\citep{baddeley-1992-working,tulving-thomson-1973-encoding}.
Grounded in the MDRP formulation, we define four core abilities as follows:

\begin{itemize}[noitemsep,topsep=0pt,parsep=0pt,partopsep=0pt,leftmargin=10pt]
  \item \textbf{Memory-Anchoring (MA):} The model's ability to anchor its behavior to the designated persona in LTM, rather than relying on superficial cues or pretrained priors.
  \item \textbf{Memory-Selecting (MS):} The model's ability to extract cues from the STM dialogue context and retrieve relevant persona facets from the LTM based on those cues.
  \item \textbf{Memory-Bounding (MB):} The model's ability to adhere to the temporal and epistemic boundaries in persona knowledge, preventing the leakage of inaccessible or out-of-scope information.
  \item \textbf{Memory-Enacting (ME):} The model's ability to transform the selected and bounded persona knowledge into a coherent, natural, and human-like in-character utterance.
\end{itemize}

\iffalse
\begin{itemize}[noitemsep,topsep=0pt,parsep=0pt,partopsep=0pt,leftmargin=10pt]
  \item \textbf{Memory-Grounding (MG):} The model's ability to acquire and anchor its behavior to the provided persona knowledge in Long-Term Memory, rather than relying on superficial cues or pretrained priors.
  \item \textbf{Memory-Selection (MS):} The model's ability to use Short-Term Memory (dialogue context) as a cue to retrieve and activate the most relevant facet(s) from the structured persona LTM.
  \item \textbf{Memory-Boundary (MB):} The model's ability to adhere to the temporal and epistemic constraints inherent to the persona knowledge, avoiding leaks of inaccessible information.
  \item \textbf{Memory-Enactment (ME):} The model's ability to transform the selected, bounded persona knowledge into a coherent, natural, and human-like in-character utterance.
\end{itemize}
\fi

Each ability is operationalized by two fine-grained metrics, yielding eight diagnostic scores.
Table~\ref{tab:mdeval_metrics} summarizes all metrics.
Concretely, for each metric, we score model outputs with an LLM-as-a-judge and then linearly calibrate the judge scores to the \emph{human rating scale} (see Appendix~\ref{app:judge_validation}), yielding an ordinal 1--10 Likert-style rating for the corresponding criterion \citep{likert1932technique}.
The full scoring rubrics are provided in Appendix~\ref{app:metrics_rubrics}.

% You: 仅将命名进行了修改
Overall, MREval provides an eight-dimensional diagnostic profile for MDRP by decomposing in-character generation into four abilities.
This enables stage-wise attribution of failures---from grounding and facet retrieval to boundary control and final enactment---rather than relying on a single holistic quality score.

\begin{table*}[t]
  \centering
  \small
  \setlength{\tabcolsep}{5pt}
  \renewcommand{\arraystretch}{1.15}
  \caption{\textbf{MREval abilities and metrics.}
MREval decomposes memory-driven role-playing into four abilities (MA/MS/MB/ME), each measured by two calibrated Likert-style metrics (eight in total).
}
  \label{tab:mdeval_metrics}
  \resizebox{\textwidth}{!}{
  %\begin{tabular}{llp{0.73\textwidth}}
  \begin{tabular}{
    >{\raggedright\arraybackslash}p{0.06\textwidth}
    >{\raggedright\arraybackslash}p{0.22\textwidth}
    p{0.73\textwidth}
  }
    \toprule
    \textbf{Ability} & \textbf{Metric} & \textbf{Definition} \\
    \hline
    \multirow{4}{*}{\textbf{MA}} & 
    \textbf{Source Invariance (SI)} &
    Measures consistency between the response generated with the anonymized persona $\hat{y}^{\text{anon}}$ and the original one $\hat{y}$. A high score indicates grounding in persona semantics, not name priors. \\
    & \textbf{Alias Fidelity (AF)} &
    Assesses whether the behavior under an anonymized persona $\hat{y}^{\text{anon}}$ remains faithful to the original intended character, using the ground-truth response $\hat{y}^{\text{gold}}$ as an anchor. \\
    \hline
    \multirow{4}{*}{\textbf{MS}} &   
    \textbf{Facet Alignment (FA)} &
    Quantifies the model's precision in selecting the correct scene facet by contrasting responses under the true $\mathcal{M}_L$ versus a counterfactual (inverted) LTM $\mathcal{M}_L^{\text{anti}}$. \\
    & \textbf{Facet Utility (FU)} &
    Measures the improvement gained by including scene-specific facets in the LTM, compared to a scene-ablated LTM $\mathcal{M}_L^{\text{no-scene}}$. \\
    \hline
    \multirow{4}{*}{\textbf{MB}} & 
    \textbf{Answer Leakage (AL)} &
    Scores the model's ability to avoid generating a forbidden reference answer $\hat{y}^{\text{out}}$ when presented with an out-of-scope prompt $c_K^{\text{out}}$ (e.g., a future plot spoiler). \\
    & \textbf{Controlled Response (CR)} &
    Assesses the appropriateness of the model's response strategy to out-of-scope prompts, favoring expressions of uncertainty, refusal, or grounded speculation over confident fabrication. \\
    \hline
    \multirow{4}{*}{\textbf{ME}} & 
    \textbf{Memory-Aligned Coherence (MAC)} &
    Rates the logical and topical coherence of the response with respect to the activated memory and context. \\
    & \textbf{Human-Like\ \ \ \ \ \ \ Enactment (HLE)} &
    Rates the naturalness, tonal appropriateness, and conversational fluency of the response, ensuring it embodies a human-like utterance consistent with the persona. \\
    \bottomrule
  \end{tabular}}
\vspace{-5mm}
\end{table*}

\subsection{MRPrompt: Inference-Time Method}
\label{sec:mrprompt}

\paragraph{Intuition: Stanislavski-inspired Memory Recall.}
Under MDRP, faithful role-playing is not only a matter of \emph{style} but a problem of \emph{contextual memory recall}.
Stanislavski's system emphasizes that authentic embodiment arises from recalling \emph{experiential memory} under the current \emph{given circumstances}, guided by an explicit rehearsal-time \emph{action plan} (e.g., ``magic if'') rather than unconstrained improvisation.
Analogously, as illustrated in Figure~\ref{fig:mrprompt}, MRPrompt equips an LLM with (i) a structured persona memory store as \emph{Long-Term Memory} (LTM) and (ii) an explicit inference-time protocol that extracts cues from the ongoing dialogue and recalls persona memory for faithful role-playing. MRPrompt is purely prompt-based at inference time, requiring no parameter updates and no external retrieval or tool use, and thus can be directly applied to any instruction-tuned LLM. 

\paragraph{(1) Narrative Schema: hierarchical, queryable persona LTM.}
MRPrompt first replaces flat trait lists with a structured Narrative Schema for LTM $\mathcal{M}_L$.
The schema organizes persona information into \emph{identity fields}, a \emph{global summary}, \emph{core traits}, and a set of \emph{scene facets} that encode context-dependent expressions under recurring situations.
Each facet is \emph{cue-addressable}: it binds cue keys (e.g., \textit{situation}, \textit{cue\_phrases}) to enactment signals (e.g., \textit{social\_role}, \textit{emotional\_state}, \textit{behavior\_pattern}, \textit{thinking\_pattern}) and boundary anchors (e.g., \textit{time\_scope}, \textit{conflict\_with\_core}).
Importantly, the fields are designed to align with our diagnostic abilities:
core traits support MA (anchoring to persona semantics beyond name priors);
cue keys support MS (selecting the relevant facet under dialogue cues);
boundary anchors support MB (enforcing temporal/epistemic constraints);
and enactment signals support ME (realizing recalled knowledge into natural utterances).
This schema mitigates style averaging and local out-of-character drift by enabling \emph{selective facet activation}, and makes memory use attributable to concrete fields (schema in Appendix~\ref{app:facet_schema}).

% \paragraph{(2) Magic-If Protocol: explicit LTM--STM control for cue-driven retrieval and boundary-aware enactment.}
\paragraph{(2) Magic-If Protocol: an explicit LTM--STM control protocol.} % for cue-driven selection, boundary-respecting generation, and faithful enactment
Building on the Narrative Schema, MRPrompt introduces a Magic-If Protocol as an explicit LTM--STM interface prompt.
It frames $\mathcal{M}_L$ as the character's internal memory store and the multi-turn dialogue $\mathcal{M}_S$ as the \emph{given circumstances} (STM cues), and specifies a minimal inference-time policy:
(\textbf{i}) establish a stable identity by grounding in core traits (MA);
(\textbf{ii}) interpret STM cues to select and activate the most relevant scene facet(s) (MS);
(\textbf{iii}) apply boundary anchors to remain within the character's temporal/epistemic knowledge and avoid out-of-scope leakage (e.g., spoilers or inaccessible claims) (MB);
(\textbf{iv}) enact the selected and bounded memory into a coherent, human-like in-character reply (ME).
Crucially, this protocol turns the LTM--STM interaction into a \emph{controllable mechanism} rather than an implicit behavior emergent from prompting, enabling stage-wise attribution and systematic ablations.

% \paragraph{Why this matters.}
MRPrompt contributes a theory-grounded separation of \emph{representation} (Narrative Schema) and \emph{control} (Magic-If Protocol):
the former makes persona memory queryable and cue-addressable, while the latter makes retrieval and boundary enforcement explicit and auditable.
Together, they support selective, bounded, and diagnosable persona utilization aligned with MA/MS/MB/ME, which is precisely what MDRP and MREval are designed to probe.
Complete prompt templates are provided in Appendix~\ref{app:prompt:roleplay}.

\subsection{MRBench: Benchmark Construction}
\label{sec:method_benchmark}
 
To evaluate MDRP in a stage-diagnostic manner, we need benchmark splits for MA/MS/MB/ME where performance differences are attributable to the \emph{targeted memory stage} rather than uncontrolled scene variation.

MRBench is constructed based on the principle of context reuse and minimal perturbation: we maximize the reuse of a shared short-term memory (STM) context pool $\mathcal{M}_S$, and create paired test conditions by applying minimal, targeted edits whenever possible to either the long-term memory (LTM) context $\mathcal{M}_L$ or the final-turn query $c_K$. This design mitigates scene-induced confounds and substantially reduces the annotation overhead compared to building separate, scene-diverse datasets for each ability. As a result, we obtain controlled $(\mathcal{M}_L,\mathcal{M}_S)$ pairs that cleanly isolate the effects of MA, MS, and MB, while ME is evaluated on the same model outputs to avoid confounding factors.
The full construction procedure is provided in Appendix~\ref{app:benchmark}.

\section{Experiments}
\label{sec:experiments}

We empirically study Memory-Driven Role-Playing (MDRP) under MREval, focusing on eight research questions:
\textbf{RQ1:} How do small LLM backbones perform on the MDRP task using MRBench?
\textbf{RQ2:} Does MRPrompt outperform baselines across diverse small LLMs?
\textbf{RQ3:} What is the contribution of each component in MRPrompt?
\textbf{RQ4:} How does MRPrompt affect the gap between small-scale and closed-source LLMs on MDRP? 
\textbf{RQ5:} How reliable is the LLM-as-a-judge for MREval?
\textbf{RQ6:} How do upstream abilities (MA/MS/MB) relate to and predict the ME ability? 
\textbf{RQ7:} What is the token cost of the baseline prompts vs.\ MRPrompt?
\textbf{RQ8:} How do baselines and MRPrompt differ in qualitative MDRP case studies?
% \textbf{RQ8:} How do closed-source LLMs compare with small-scale LLMs on MDRP under MRPrompt? 
%, as measured by MREval?
Due to space limitations, discussions of \textbf{RQ7}--\textbf{RQ8} are provided in the Appendix.

\begin{table*}[t]
  \centering
  \scriptsize
  \setlength{\tabcolsep}{2pt}
  \renewcommand{\arraystretch}{1.08}
  \caption{\textbf{Experimental results on MRBench.}
  Human-calibrated GPT-4.1-mini scores (higher is better) for the eight MREval metrics in English/Chinese (en/zh). Avg.\ Score is the mean over all 8 metrics $\times$ 2 languages.}
  %Avg.\ Score (higher is better) is the arithmetic mean over all 16 metric--language columns (8 metrics $\times$ \{en, zh\}) for each persona condition.}
  \label{tab:rq1_calibrated_scores}
  \resizebox{\textwidth}{!}{%
  \begin{tabular}{llccccccccccccccccc}
    \toprule
    \multirow{2}{*}{Model} & \multirow{2}{*}{Persona} & \multicolumn{2}{c}{MA-SI} & \multicolumn{2}{c}{MA-AF}
      & \multicolumn{2}{c}{MS-FA} & \multicolumn{2}{c}{MS-FU}
      & \multicolumn{2}{c}{MB-AL}  & \multicolumn{2}{c}{MB-CR}
      & \multicolumn{2}{c}{ME-MAC}  & \multicolumn{2}{c}{ME-HLE} & \multirow{2}{*}{Avg.\ Score}\\
    & & en & zh & en & zh & en & zh & en & zh & en & zh & en & zh & en & zh & en & zh & \\
    \hline

\multirow{3}{*}{Qwen3-0.6B}
 & Base & 7.38 & 7.21 & 5.94 & 7.80 & 6.13 & 8.35 & 7.45 & 7.47 & 8.46 & 8.08 & 6.20 & 6.45 & 5.99 & 6.60 & 5.28 & 5.83 & 6.91 \\
 & Card & 7.88 & 7.93 & 6.12 & 7.63 & 5.51 & 8.18 & 7.20 & 7.50 & 8.53 & 8.21 & 6.37 & 6.37 & 6.22 & 6.79 & 5.52 & 6.06 & 7.00 \\
 & MRPrompt & \textbf{8.02} & \textbf{8.06} & \textbf{6.67} & \textbf{7.81} & \textbf{6.46} & \textbf{8.69} & \textbf{7.63} & \textbf{7.77} & \textbf{8.65} & \textbf{8.23} & \textbf{6.54} & \textbf{6.56} & \textbf{6.33} & \textbf{6.97} & \textbf{5.86} & \textbf{6.35} & \textbf{7.29} \\
\hline

\multirow{3}{*}{Qwen3-4B}
 & Base & 8.65 & 7.80 & 7.54 & 7.99 & \textbf{7.97} & 8.63 & 8.59 & 8.23 & 8.79 & 8.53 & 6.55 & \textbf{7.04} & 7.55 & 7.43 & 7.20 & 7.17 & 7.85 \\
 & Card & 8.72 & 8.53 & 7.57 & 7.80 & 6.72 & 8.40 & 8.33 & 8.16 & 8.78 & 8.48 & 6.70 & 6.95 & 7.73 & 7.41 & 7.39 & 7.21 & 7.81 \\
 & MRPrompt & \textbf{8.88} & \textbf{8.61} & \textbf{7.69} & \textbf{8.23} & 7.61 & \textbf{8.81} & \textbf{8.69} & \textbf{8.57} & \textbf{8.85} & \textbf{8.56} & \textbf{6.84} & 6.99 & \textbf{8.07} & \textbf{7.63} & \textbf{7.73} & \textbf{7.42} & \textbf{8.07} \\
\hline

\multirow{3}{*}{Qwen3-8B}
 & Base & 8.67 & 8.08 & 7.52 & 8.02 & \textbf{8.03} & 8.60 & \textbf{8.83} & 8.23 & 8.85 & 8.51 & 6.63 & \textbf{7.36} & 7.83 & 7.34 & 7.50 & 7.12 & 7.95 \\
 & Card & 8.88 & 8.49 & 7.66 & 8.05 & 7.09 & 8.47 & 8.54 & 8.28 & 8.81 & \textbf{8.55} & 6.64 & 6.98 & 7.91 & 7.52 & 7.59 & 7.16 & 7.91 \\
 & MRPrompt & \textbf{8.97} & \textbf{8.73} & \textbf{7.76} & \textbf{8.25} & 7.56 & \textbf{8.99} & 8.64 & \textbf{8.56} & \textbf{8.88} & 8.52 & \textbf{6.92} & 7.07 & \textbf{8.13} & \textbf{7.64} & \textbf{7.83} & \textbf{7.42} & \textbf{8.12} \\
\hline

\multirow{3}{*}{GLM-4-9B-Chat}
 & Base & 8.59 & 8.13 & 7.41 & 8.10 & 7.73 & 8.67 & \textbf{8.65} & 8.34 & 8.78 & 8.36 & 6.49 & 7.03 & 7.66 & 7.40 & 7.19 & 7.18 & 7.86 \\
 & Card & \textbf{8.87} & 8.61 & 7.48 & 7.94 & 6.88 & 8.34 & 8.42 & 8.35 & 8.82 & 8.42 & 6.46 & 6.93 & 7.85 & 7.42 & 7.41 & 7.17 & 7.84 \\
 & MRPrompt & 8.77 & \textbf{8.70} & \textbf{7.59} & \textbf{8.23} & \textbf{7.93} & \textbf{8.88} & 8.53 & \textbf{8.52} & \textbf{8.83} & \textbf{8.54} & \textbf{6.76} & \textbf{7.07} & \textbf{8.02} & \textbf{7.63} & \textbf{7.53} & \textbf{7.41} & \textbf{8.06} \\
\hline

\multirow{3}{*}{Llama-3-8B-Instruct}
 & Base & 8.23 & 7.40 & 7.13 & \textbf{7.80} & \textbf{7.02} & 8.62 & \textbf{8.37} & 7.52 & 8.78 & 8.41 & 6.50 & 6.95 & 7.50 & 7.00 & 6.74 & 6.57 & 7.53 \\
 & Card & \textbf{8.78} & 8.09 & 7.33 & 7.32 & 6.29 & 8.55 & 8.13 & 7.73 & 8.81 & 8.28 & 6.69 & 6.84 & 7.75 & 6.84 & 7.05 & 6.37 & 7.55 \\
 & MRPrompt & 8.69 & \textbf{8.20} & \textbf{7.51} & 7.68 & 6.83 & \textbf{8.63} & 8.32 & \textbf{8.03} & \textbf{8.87} & \textbf{8.47} & \textbf{6.93} & \textbf{7.13} & \textbf{8.05} & \textbf{7.20} & \textbf{7.37} & \textbf{6.69} & \textbf{7.79} \\
\hline

\multirow{3}{*}{Llama-3.2-3B-Instruct}
 & Base & 8.43 & 5.56 & 7.12 & 7.00 & 7.63 & 7.93 & \textbf{8.45} & 6.51 & 8.77 & 7.83 & 6.74 & 6.35 & 7.39 & 6.18 & 6.79 & 5.15 & 7.11 \\
 & Card & 8.55 & 6.43 & 7.21 & 6.77 & 6.53 & 8.14 & 8.36 & 7.06 & 8.85 & 7.75 & 6.73 & 6.47 & 7.63 & 6.24 & 6.99 & 5.17 & 7.18 \\
 & MRPrompt & \textbf{8.73} & \textbf{7.35} & \textbf{7.50} & \textbf{7.19} & \textbf{7.69} & \textbf{8.63} & \textbf{8.45} & \textbf{7.36} & \textbf{8.87} & \textbf{8.19} & \textbf{7.10} & \textbf{6.65} & \textbf{8.04} & \textbf{6.60} & \textbf{7.55} & \textbf{5.88} & \textbf{7.61} \\
\hline

\multirow{3}{*}{InternLM2.5-7B-Chat}
 & Base & 7.97 & 6.97 & 7.11 & 7.96 & 7.49 & 8.80 & 8.39 & 8.09 & 8.64 & 8.36 & 6.24 & 6.94 & 6.96 & 7.31 & 6.74 & 6.88 & 7.55 \\
 & Card & \textbf{8.26} & 7.76 & 7.17 & 7.85 & 7.12 & 8.70 & 8.17 & 7.94 & 8.71 & 8.40 & 6.55 & \textbf{7.07} & 7.39 & 7.26 & 6.81 & 6.87 & 7.63 \\
 & MRPrompt & 8.08 & \textbf{7.99} & \textbf{7.31} & \textbf{7.98} & \textbf{7.89} & \textbf{9.10} & \textbf{8.40} & \textbf{8.33} & \textbf{8.75} & \textbf{8.46} & \textbf{6.71} & 6.92 & \textbf{7.71} & \textbf{7.36} & \textbf{7.13} & \textbf{6.95} & \textbf{7.82}\\
\bottomrule
  \end{tabular}}
  \vspace{-3mm}
\end{table*}

\subsection{Experimental Setup}
\label{sec:exp_setup}

\paragraph{Benchmark.}
We construct MRBench, a bilingual benchmark instantiated under MDRP via controlled variants of the inputs $(\mathcal{M}_L,\mathcal{M}_S)$ to target MA/MS/MB/ME.
Sourced from a collection of ten English and six Chinese novels,
MRBench contains 200 English and 200 Chinese instances per ability family; all instances are single-turn.
The detailed data source information is provided in Appendix~\ref{app:benchmark}.

\paragraph{Models.}
We evaluate \textbf{twelve} instruction-tuned LLMs that support Chinese--English role-play, spanning open-source and API-based families.
Our \textbf{seven open-source} backbones include Llama-3-8B-Instruct~\citep{llama3modelcard}, Llama-3.2-3B-Instruct~\citep{meta2024llama32}, Qwen3-\{0.6B, 4B, 8B\}~\citep{qwen3technicalreport}, GLM-4-9B-Chat~\citep{teamglm2024chatglm}, and InternLM2.5-7B-chat~\citep{cai2024internlm2}.
For \textbf{closed-source} comparison, we additionally include five API-based LLMs: GPT-5.2, GLM-4.7, DeepSeek-Chat, Qwen3-Max, and Doubao-Seed-1.6-250615.\footnote{Official model documentation pages: \citep{openai2025gpt52model,zai2025glm47overview,deepseek2025pricing,qwen2025qwen3max,volcengine2025doubao_seed16}.}

\paragraph{Hyperparameters.}
All models are used \emph{as is} (no fine-tuning), with prompt-only interventions.
We perform no hyperparameter search.
In generation, we set temperature to $T{=}0.7$; other decoding parameters follow provider defaults when not explicitly configurable.

\paragraph{Compared prompting conditions.}
We compare two baseline persona prompting formats with MRPrompt.
For each character, we derive persona content from the same source materials and keep the same STM input $\mathcal{M}_S$ fixed; conditions differ only in persona representation and usage guidance.
Specifically, we use (i) \textbf{Base}, a narrative persona summary; (ii) \textbf{Card}, a lightweight profile-card baseline following the CharacterEval persona format~\citep{tu-etal-2024-charactereval}; and (iii) \textbf{MRPrompt}, our facet-structured LTM with an explicit LTM--STM interface.
Construction prompts are provided in Appendix~\ref{app:prompt:profiles}, and persona specifications with examples are given in Appendix~\ref{app:persona_construction}.

\paragraph{Reporting.}
Unless otherwise noted, we report \emph{mean} scores over instances for each split and language.
Due to evaluation cost, results are obtained from a single run per model and prompting condition under fixed decoding settings.

\subsection{Main Experiments (RQ1 \& RQ2)}
\label{sec:main:experiment}

We evaluate seven off-the-shelf instruction-tuned LLMs on MRBench under MREval, comparing two baselines (Base, Card) with MRPrompt.
Unless otherwise stated, all results reported in the main paper are based on human-calibrated judge scores. Raw judge scores are provided in Appendix~\ref{app:raw_scores}, and calibration details are described in \S\ref{sec:judge_validation}.
From the experimental results presented in Table~\ref{tab:rq1_calibrated_scores}, three key observations can be drawn.

\textbf{(1) Scaling helps, but not uniformly across stages.}
Overall performance tends to improve with model capacity across MA/MS/MB/ME and both languages, suggesting stable capability differences under MDRP.
However, the eight-metric breakdown is non-uniform: gains are often stronger in MA/ME than in MS/MB, indicating that \emph{anchoring} and \emph{surface realization} scale more readily than \emph{cue-driven facet recall} and \emph{boundary control}.

\textbf{(2) Baseline structuring (Card) is not enough.}
Comparing Base and Card, Avg.\ Score changes are modest and mixed across backbones, and the gains are uneven across abilities.
Card-style formatting can help MA/ME (persona uptake and response organization), yet it does not reliably improve MS or MB, leaving the core MDRP failure modes---\emph{facet mis-recall} and \emph{boundary violations}---largely unresolved.

\textbf{(3) MRPrompt yields diagnostic, stage-specific gains.}
MRPrompt achieves the highest Avg.\ Score for every backbone relative to Base/Card.
Crucially, improvements are diagnostic rather than cosmetic: gains frequently concentrate in MS and MB, while MA also improves (especially for smaller models), and ME increases more modestly but consistently.

Overall, the MRPrompt design---a retrieval-oriented structured LTM plus an explicit LTM--STM control protocol---primarily strengthens MA/MS/MB and is consistently accompanied by improved enactment (ME).

\begin{table}[t]
  \centering
  \scriptsize
  \setlength{\tabcolsep}{3pt}
  \renewcommand{\arraystretch}{1.05}
  \caption{\textbf{Component ablation.}
%Results on Qwen3-4B and GLM-4-9B-Chat comparing Base, +Protocol, +Schema, and MRPrompt.
Each ability score (MA/MS/MB/ME) is averaged over its own 2 metrics $\times$ 2 languages (en/zh).
%Avg.\ Score is the mean over MG/MS/MB/ME; Bold denotes the best setting within each backbone.
}
  \vspace{-3mm}
  \label{tab:ablation}
  \resizebox{\columnwidth}{!}{%
  \begin{tabular}{llccccc}
    \toprule
    Model & Condition & MA & MS & MB & ME & Avg.\ Score \\
    \hline
    \multirow{4}{*}{Qwen3-4B}
      & Base                    & 8.00 & 8.36 & 7.73 & 7.34 & 7.85 \\
      & +Protocol           & 8.17 & \textbf{8.42} & 7.71 & 7.41 & 7.93 \\
      & +Schema         & 8.30 & 8.24 & 7.77 & 7.58 & 7.97 \\
      & MRPrompt                    & \textbf{8.35} & \textbf{8.42} & \textbf{7.81} & \textbf{7.71} & \textbf{8.07} \\
    \hline
    \multirow{4}{*}{GLM-4-9B-Chat}
      & Base                    & 8.06 & 8.35 & 7.67 & 7.36 & 7.86 \\
      & +Protocol           & 8.11 & 8.34 & 7.68 & 7.36 & 7.87 \\
      & +Schema         & \textbf{8.32} & 8.44 & 7.72 & \textbf{7.67} & 8.04 \\
      & MRPrompt                    & \textbf{8.32} & \textbf{8.47} & \textbf{7.80} & 7.65 & \textbf{8.06} \\
    \bottomrule
  \end{tabular}}
\end{table}

\subsection{Component Ablations (RQ3)}
\label{sec:rq3}

We ablate MRPrompt on two backbones (Qwen3-4B and GLM-4-9B-Chat) by isolating its two components: facet-structured LTM (Schema) and the LTM--STM interface (Protocol).
We compare four conditions---Base, +Protocol, +Schema, and MRPrompt---and report ability-level averages in Table~\ref{tab:ablation}.
Overall, Schema contributes the larger \emph{overall} gain: replacing a narrative persona with facet-structured LTM yields a stronger improvement in Avg.\ Score on both backbones, reflecting more reliable persona anchoring and downstream enactment.
Meanwhile, Protocol is complementary and more model-dependent: it brings limited change on top of Base (especially on GLM-4-9B-Chat), but provides further gains once structured LTM is present, making MRPrompt best overall.
This pattern matches MDRP’s division of labor: Schema builds a cue-addressable memory space, and Protocol more reliably elicits goal-directed retrieval and boundary-aware responses.

% requires: \usepackage{booktabs,graphicx}
\begin{table}[t]
  \centering
  \scriptsize
  \setlength{\tabcolsep}{3.2pt}
  \renewcommand{\arraystretch}{1.05}
  \caption{\textbf{Backbone comparison (Large vs.\ Small).}
%MA/MS/MB/ME average over their 2 metrics $\times$ 2 languages (en/zh), and 
The best and second-best results per column are bold and underlined.}
  \vspace{-3mm}
  \label{tab:backbone_base_vs_mrprompt_flat}
  \resizebox{\columnwidth}{!}{%
  \begin{tabular}{lcccccc}
    \toprule
    Model & Condition & MA & MS & MB & ME & Avg.\ Score \\
    \hline
    Qwen3-8B & Base          & 8.07 & 8.42 & 7.84 & 7.45 & 7.95 \\
    GLM-4-9B-Chat & Base     & 8.06 & 8.35 & 7.67 & 7.36 & 7.86 \\
    Qwen3-8B & MRPrompt          & \textbf{8.43} & 8.44 & 7.85 & \textbf{7.76} & \underline{8.12} \\
    GLM-4-9B-Chat & MRPrompt     & 8.32 & 8.47 & 7.80 & 7.65 & 8.06 \\
    \hline
    Qwen3-Max & Base         & 8.31 & 8.41 & 7.89 & 7.73 & 8.08 \\
    GLM-4.7 & Base           & 8.31 & \textbf{8.60} & 7.87 & 7.65 & 8.11 \\
    Qwen3-Max & MRPrompt         & 8.33 & 8.22 & \textbf{8.02} & 7.73 & 8.07 \\
    GLM-4.7 & MRPrompt           & \underline{8.34} & \underline{8.54} & \underline{7.91} & \underline{7.75} & \textbf{8.13} \\
    \bottomrule
  \end{tabular}}
\end{table}

\subsection{Large vs.\ Small LLMs (RQ4)}
\label{sec:L:LLM}

\begin{figure}[t]
  \centering
  \begin{minipage}[t]{0.49\columnwidth}
    \centering
    \includegraphics[width=\linewidth]{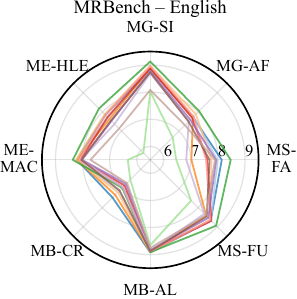}
    \scriptsize (a) English.
  \end{minipage}\hfill
  \begin{minipage}[t]{0.49\columnwidth}
    \centering
    \includegraphics[width=\linewidth]{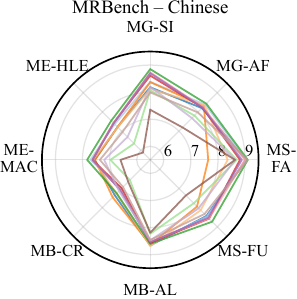}
    \scriptsize (b) Chinese.
  \end{minipage}

  \vspace{2pt}
  \includegraphics[width=\columnwidth]{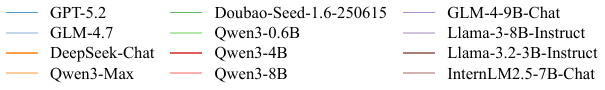}

  \caption{\textbf{All-model radar profiles on MRBench (MRPrompt).}
  Eight-axis MREval metric profiles for English and Chinese with a shared legend.}
  \label{fig:rq8_radar_allmodels}
\end{figure}

Under the same MRPrompt setup, Figure~\ref{fig:rq8_radar_allmodels} shows a clear capacity-shaped profile across both English and Chinese: closed-source SOTA models exhibit consistently strong, low-variance performance across upstream memory use and downstream enactment.
For instance, Doubao-Seed-1.6-250615 leads MA/MS while remaining top-tier on ME; GPT-5.2 is best on \textsc{CR} in English and ties the top tier on \textsc{AL}. 
In contrast, small models are more uneven: several mid-sized LLMs (e.g., Qwen3-8B/4B, GLM-4-9B-Chat) can approach SOTA on upstream anchoring and facet selection, yet lag more on ME, suggesting that structured prompting stabilises \emph{memory use} but cannot fully compensate for limited \emph{surface enactment} capacity; this bottleneck is most evident for the smallest model (Qwen3-0.6B). 
Constraint robustness is consistently harder than other upstream dimensions, with \textsc{CR} typically trailing \textsc{AL} across models; while overall trends align between English and Chinese, several metrics still show language-specific shifts (full details in Appendix~\ref{app:rq8_sota_vs_small}).

More importantly, \textbf{MRPrompt enables smaller LLMs to rival the performance of much larger counterparts.} As shown in Table~\ref{tab:backbone_base_vs_mrprompt_flat}, the Qwen3-8B equipped with MRPrompt achieves an average score of 8.12, surpassing the standard versions of the much larger GLM-4.7 (8.11) and Qwen3-Max (8.08) models, and nearly matching the SOTA performance of GLM-4.7 with MRPrompt (8.13). This result underscores the unique value of MRPrompt: effectively enhancing smaller LLMs to achieve faithful, high-quality role-playing that is competitive with cutting-edge, scaled-up models.

\iffalse
Under the same MRPrompt setup, Figure~\ref{fig:rq8_radar_allmodels} shows a clear capacity-shaped pattern in both English and Chinese: closed-source SOTA models deliver uniformly high, low-variance scores across upstream memory use and downstream enactment, whereas small models are much more uneven---mid-sized LLMs can be competitive on anchoring/selection but lag on ME, and the smallest model drops sharply on ME despite relatively stable upstream/constraint-aligned scores. A consistent bottleneck across scales is constraint robustness, with \textsc{CR} typically below \textsc{AL}; overall trends are similar across languages but exhibit metric-level shifts (full details in Appendix~\ref{app:rq8_sota_vs_small}).
\fi

\subsection{Judge Validation (RQ5)}
\label{sec:judge_validation}

We use GPT-4.1-mini as the LLM judge for all eight MREval metrics (1--10).
To assess reliability, we sample a bilingual validation set with 100 instances per metric (50 en / 50 zh; 800 total) and collect blind human ratings from a bilingual annotator.
Across all metric--language pairs, judge--human agreement is consistently strong. %Figure~\ref{fig:judge_corr_heatmap} visualises \emph{Pearson} correlations, while 

Table~\ref{tab:judge_corr_mgms} and Table~\ref{tab:judge_corr_mbme} report the full correlation statistics (Pearson/Spearman/Kendall and $p$-values). Meanwhile, mean-score differences suggest mild systematic bias (Table~\ref{tab:judge_means}), motivating per-metric calibration on the validation set.
Appendix~\ref{app:judge_validation} reports full details, calibration parameters, and additional visual checks.

\subsection{Ability Interactions (RQ6)}
\label{sec:rq6}

MREval decomposes MDRP into four stages, but effective role-playing should behave as a coupled pipeline.
We therefore test (i) whether scores of different abilities are correlated across evaluated instances, and (ii) whether stronger upstream memory abilities (MA/MS/MB) are associated with better downstream enactment (ME).

\begin{figure}[t]
  \centering
  \includegraphics[width=1.0\linewidth]{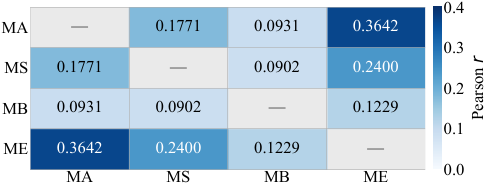}
  \caption{\textbf{Ability correlations.}
  % (pooled EN+ZH)
  Heatmap of Pearson correlations ($r$) between ability-level scores over multiple models. %  2,562 instances (
  %Diagonal self-correlations are omitted for readability, and the color scale is set to the off-diagonal range.
  }
  \label{fig:rq5_corr_all}
\end{figure}

\paragraph{Ability correlations.}
We aggregate the two metrics within each ability to obtain ability-level scores, and compute Pearson correlations over instances where all four abilities are available.
Figure~\ref{fig:rq5_corr_all} reports correlations on the pooled EN+ZH set: correlations are uniformly positive but weak-to-moderate, indicating that abilities are \emph{coupled yet non-redundant}.
Notably, MA exhibits the strongest association with ME, while MS and MB show weaker correlations with other abilities, suggesting that (a) anchoring in persona memory often translates to better surface enactment, but (b) \emph{selection} and \emph{boundary control} remain comparatively independent bottlenecks rather than automatically improving with general response quality.
Language-split analyses and additional statistics are provided in Appendix~\ref{app:rq5_language_corr}.

\paragraph{Upstream association.}
Aggregating MA/MS/MB into a single upstream score, we find a moderate positive association with ME on the pooled set ($r{=}0.35$), implying that better anchoring/selecting/bounding behavior tends to accompany improved enactment, but does not fully determine it.
Appendix~\ref{app:rq5_language_corr} reports language-split correlations and further details.

\section{Conclusion}
\label{sec:conclusion}

We study Memory-Driven Role-Playing (MDRP) from a memory perspective, where persona knowledge serves as LTM, and dialogue context serves as STM.
We introduce MREval, a diagnostic evaluation that decomposes MDRP into four abilities with eight Likert-style metrics, and construct a bilingual benchmark MRBench.
Building on this formulation, we propose MRPrompt to enable cue-driven retrieval and boundary-aware generation.
Experiments show consistent improvements over narrative and profile-card baselines, with the largest gains in selection and boundary-related diagnostics.
Overall, our results suggest that making memory representation and use explicit is an effective route to more faithful role-playing.

\section*{Acknowledgments}
This research is supported by the National Natural Science Foundation in China
(Grant: 62506100).
Dr. Yang Zhang did not receive financial support for this work from this grant or from any other external project. His contribution was undertaken as independent academic research during his appointment as a Research Fellow at Macquarie University.

\section*{Limitations}

\label{sec:limitation}

\textbf{Evaluation scope under MDRP.}
We evaluate MDRP in an \emph{episodic, single-turn} setting: for each instance, the model receives a fixed input pair $(\mathcal{M}_L,\mathcal{M}_S)$ and generates one in-character reply, with no \emph{within-session} memory update.
While this design enables clear, stage‑diagnostic comparisons under the MREval framework, it has not yet addressed interactive scenarios where persona memory is gradually accumulated, revised, or negotiated over time—such as in multi‑turn conversation carryover, shifting user goals, or multi‑character coordination. In these more dynamic regimes, the coupling among four memory abilities may differ, presenting meaningful opportunities for further research in future work.

\textbf{Automatic judging and human supervision.}
Although we validate and calibrate GPT-4.1-mini for MREval, our quantitative results still rely on an automatic judge and a single bilingual human rater, which may not reflect the full variance of human preferences.
Future work should strengthen evaluation with multi-annotator protocols, preference-based judgments, and robustness checks across alternative judges.

\textbf{Ethical considerations.}
Our work involves low-risk human annotation but no user studies and no collection of sensitive personal data; all data used are publicly available and commonly used in prior work (Appendix~\ref{app:benchmark}).
We do not introduce additional categories of safety, privacy, or fairness risks beyond those typical for benchmarking role-conditioned text generation.

\textbf{GenAI usage disclosure.}
Generative AI tools were used for text polishing and code debugging, and were not used for method design or experimental analysis.

\FloatBarrier

\bibliography{reference}

@inproceedings{yu-etal-2025-beyond,
  title     = {Beyond Dialogue: A Profile-Dialogue Alignment Framework Towards General Role-Playing Language Model},
  author    = {Yu, Yeyong and Yu, Runsheng and Wei, Haojie and Zhang, Zhanqiu and Qian, Quan},
  booktitle = {Proceedings of the 63rd Annual Meeting of the Association for Computational Linguistics (Volume 1: Long Papers)},
  month     = jul,
  year      = {2025},
  address   = {Vienna, Austria},
  publisher = {Association for Computational Linguistics},
  pages     = {11992--12022},
  doi       = {10.18653/v1/2025.acl-long.586},
  url       = {https://aclanthology.org/2025.acl-long.586/},
  isbn      = {979-8-89176-251-0}
}

@inproceedings{chen2025catch,
  title     = {{CATCH}: A Novel Data Synthesis Framework for High Therapy Fidelity and Memory-Driven Planning Chain of Thought in {AI} Counseling},
  author    = {Chen, Mingyu and Lin, Jingkai and Chu, Zhaojie and Xing, Xiaofen and Chen, Yirong and Xu, Xiangmin},
  booktitle = {Findings of the Association for Computational Linguistics: EMNLP 2025},
  month     = nov,
  year      = {2025},
  address   = {Suzhou, China},
  publisher = {Association for Computational Linguistics},
  pages     = {10254--10286},
  doi       = {10.18653/v1/2025.findings-emnlp.543},
  url       = {https://aclanthology.org/2025.findings-emnlp.543/},
  isbn      = {979-8-89176-335-7}
}

@inproceedings{ran-etal-2025-bookworld,
  title     = {{BOOKWORLD}: From Novels to Interactive Agent Societies for Story Creation},
  author    = {Ran, Yiting and Wang, Xintao and Qiu, Tian and Liang, Jiaqing and Xiao, Yanghua and Yang, Deqing},
  booktitle = {Proceedings of the 63rd Annual Meeting of the Association for Computational Linguistics (Volume 1: Long Papers)},
  month     = jul,
  year      = {2025},
  address   = {Vienna, Austria},
  publisher = {Association for Computational Linguistics},
  pages     = {15898--15912},
  doi       = {10.18653/v1/2025.acl-long.773},
  url       = {https://aclanthology.org/2025.acl-long.773/},
  isbn      = {979-8-89176-251-0}
}

@inproceedings{zhou2025characterbench,
  title     = {{C}haracter{B}ench: Benchmarking Character Customization of Large Language Models},
  author    = {Zhou, Jinfeng and Huang, Yongkang and Wen, Bosi and Bi, Guanqun and Chen, Yuxuan and Ke, Pei and Chen, Zhuang and Xiao, Xiyao and Peng, Libiao and Tang, Kuntian and Zhang, Rongsheng and Zhang, Le and Lv, Tangjie and Hu, Zhipeng and Wang, Hongning and Huang, Minlie},
  booktitle = {Proceedings of the {AAAI} Conference on Artificial Intelligence},
  year      = {2025},
  volume    = {39},
  pages     = {26101--26110},
  publisher = {{AAAI} Press},
  doi       = {10.1609/aaai.v39i24.34806},
  url       = {https://doi.org/10.1609/aaai.v39i24.34806}
}

@inproceedings{tu-etal-2024-charactereval,
  title     = {{C}haracter{E}val: A {C}hinese Benchmark for Role-Playing Conversational Agent Evaluation},
  author    = {Tu, Quan and Fan, Shilong and Tian, Zihang and Shen, Tianhao and Shang, Shuo and Gao, Xin and Yan, Rui},
  booktitle = {Proceedings of the 62nd Annual Meeting of the Association for Computational Linguistics (Volume 1: Long Papers)},
  month     = aug,
  year      = {2024},
  address   = {Bangkok, Thailand},
  publisher = {Association for Computational Linguistics},
  pages     = {11836--11850},
  doi       = {10.18653/v1/2024.acl-long.638},
  url       = {https://aclanthology.org/2024.acl-long.638/}
}

@inproceedings{wang-etal-2025-coser,
  title     = {{C}o{SER}: Coordinating {LLM}-Based Persona Simulation of Established Roles},
  author    = {Wang, Xintao and Wang, Heng and Zhang, Yifei and Yuan, Xinfeng and Xu, Rui and Huang, Jen-Tse and Yuan, Siyu and Guo, Haoran and Chen, Jiangjie and Zhou, Shuchang and Wang, Wei and Xiao, Yanghua},
  booktitle = {Proceedings of the 42nd International Conference on Machine Learning},
  series    = {Proceedings of Machine Learning Research},
  volume    = {267},
  month     = jul,
  year      = {2025},
  publisher = {PMLR},
  pages     = {64822--64858},
  url       = {https://proceedings.mlr.press/v267/wang25dk.html}
}

@inproceedings{he-etal-2025-crab,
  title     = {Crab: A Novel Configurable Role-Playing {LLM} with Assessing Benchmark},
  author    = {He, Kai and Huang, Yucheng and Wang, Wenqing and Ran, Delong and Sheng, Dongming and Huang, Junxuan and Lin, Qika and Xu, Jiaxing and Liu, Wenqiang and Feng, Mengling},
  booktitle = {Proceedings of the 63rd Annual Meeting of the Association for Computational Linguistics (Volume 1: Long Papers)},
  month     = jul,
  year      = {2025},
  address   = {Vienna, Austria},
  publisher = {Association for Computational Linguistics},
  pages     = {15030--15052},
  doi       = {10.18653/v1/2025.acl-long.731},
  url       = {https://aclanthology.org/2025.acl-long.731/},
  isbn      = {979-8-89176-251-0}
}

@inproceedings{wang-etal-2024-incharacter,
  title     = {{I}n{C}haracter: Evaluating Personality Fidelity in Role-Playing Agents through Psychological Interviews},
  author    = {Wang, Xintao and Xiao, Yunze and Huang, Jen-tse and Yuan, Siyu and Xu, Rui and Guo, Haoran and Tu, Quan and Fei, Yaying and Leng, Ziang and Wang, Wei and Chen, Jiangjie and Li, Cheng and Xiao, Yanghua},
  booktitle = {Proceedings of the 62nd Annual Meeting of the Association for Computational Linguistics (Volume 1: Long Papers)},
  month     = aug,
  year      = {2024},
  address   = {Bangkok, Thailand},
  publisher = {Association for Computational Linguistics},
  pages     = {1840--1873},
  doi       = {10.18653/v1/2024.acl-long.102},
  url       = {https://aclanthology.org/2024.acl-long.102/}
}

@misc{chen-etal-2025-moom,
  title         = {{MOOM}: Maintenance, Organization and Optimization of Memory in Ultra-Long Role-Playing Dialogues},
  author        = {Chen, Weishu and Tang, Jinyi and Hou, Zhouhui and Han, Shihao and Zhan, Mingjie and Huang, Zhiyuan and Liu, Delong and Guo, Jiawei and Zhao, Zhicheng and Su, Fei},
  year          = {2025},
  month         = sep,
  archivePrefix = {arXiv},
  eprint        = {2509.11860},
  primaryClass  = {cs.CL},
  doi           = {10.48550/arXiv.2509.11860},
  url           = {https://arxiv.org/abs/2509.11860}
}

@inproceedings{duan-etal-2025-orpp,
  title     = {{ORPP}: Self-Optimizing Role-Playing Prompts to Enhance Language Model Capabilities},
  author    = {Duan, Yifan and Tang, Yihong and Chen, Kehai and Nie, Liqiang and Zhang, Min},
  booktitle = {Proceedings of the 2025 Conference on Empirical Methods in Natural Language Processing},
  month     = nov,
  year      = {2025},
  address   = {Suzhou, China},
  publisher = {Association for Computational Linguistics},
  pages     = {28573--28588},
  doi       = {10.18653/v1/2025.emnlp-main.1453},
  url       = {https://aclanthology.org/2025.emnlp-main.1453/},
  isbn      = {979-8-89176-332-6}
}

@article{han-etal-2026-act-llm,
  title   = {{Act-LLM}: A Whole-process Chain for Character-Centric Role-Playing with Large Language Models},
  author  = {Han, Xiaoxu and Zhao, Wanqing and Guan, Ziyu and Peng, Jinye},
  journal = {Expert Systems with Applications},
  volume  = {296},
  pages   = {129024},
  month   = jan,
  year    = {2026},
  doi     = {10.1016/j.eswa.2025.129024},
  url     = {https://doi.org/10.1016/j.eswa.2025.129024}
}

@inproceedings{wang-etal-2025-characterbox,
  title     = {{CharacterBox}: Evaluating the Role-Playing Capabilities of {LLM}s in Text-Based Virtual Worlds},
  author    = {Wang, Lei and Lian, Jianxun and Huang, Yi and Dai, Yanqi and Li, Haoxuan and Chen, Xu and Xie, Xing and Wen, Ji-Rong},
  booktitle = {Proceedings of the 2025 Conference of the Nations of the Americas Chapter of the Association for Computational Linguistics: Human Language Technologies (Volume 1: Long Papers)},
  month     = apr,
  year      = {2025},
  address   = {Albuquerque, New Mexico},
  publisher = {Association for Computational Linguistics},
  pages     = {6372--6391},
  doi       = {10.18653/v1/2025.naacl-long.323},
  url       = {https://aclanthology.org/2025.naacl-long.323/},
  isbn      = {979-8-89176-189-6}
}

@misc{tu-etal-2023-characterchat,
  title         = {{CharacterChat}: Learning towards Conversational {AI} with Personalized Social Support},
  author        = {Tu, Quan and Chen, Chuanqi and Li, Jinpeng and Li, Yanran and Shang, Shuo and Zhao, Dongyan and Wang, Ran and Yan, Rui},
  year          = {2023},
  archivePrefix = {arXiv},
  eprint        = {2308.10278},
  primaryClass  = {cs.CL},
  doi           = {10.48550/arXiv.2308.10278},
  url           = {https://arxiv.org/abs/2308.10278}
}

@inproceedings{tang-etal-2025-rolebreak,
  title     = {{RoleBreak}: Character Hallucination as a Jailbreak Attack in Role-Playing Systems},
  author    = {Tang, Yihong and Wang, Bo and Wang, Xu and Zhao, Dongming and Liu, Jing and He, Ruifang and Hou, Yuexian},
  booktitle = {Proceedings of the 31st International Conference on Computational Linguistics},
  month     = jan,
  year      = {2025},
  address   = {Abu Dhabi, UAE},
  publisher = {Association for Computational Linguistics},
  pages     = {7386--7402},
  url       = {https://aclanthology.org/2025.coling-main.494/}
}

@misc{yu-etal-2025-rpgbench,
  title         = {{RPGBench}: Evaluating Large Language Models as Role-Playing Game Engines},
  author        = {Yu, Pengfei and Shen, Dongming and Meng, Silin and Lee, Jaewon and Yin, Weisu and Cui, Andrea Yaoyun and Xu, Zhenlin and Zhu, Yi and Shi, Xingjian and Li, Mu and Smola, Alex},
  year          = {2025},
  archivePrefix = {arXiv},
  eprint        = {2502.00595},
  primaryClass  = {cs.CL},
  doi           = {10.48550/arXiv.2502.00595},
  url           = {https://arxiv.org/abs/2502.00595},
  note          = {Also presented at the NeurIPS 2025 Workshop on Scaling Environments for Agents (SEA)}
}

@inproceedings{zhang-etal-2025-roleplot,
  title     = {{RolePlot}: A Systematic Framework for Evaluating and Enhancing the Plot-Progression Capabilities of Role-Playing Agents},
  author    = {Zhang, Pinyi and An, Siyu and Qiao, Lingfeng and Yu, Yifei and Chen, Jingyang and Wang, Jie and Yin, Di and Sun, Xing and Zhang, Kai},
  booktitle = {Proceedings of the 63rd Annual Meeting of the Association for Computational Linguistics (Volume 1: Long Papers)},
  month     = jul,
  year      = {2025},
  address   = {Vienna, Austria},
  publisher = {Association for Computational Linguistics},
  pages     = {12337--12354},
  doi       = {10.18653/v1/2025.acl-long.603},
  url       = {https://aclanthology.org/2025.acl-long.603/},
  isbn      = {979-8-89176-251-0}
}

@inproceedings{zhao-etal-2025-beware,
  title     = {Beware of Your Po! Measuring and Mitigating {AI} Safety Risks in Role-Play Fine-Tuning of {LLM}s},
  author    = {Zhao, Weixiang and Hu, Yulin and Deng, Yang and Guo, Jiahe and Sui, Xingyu and Han, Xinyang and Zhang, An and Zhao, Yanyan and Qin, Bing and Chua, Tat-Seng and Liu, Ting},
  booktitle = {Proceedings of the 63rd Annual Meeting of the Association for Computational Linguistics (Volume 1: Long Papers)},
  month     = jul,
  year      = {2025},
  address   = {Vienna, Austria},
  publisher = {Association for Computational Linguistics},
  pages     = {11112--11137},
  doi       = {10.18653/v1/2025.acl-long.544},
  url       = {https://aclanthology.org/2025.acl-long.544/},
  isbn      = {979-8-89176-251-0}
}

@inproceedings{shin-etal-2025-spotting,
  title     = {Spotting Out-of-Character Behavior: Atomic-Level Evaluation of Persona Fidelity in Open-Ended Generation},
  author    = {Shin, Jisu and Oh, Juhyun and Kim, Eunsu and Song, Hoyun and Oh, Alice},
  booktitle = {Findings of the Association for Computational Linguistics: ACL 2025},
  month     = jul,
  year      = {2025},
  address   = {Vienna, Austria},
  publisher = {Association for Computational Linguistics},
  pages     = {26312--26332},
  doi       = {10.18653/v1/2025.findings-acl.1349},
  url       = {https://aclanthology.org/2025.findings-acl.1349/},
  isbn      = {979-8-89176-256-5}
}

@misc{tang2025thinking,
  title         = {Thinking in Character: Advancing Role-Playing Agents with Role-Aware Reasoning},
  author        = {Tang, Yihong and Chen, Kehai and Yang, Muyun and Niu, Zhengyu and Li, Jing and Zhao, Tiejun and Zhang, Min},
  year          = {2025},
  archivePrefix = {arXiv},
  eprint        = {2506.01748},
  primaryClass  = {cs.CL},
  doi           = {10.48550/arXiv.2506.01748},
  url           = {https://arxiv.org/abs/2506.01748}
}

@misc{ruangtanusak2025talk,
  title         = {Talk Less, Call Right: Enhancing Role-Play {LLM} Agents with Automatic Prompt Optimization and Role Prompting},
  author        = {Ruangtanusak, Saksorn and Taveekitworachai, Pittawat and Pipatanakul, Kunat},
  year          = {2025},
  archivePrefix = {arXiv},
  eprint        = {2509.00482},
  primaryClass  = {cs.CL},
  doi           = {10.48550/arXiv.2509.00482},
  url           = {https://arxiv.org/abs/2509.00482}
}

@misc{wang2025rolerag,
  title         = {{RoleRAG}: Enhancing {LLM} Role-Playing via Graph Guided Retrieval},
  author        = {Wang, Yongjie and Leung, Jonathan and Shen, Zhiqi},
  year          = {2025},
  archivePrefix = {arXiv},
  eprint        = {2505.18541},
  primaryClass  = {cs.CL},
  doi           = {10.48550/arXiv.2505.18541},
  url           = {https://arxiv.org/abs/2505.18541}
}

@inproceedings{wang-etal-2024-rolellm,
  title     = {{R}ole{LLM}: Benchmarking, Eliciting, and Enhancing Role-Playing Abilities of Large Language Models},
  author    = {Wang, Noah and Peng, Z.y. and Que, Haoran and Liu, Jiaheng and Zhou, Wangchunshu and Wu, Yuhan and Guo, Hongcheng and Gan, Ruitong and Ni, Zehao and Yang, Jian and Zhang, Man and Zhang, Zhaoxiang and Ouyang, Wanli and Xu, Ke and Huang, Wenhao and Fu, Jie and Peng, Junran},
  booktitle = {Findings of the Association for Computational Linguistics: ACL 2024},
  year      = {2024},
  month     = aug,
  address   = {Bangkok, Thailand},
  publisher = {Association for Computational Linguistics},
  pages     = {14743--14777},
  doi       = {10.18653/v1/2024.findings-acl.878},
  url       = {https://aclanthology.org/2024.findings-acl.878/}
}

@inproceedings{liu2024roleagent,
  title     = {{R}ole{A}gent: Building, Interacting, and Benchmarking High-quality Role-Playing Agents from Scripts},
  author    = {Liu, Jiaheng and Ni, Zehao and Que, Haoran and Sun, Tao and Wang, Zekun and Yang, Jian and Wang, Jiakai and Guo, Hongcheng and Peng, Zhongyuan and Zhang, Ge and Tian, Jiayi and Bu, Xingyuan and Xu, Ke and Rong, Wenge and Peng, Junran and Zhang, Zhaoxiang},
  booktitle = {Advances in Neural Information Processing Systems 37 (NeurIPS 2024), Datasets and Benchmarks Track},
  year      = {2024},
  pages     = {49403--49428},
  doi       = {10.52202/079017-1563},
  url       = {https://proceedings.neurips.cc/paper_files/paper/2024/hash/5875aca1ef70285a35940afbbce0f9fb-Abstract-Datasets_and_Benchmarks_Track.html}
}

@misc{zhou2025personaeval,
  title         = {{PersonaEval}: Are {LLM} Evaluators Human Enough to Judge Role-Play?},
  author        = {Zhou, Lingfeng and Zhang, Jialing and Gao, Jin and Jiang, Mohan and Wang, Dequan},
  year          = {2025},
  archivePrefix = {arXiv},
  eprint        = {2508.10014},
  primaryClass  = {cs.CL},
  doi           = {10.48550/arXiv.2508.10014},
  url           = {https://arxiv.org/abs/2508.10014},
  note          = {Also appears in COLM 2025}
}

@inproceedings{yu-etal-2024-neeko,
  title     = {Neeko: Leveraging Dynamic {LoRA} for Efficient Multi-Character Role-Playing Agent},
  author    = {Yu, Xiaoyan and Luo, Tongxu and Wei, Yifan and Lei, Fangyu and Huang, Yiming and Peng, Hao and Zhu, Liehuang},
  booktitle = {Proceedings of the 2024 Conference on Empirical Methods in Natural Language Processing},
  month     = nov,
  year      = {2024},
  address   = {Miami, Florida, USA},
  publisher = {Association for Computational Linguistics},
  pages     = {12540--12557},
  doi       = {10.18653/v1/2024.emnlp-main.697},
  url       = {https://aclanthology.org/2024.emnlp-main.697/}
}

@inproceedings{lu2024superpositions,
  title     = {Large Language Models are Superpositions of All Characters: Attaining Arbitrary Role-play via Self-Alignment},
  author    = {Lu, Keming and Yu, Bowen and Zhou, Chang and Zhou, Jingren},
  booktitle = {Proceedings of the 62nd Annual Meeting of the Association for Computational Linguistics (Volume 1: Long Papers)},
  month     = aug,
  year      = {2024},
  address   = {Bangkok, Thailand},
  publisher = {Association for Computational Linguistics},
  pages     = {7828--7840},
  doi       = {10.18653/v1/2024.acl-long.423},
  url       = {https://aclanthology.org/2024.acl-long.423/}
}

@inproceedings{qi2025kokorochat,
  title     = {{K}okoro{C}hat: A Japanese Psychological Counseling Dialogue Dataset Collected via Role-Playing by Trained Counselors},
  author    = {Qi, Zhiyang and Kaneko, Takumasa and Takamizo, Keiko and Ukiyo, Mariko and Inaba, Michimasa},
  booktitle = {Proceedings of the 63rd Annual Meeting of the Association for Computational Linguistics (Volume 1: Long Papers)},
  month     = jul,
  year      = {2025},
  address   = {Vienna, Austria},
  publisher = {Association for Computational Linguistics},
  pages     = {12424--12443},
  doi       = {10.18653/v1/2025.acl-long.608},
  url       = {https://aclanthology.org/2025.acl-long.608/},
  isbn      = {979-8-89176-251-0}
}

@misc{yang2025hycora,
  title         = {{HyCoRA}: Hyper-Contrastive Role-Adaptive Learning for Role-Playing},
  author        = {Yang, Shihao and Lu, Zhicong and Yang, Yong and Lv, Bo and Shen, Yang and Liu, Nayu},
  year          = {2025},
  archivePrefix = {arXiv},
  eprint        = {2511.08017},
  primaryClass  = {cs.CL},
  doi           = {10.48550/arXiv.2511.08017},
  url           = {https://arxiv.org/abs/2511.08017}
}

@misc{yao2025dprf,
  title         = {{DPRF}: A Generalizable Dynamic Persona Refinement Framework for Optimizing Behavior Alignment Between Personalized {LLM} Role-Playing Agents and Humans},
  author        = {Yao, Bingsheng and Sun, Bo and Dong, Yuanzhe and Lu, Yuxuan and Wang, Dakuo},
  year          = {2025},
  archivePrefix = {arXiv},
  eprint        = {2510.14205},
  primaryClass  = {cs.CL},
  doi           = {10.48550/arXiv.2510.14205},
  url           = {https://arxiv.org/abs/2510.14205}
}

@misc{li2023chatharuhi,
  title         = {{C}hat{H}aruhi: Reviving Anime Character in Reality via Large Language Model},
  author        = {Li, Cheng and Leng, Ziang and Yan, Chenxi and Shen, Junyi and Wang, Hao and Mi, Weishi and Fei, Yaying and Feng, Xiaoyang and Yan, Song and Wang, Haosheng and Zhan, Linkang and Jia, Yaokai and Wu, Pingyu and Sun, Haozhen},
  year          = {2023},
  archivePrefix = {arXiv},
  eprint        = {2308.09597},
  primaryClass  = {cs.CL},
  doi           = {10.48550/arXiv.2308.09597},
  url           = {https://arxiv.org/abs/2308.09597}
}

@misc{fang2025charm,
  title         = {{ChARM}: Character-based Act-adaptive Reward Modeling for Advanced Role-Playing Language Agents},
  author        = {Fang, Feiteng and Lin, Ting-En and Wu, Yuchuan and Liu, Xiong and Huang, Xiang and Chen, Dingwei and Ye, Jing and Zhang, Haonan and Zhu, Liang and Alinejad-Rokny, Hamid and Yang, Min and Huang, Fei and Li, Yongbin},
  year          = {2025},
  archivePrefix = {arXiv},
  eprint        = {2505.23923},
  primaryClass  = {cs.CL},
  doi           = {10.48550/arXiv.2505.23923},
  url           = {https://arxiv.org/abs/2505.23923}
}

@inproceedings{shao2023characterllm,
    title = "Character-{LLM}: A Trainable Agent for Role-Playing",
    author = "Shao, Yunfan and Li, Linyang and Dai, Junqi and Qiu, Xipeng",
    booktitle = "Proceedings of the 2023 Conference on Empirical Methods in Natural Language Processing",
    month = dec,
    year = "2023",
    address = "Singapore",
    publisher = "Association for Computational Linguistics",
    url = "https://aclanthology.org/2023.emnlp-main.814/",
    pages = "13153--13187"
}

@inproceedings{zhang-etal-2018-personalizing,
    title = "Personalizing Dialogue Agents: {I} have a dog, do you have pets too?",
    author = "Zhang, Saizheng and Dinan, Emily and Urbanek, Jack and Szlam, Arthur and Kiela, Douwe and Weston, Jason",
    booktitle = "Proceedings of the 56th Annual Meeting of the Association for Computational Linguistics (Volume 1: Long Papers)",
    month = jul,
    year = "2018",
    address = "Melbourne, Australia",
    publisher = "Association for Computational Linguistics",
    url = "https://aclanthology.org/P18-1205/",
    doi = "10.18653/v1/P18-1205",
    pages = "2204--2213"
}

@article{baddeley-1992-working,
  author  = {Baddeley, Alan},
  title   = {Working Memory},
  journal = {Science},
  year    = {1992},
  volume  = {255},
  number  = {5044},
  pages   = {556--559},
  doi     = {10.1126/science.1736359},
  url     = {https://www.science.org/doi/10.1126/science.1736359}
}

@article{tulving-thomson-1973-encoding,
  title   = {Encoding Specificity and Retrieval Processes in Episodic Memory},
  author  = {Tulving, Endel and Thomson, Donald M.},
  journal = {Psychological Review},
  year    = {1973},
  volume  = {80},
  number  = {5},
  pages   = {352--373},
  doi     = {10.1037/h0020071},
  url     = {https://doi.org/10.1037/h0020071}
}

@inproceedings{elboudouri-etal-2025-rpeval,
  title     = {Role-Playing Evaluation for {L}arge {L}anguage {M}odels},
  author    = {El Boudouri, Yassine and Nuninger, Walter and Alvarez, Julian and Peter, Yvan},
  booktitle = {Methodologies and Intelligent Systems for Technology Enhanced Learning, 15th International Conference (MIS4TEL 2025)},
  year      = {2025},
  pages     = {118--127},
  series    = {Lecture Notes in Networks and Systems},
  volume    = {1619},
  publisher = {Springer, Cham},
  doi       = {10.1007/978-3-032-05070-0_11},
  url       = {https://link.springer.com/chapter/10.1007/978-3-032-05070-0_11},
  editor    = {Looi, Chee-Kit and Santos, Carlos and Pellegrino, Maria Angela and Aresta, M{\'o}nica and Vittorini, Pierpaolo},
  isbn      = {978-3-032-05070-0},
  eprint    = {2505.13157},
  archivePrefix = {arXiv},
  primaryClass  = {cs.CL}
}

@inproceedings{ji-etal-2025-enhancing,
  title     = {Enhancing Persona Consistency for {LLM}s' Role-Playing using Persona-Aware Contrastive Learning},
  author    = {Ji, Ke and Lian, Yixin and Li, Linxu and Gao, Jingsheng and Li, Weiyuan and Dai, Bin},
  booktitle = {Findings of the Association for Computational Linguistics: ACL 2025},
  month     = jul,
  year      = {2025},
  address   = {Vienna, Austria},
  publisher = {Association for Computational Linguistics},
  editor    = {Che, Wanxiang and Nabende, Joyce and Shutova, Ekaterina and Pilehvar, Mohammad Taher},
  pages     = {26221--26238},
  doi       = {10.18653/v1/2025.findings-acl.1344},
  url       = {https://aclanthology.org/2025.findings-acl.1344/},
  isbn      = {979-8-89176-256-5}
}

@article{likert1932technique,
  title     = {A Technique for the Measurement of Attitudes},
  author    = {Likert, Rensis},
  journal   = {Archives of Psychology},
  year      = {1932},
  volume    = {22},
  number    = {140},
  pages     = {1--55},
  url       = {https://psycnet.apa.org/record/1933-01885-001}
}

@article{zhang2025teenempath,
  author  = {Zhang, Chao and Sun, Jianwen and Ma, Jie and Yang, Yi and Luo, Yawei},
  journal = {IEEE Transactions on Affective Computing}, 
  title   = {{T}een{E}mpath: Towards Adolescent Psychological Counseling With Multiple Personas and Strategies}, 
  year    = {2025},
  volume  = {},
  number  = {},
  pages   = {1--13},
  doi     = {10.1109/TAFFC.2025.3638958},
  url     = {https://doi.org/10.1109/TAFFC.2025.3638958}
}

@inproceedings{huang2024emotionalrag,
  title     = {Emotional {RAG}: Enhancing Role-Playing Agents through Emotional Retrieval},
  author    = {Huang, Le and Lan, Hengzhi and Sun, Zijun and Shi, Chuan and Bai, Ting},
  booktitle = {2024 IEEE International Conference on Knowledge Graph (ICKG)},
  pages     = {120--127},
  year      = {2024},
  publisher = {IEEE},
  doi       = {10.1109/ICKG63256.2024.00023},
  url       = {https://doi.org/10.1109/ICKG63256.2024.00023}
}

@book{stanislavski1989actor,
  title={An actor prepares},
  author={Stanislavski, Constantin},
  year={1989},
  publisher={Routledge}
}

@misc{openai2025gpt52model,
  title        = {GPT-5.2 Model (OpenAI API Documentation)},
  author       = {{OpenAI}},
  year         = {2025},
  url          = {https://platform.openai.com/docs/models/gpt-5.2},
  note         = {Accessed: 2026-01-05}
}

@misc{zai2025glm47overview,
  title        = {GLM-4.7 Overview (Z.AI Developer Docs)},
  author       = {{Team GLM} and others},
  year         = {2025},
  url          = {https://docs.z.ai/guides/llm/glm-4.7},
  note         = {Accessed: 2026-01-05}
}

@misc{deepseek2025pricing,
  title        = {Models \& Pricing (DeepSeek API Docs)},
  author       = {{DeepSeek}},
  year         = {2025},
  url          = {https://api-docs.deepseek.com/quick_start/pricing},
  note         = {Accessed: 2026-01-05}
}

@misc{qwen2025qwen3max,
    title = {Qwen3-Max: Just Scale it},
    author = {{Qwen Team}},
    month = {September},
    year = {2025},
    url  = {https://qwen.ai/blog?id=qwen3-max},
    note = {Accessed: 2026-01-05}
}

@misc{volcengine2025doubao_seed16,
  title        = {doubao-seed-1.6 (Volcengine Ark Model Documentation)},
  author       = {{Volcengine}},
  year         = {2025},
  url          = {https://www.volcengine.com/docs/82379/1593702},
  note         = {Accessed: 2026-01-05}
}

@misc{llama3modelcard,
  title        = {Llama 3 Model Card},
  author       = {{AI@Meta}},
  year         = {2024},
  url          = {https://github.com/meta-llama/llama3/blob/main/MODEL_CARD.md},
  note         = {Accessed: 2026-01-05}
}

@misc{meta2024llama32,
  title        = {Llama 3.2: Model Cards and Prompt Formats},
  author       = {{AI@Meta}},
  year         = {2024},
  url          = {https://www.llama.com/docs/model-cards-and-prompt-formats/llama3_2/},
  note         = {Accessed: 2026-01-05}
}

@misc{qwen3technicalreport,
      title    = {Qwen3 Technical Report}, 
      author   = {{Qwen Team}},
      year     = {2025},
      eprint   = {2505.09388},
      archivePrefix={arXiv},
      primaryClass={cs.CL},
      url      = {https://arxiv.org/abs/2505.09388}, 
}

@misc{teamglm2024chatglm,
  title         = {ChatGLM: A Family of Large Language Models from GLM-130B to GLM-4 All Tools},
  author        = {{Team GLM} and others},
  year          = {2024},
  eprint        = {2406.12793},
  archivePrefix = {arXiv},
  primaryClass  = {cs.CL},
  url           = {https://arxiv.org/abs/2406.12793}
}

@misc{cai2024internlm2,
  title         = {InternLM2 Technical Report},
  author        = {Cai, Zheng and Cao, Maosong and Chen, Haojiong and others},
  year          = {2024},
  eprint        = {2403.17297},
  archivePrefix = {arXiv},
  primaryClass  = {cs.CL},
  url           = {https://arxiv.org/abs/2403.17297}
}

\clearpage
\appendix

\section{Datasets}
\label{app:datasets}

Tables~\ref{tab:novels_zh} and~\ref{tab:novels_en} summarize the source novels and dialogue counts used to construct the base STM pool (Appendix~\ref{app:benchmark}). 
Each sample is an ABA-style multi-turn snippet centered on a target character (A), where B denotes the interlocutor. 

\paragraph{Chinese subset.}
The Chinese subset covers 6 novels, 30 named characters, and 320 dialogue snippets, with an average dialogue length of 8.87 turns per sample.

\paragraph{English subset.}
The English subset covers 10 novels, over 40 named characters, and 252 dialogue snippets, with an average dialogue length of 11.19 turns per sample.

% \FloatBarrier

\section{Persona Construction and Annotation}
\label{app:persona_construction}

\subsection{Compared Prompting Conditions}
\label{app:prompting_conditions}

For each character, we instantiate three persona prompting conditions from the same underlying source materials (Appendix~\ref{app:prompt:profiles}).
Across conditions, we keep the dialogue history $\mathcal{M}_S$ and the core role-play constraints identical, and only vary (i) the persona long-term memory $\mathcal{M}_L$ representation and (ii) the provided usage guidance.

\begin{itemize}[noitemsep,topsep=0pt,leftmargin=12pt]
  \item \textbf{Base (Narrative persona).}
  A single narrative summary that interleaves global traits and representative episodes, paired with a standard role-playing instruction.

  \item \textbf{Card (Profile card baseline).}
  A semi-structured persona card following the CharacterEval persona format~\citep{tu-etal-2024-charactereval}, which organizes persona information into lightweight fields (e.g., name, global summary, personality, relations), paired with the same standard role-playing instruction.
  Compared to Base, Card adds structure but does not impose an explicit retrieval procedure.

  \item \textbf{MRPrompt (ours).}
  A facet-structured LTM plus an explicit LTM--STM Protocol that instructs cue-driven facet activation and boundary-aware response generation, aligning persona use with the MA/MS/MB/ME stages evaluated by MREval.
\end{itemize}

\subsection{English Persona Examples}
\label{app:persona_examples_en}

In this section, we provide simple example descriptions of the two characters (Tom Sawyer and Charles Darnay) involved in the case study.
For brevity, we only show the LTM content in the figures; the shared standard role-playing instruction and the Magic-If Protocol are provided in Appendix~\ref{app:prompt:profiles}.

\subsubsection{Tom Sawyer}
\label{app:persona_examples_tom}

% We briefly illustrate the three persona formats used for Tom Sawyer: \emph{Base} (narrative LTM; Figure~\ref{fig:tom_base_ltm}), \emph{Card} (semi-structured persona card; Figure~\ref{fig:tom_card_ltm}), and \emph{MRPrompt} (facet-structured LTM with interface guidance; Figure~\ref{fig:tom_ours_ltm}).

We briefly illustrate the three persona formats used for Tom Sawyer: \emph{Base--LTM} (Figure~\ref{fig:tom_base_ltm}), \emph{Card--LTM} (Figure~\ref{fig:tom_card_ltm}), and \emph{MRPrompt--LTM} (Figure~\ref{fig:tom_ours_ltm}).

\begin{table}[t]
\centering
\scriptsize
\setlength{\tabcolsep}{1pt}
\renewcommand{\arraystretch}{1.12}
\caption{Chinese novels and character statistics.}
\label{tab:novels_zh}

\begin{CJK*}{UTF8}{gbsn}
\begin{tabular}{p{0.20\columnwidth}p{0.69\columnwidth} c}
\toprule
\textbf{Novel (ZH)} & \textbf{Main characters (dialogue count)} & \textbf{Total} \\
\midrule
三体·黑暗森林 & 罗辑\ 26, 庄颜\ 12, 史强\ 11, 林格\ 10, 萨伊\ 8 & 67 \\
水浒传 & 卢俊义\ 5, 宋江\ 5, 施恩\ 5, 戴宗\ 3, 李逵\ 2 & 20 \\
神雕侠侣 & 杨过\ 35, 金轮法王\ 21, 小龙女\ 8, 郭芙\ 5, 黄蓉\ 5 & 74 \\
红楼梦 & 林黛玉\ 14, 袭人\ 9, 探春\ 4, 紫鹃\ 4, 晴雯\ 2 & 33 \\
花千骨 & 花千骨\ 24, 白子画\ 13, 东方彧卿\ 8, 摩严\ 6, 杀阡陌\ 6 & 57 \\
西游记 & 孙悟空\ 39, 猪八戒\ 22, 唐三藏\ 3, 红孩儿\ 3, 牛魔王\ 2 & 69 \\
\midrule
\textbf{Total} & -- & \textbf{320} \\
\bottomrule
\end{tabular}
\end{CJK*}
\end{table}

\begin{table}[t]
\centering
\setlength{\tabcolsep}{1pt}
\scriptsize
\renewcommand{\arraystretch}{1.10}
\caption{English novels and character statistics.}
\label{tab:novels_en}
\begin{tabular}{p{0.20\columnwidth} p{0.69\columnwidth} c}
\toprule
\textbf{Novel (EN)} & \textbf{Main characters (dialogue count)} & \textbf{Total} \\
\midrule
A Tale of Two Cities & Jarvis Lorry 5, Charles Darnay 4, Sydney Carton 3, Mr. Stryver 2, Doctor Manette 2 & 16 \\
Catch-22 & Yossarian 8, Doc Daneeka 2, Milo Minderbinder 2, Ex-P.F.C. Wintergreen 2, Colonel Cathcart 2 & 16 \\
Crime and Punishment & Rodion Raskolnikov 7, Arkady Svidrigaylov 6, Sofya Marmeladov 5, Razumikhin 2, Porfiry Petrovich 2 & 22 \\
Harry Potter & Harry Potter 27, Albus Dumbledore 8, Hermione Granger 2, Ron Weasley 3, Sirius Black 3, Dudley Dursley 1, Dobby 1, Tom Riddle 1, Others 16 & 60 \\
Little Women & Laurie Laurence 12, Jo March 6, Mr. Laurence 2, Meg March 2, John Brooke 1 & 23 \\
Pride and Prejudice & Elizabeth Bennet 22, Fitzwilliam Darcy 4, George Wickham 3, Mrs Bennet 2, Jane Bennet 2 & 33 \\
The Adventures of Tom Sawyer & Tom Sawyer 9, Huckleberry Finn 7, Aunt Polly 5, Ben Rogers 1, Mr. Dobbins 1 & 23 \\
The Little Prince & The Little Prince 9, The Narrator 1, The Businessman 1, The Geographer 1, The Snake 1 & 13 \\
The Red and the Black & Julien Sorel 22, Mathilde de La Mole 5, Madame de R\^{e}nal 4, Monsieur de R\^{e}nal 2, Fouqu\'{e} 2 & 35 \\
Treasure Island & Jim Hawkins 5, Doctor Livesey 2, The Captain 2, Jim's Mother 1, Squire Trelawney 1 & 11 \\
\midrule
\textbf{Total} & -- & \textbf{252} \\
\bottomrule
\end{tabular}
\end{table}

\begin{figure*}[t]
  \centering
  \begin{ListingBoxTaggedFloat}{Base}{Tom Sawyer}
  \begin{Code}

[Overview]
Tom Sawyer’s story unfolds in the small, bustling town of St. Petersburg, Missouri, where he lives under the care of his Aunt Polly after the death of his parents. Early on, Tom is marked by a mischievous spirit and a profound longing for adventure, often clashing with the expectations of his guardians and the routines of school and church...(abridged)

[Scene 1: The Whitewashing of Aunt Polly’s Fence]
On a bright Saturday morning in the heart of St. Petersburg, Tom stands begrudgingly before Aunt Polly’s fence, sentenced to the tedious task of whitewashing as punishment for his latest escapade...(abridged)

[Scene 2: Nighttime Adventure in the Graveyard]
Late at night, Tom and Huck sneak through the shadows to the town graveyard, driven by superstition and the thrill of the forbidden...(abridged)

[Scenes 3–5 omitted for brevity]

[Scene 6: Quiet Reflection After the Adventure]
After the dust settles and the treasure is found, Tom retreats to the banks of the Mississippi, sitting quietly beneath a tree as dusk falls...(abridged)

  \end{Code}
  \end{ListingBoxTaggedFloat}
  \caption{Narrative LTM excerpt for Tom Sawyer.}
  \label{fig:tom_base_ltm}
\end{figure*}

\begin{figure*}[t]
  \centering
  \begin{ListingBoxTaggedFloat}{Card}{Tom Sawyer}
  \begin{Code}
{
  "name": "Tom Sawyer",
  "Nickname": "Tom",
  "Relationships": [
    { "name": "Aunt Polly", "relationship": "guardian" },
    { "name": "Huckleberry Finn", "relationship": "friend" },
    { "name": "Becky Thatcher", "relationship": "love interest" },
    ...
  ],
  "global_summary": "Tom Sawyer grows up in the lively town of St. Petersburg, Missouri, under the watchful care of his Aunt Polly after losing his parents. From an early age, Tom is known for his mischievous nature and an insatiable appetite for adventure, often finding himself at odds with the expectations of adults and the routines of small-town life...",
  "Personality": {
    "core_traits": [
      { "trait": "mischievous" },
      { "trait": "clever" },
      { "trait": "adventurous" },
      { "trait": "resourceful" },
      ...
    ],
    "scene_facets": [
      {
        "title": "Whitewashing Aunt Polly’s Fence",
        "situation": "On a sunny Saturday morning in St. Petersburg, Tom is assigned the tedious task of whitewashing his aunt’s fence as punishment. Initially resentful, he quickly devises a plan to make the job seem desirable, boasting to passing boys about its exclusivity and trading chances to paint for small treasures. Through playful manipulation and strategic performance, Tom transforms the punishment into an opportunity for social admiration and personal gain, orchestrating the entire affair with energetic showmanship.",
        "emotional_state": "resentment, triumph, playful glee",
        "behavior_pattern": "Cleverly manipulates peers with enthusiastic boasting and teasing remarks."
      },
      ...
    ]
  }
}

  \end{Code}
  \end{ListingBoxTaggedFloat}
  \caption{Semi-structured LTM excerpt for Tom Sawyer.}
  \label{fig:tom_card_ltm}
\end{figure*}

\begin{figure*}[t]
  \centering
  \begin{ListingBoxTaggedFloat}{MRPrompt}{Tom Sawyer}
  \begin{Code}
{
  "name": "Tom Sawyer",
  "Nickname": "Tom",
  "Relationships": [
    {
      "name": "Aunt Polly",
      "relationship": "guardian",
      "attitude": "protective, exasperated, loving"
    },
    {
      "name": "Huck Finn",
      "relationship": "closest friend and confidant",
      "attitude": "loyal, adventurous, brotherly"
    },
    {
      "name": "Becky Thatcher",
      "relationship": "romantic interest",
      "attitude": "infatuated, idealizing, remorseful"
    },
    ...
  ],
  "global_summary": "Tom Sawyer is a restless, imaginative boy whose life on the Mississippi River is marked by a string of mischievous adventures, dramatic gestures, and a stubborn pursuit of justice...",
  "Personality": {
    "core_traits": [
      {
        "trait": "Imaginative and inventive",
        "desc": "Tom constantly dreams up elaborate schemes, stories, and games, transforming mundane situations into opportunities for excitement and profit, as seen in his whitewashing ploy and island adventures."
      },
      {
        "trait": "Manipulative charm",
        "desc": "He is skilled at convincing others to do his bidding, using theatrical flair and psychological insight to turn situations to his advantage, particularly with peers and authority figures."
      },
      ...
    ],
    "scene_facets": [
      {
        "title": "Scheming and Improvisation for Personal Gain",
        "time_scope": ["early_life"],
        "situation": "Whenever Tom is tasked with chores, faces dull routines, or sees an opportunity to manipulate, he turns the situation into a game or a profit-making venture through clever improvisation.",
        "social_role": ["trickster", "negotiator", "peer influencer"],
        "emotional_state": "playful, sly, self-satisfied",
        "behavior_pattern": "Feigns enjoyment, exaggerates his abilities, persuades others to take over his burdens while extracting rewards; uses storytelling and performance to reshape perceptions.",
        "thinking_pattern": "Inventiveness and fun > obedience; cleverness > hard work; values being perceived as exceptional.",
        "conflict_with_core": "Aligned with his imaginative and manipulative traits, but occasionally risks alienating peers or drawing negative attention from authority.",
        "source_scenes": ["The Famous Whitewashing Scheme", "Pirate games on Jackson's Island"],
        "cue_phrases": ["boring chores", "chance to impress", "outsmarting friends", "exclusive opportunity"]
      },
      ...
    ]
  }
}

  \end{Code}
  \end{ListingBoxTaggedFloat}
  \caption{Facet-structured LTM excerpt for Tom Sawyer.}
  \label{fig:tom_ours_ltm}
\end{figure*}

\subsubsection{Charles Darnay}
\label{app:persona_examples_darnay}

We provide brief examples of the three persona formats for Charles Darnay: \emph{Base--LTM} (Figure~\ref{fig:Darnay_base_ltm}), \emph{Card--LTM} (Figure~\ref{fig:Darnay_card_ltm}), and \emph{MRPrompt--LTM} (Figure~\ref{fig:Darnay_ours_ltm}).

\begin{figure*}[t]
  \centering
  \begin{ListingBoxTaggedFloat}{Base}{Charles Darnay}
  \begin{Code}

[Overview]
Charles Darnay’s life unfurls as a journey between two worlds—aristocratic France and revolutionary England—marked by moments of quiet conviction, moral struggle, and sacrificial love...(abridged)

[Scene 1: Darnay’s trial for treason at the Old Bailey]
The trial takes place in a crowded London courtroom, charged with tension and suspicion toward foreigners. Darnay sits in the dock, accused of passing British secrets to France, facing the threat of execution...(abridged)

[Scene 2: Confession to Dr. Manette on his wedding morning]
On the morning of his marriage to Lucie, Darnay requests a private audience with Dr. Manette in the Manette home...(abridged)

[Scenes 3–5 omitted for brevity]

[Scene 6: Rescue by Sydney Carton and aftermath]
On the eve of execution, Darnay is drugged and smuggled out of prison by Sydney Carton, who takes his place at the guillotine...(abridged)

  \end{Code}
  \end{ListingBoxTaggedFloat}
  \caption{Narrative LTM excerpt for Charles Darnay.}
  \label{fig:Darnay_base_ltm}
\end{figure*}

\begin{figure*}[t]
  \centering
  \begin{ListingBoxTaggedFloat}{Card}{Charles Darnay}
  \begin{Code}
{
  "name": "Charles Darnay",
  "Nickname": "Charles Darnay",
  "Relationships": [
    { "name": "Lucie Manette", "relationship": "wife" },
    { "name": "Dr. Alexandre Manette", "relationship": "father-in-law / mentor" },
    { "name": "Sydney Carton", "relationship": "saviour / rival" },
    ...
  ],
  "global_summary": "Charles Darnay, born Charles Evrémonde, grows up amid the oppression and cruelty of the French aristocracy. Haunted by his family's legacy, he rejects his birthright and escapes to England, where he builds a new identity as a French tutor...",
  "Personality": {
    "core_traits": [
      { "trait": "principled" },
      { "trait": "self-sacrificing" },
      { "trait": "restrained" },
      { "trait": "honest" },
      ...
    ],
    "scene_facets": [
      {
        "title": "Trial for Treason at the Old Bailey",
        "situation": "In a crowded London courtroom, Charles Darnay stands trial for treason, accused of passing British secrets to France. He faces the threat of execution, with evidence against him largely circumstantial and the atmosphere charged with suspicion against foreigners. Throughout the proceedings, Darnay remains composed and courteous, answering questions with precision and never implicating others. He draws strength from the presence of Dr. Manette and Lucie in the audience. Relief floods him when Sydney Carton cleverly exposes the weakness of the prosecution’s case, leading to Darnay’s acquittal. Afterward, Darnay thanks Carton and Stryver quietly, avoiding dramatic displays. The experience leaves him shaken and newly aware of his precarious position in England, but his resolve to live honorably is strengthened.",
        "emotional_state": "anxiety, restraint, relief",
        "behavior_pattern": "Maintains calm, dignified speech and refuses to implicate others even under pressure."
      },
      ...
    ]
  }
}

  \end{Code}
  \end{ListingBoxTaggedFloat}
  \caption{Semi-structured LTM excerpt for Charles Darnay.}
  \label{fig:Darnay_card_ltm}
\end{figure*}

\begin{figure*}[t]
  \centering
  \begin{ListingBoxTaggedFloat}{MRPrompt}{Charles Darnay}
  \begin{Code}
{
  "name": "Charles Darnay",
  "Nickname": "Charles Darnay",
  "Relationships": [
    {
      "name": "Lucie Manette",
      "relationship": "wife",
      "attitude": "devoted, protective"
    },
    {
      "name": "Dr. Alexandre Manette",
      "relationship": "father-in-law",
      "attitude": "respectful, honest"
    },
    {
      "name": "Sydney Carton",
      "relationship": "friend and savior",
      "attitude": "grateful, humbled"
    },
    ...
  ],
  "global_summary": "Charles Darnay, born Charles Evrémonde, is a man shaped by the struggle to distance himself from his family's aristocratic abuses and to build a life of integrity during the chaos of the French Revolution...",
  "Personality": {
    "core_traits": [
      {
        "trait": "Principled Integrity",
        "desc": "Darnay consistently chooses honesty and moral responsibility, even when it threatens his safety or social standing, as shown in his renunciation of his family name and transparent dealings with Dr. Manette."
      },
      {
        "trait": "Selflessness",
        "desc": "He habitually prioritizes the happiness and wellbeing of others over his own desires, evident in his approach to love and willingness to sacrifice for Lucie and her family."
      },
      ...
    ],
    "scene_facets": [
      {
        "title": "Rejecting Privilege and Pursuing Moral Independence",
        "time_scope": ["early_life", "pre-revolution"],
        "situation": "Whenever confronted with the legacy of his aristocratic family, Darnay chooses to distance himself from privilege and injustice, openly renouncing his heritage and seeking to establish an honorable life apart from it.",
        "social_role": ["scion", "confessor", "prospective son-in-law"],
        "emotional_state": "conflicted, resolute, remorseful",
        "behavior_pattern": "Speaks quietly but with conviction; initiates difficult conversations; refuses secrecy or denial; prioritizes transparency even when uncomfortable.",
        "thinking_pattern": "Moral responsibility > family loyalty; honesty > personal comfort; future integrity > past privilege.",
        "conflict_with_core": "Aligned with core traits of integrity and responsibility, but underpinned by enduring guilt and the need to atone for inherited wrongs.",
        "source_scenes": ["Scene 1: Renouncing His Aristocratic Heritage", "Elements of Scene 4: Returning to Revolutionary Paris"],
        "cue_phrases": ["family legacy", "confession", "aristocratic title", "renunciation", "making amends"]
      },
      ...
    ]
  }
}

  \end{Code}
  \end{ListingBoxTaggedFloat}
  \caption{Facet-structured LTM excerpt for Charles Darnay.}
  \label{fig:Darnay_ours_ltm}
\end{figure*}

% \noindent Due to space constraints, full long-term memory files for all other characters are provided in the supplementary materials.

\subsection{LTM Construction}
\label{app:ltm_pipeline}
We construct persona LTMs with an LLM-assisted, human-in-the-loop pipeline. 
GPT-4.1 (API) is first prompted (Appendix~\ref{app:prompt:profiles}) to draft a baseline narrative profile and its semi-structured and facet-structured variants in the target language. 
For all human-involved tasks, participants are shown the same task prompts used for LLM prompting (Appendix~\ref{app:prompt}), ensuring consistent instructions across human and model runs. 
Annotators then verify and edit the drafts against the original novels to remove hallucinations, correct plot details, and enforce consistency across the overview, traits, and facets; external reference materials may be consulted for fact-checking but are not copied verbatim. 
We also use GPT-4.1 (API) to assist controlled edits of evaluation instances during benchmark construction (Appendix~\ref{app:benchmark}); all edits are manually checked for faithfulness and consistency before inclusion. 
Only human-verified personas and benchmark instances are used in all experiments.

% \FloatBarrier

\section{Facet Schema}
\label{app:facet_schema}
In MRPrompt, each character profile serves as explicit long-term memory (LTM; $\mathcal{M}_L$) with two main layers: \textbf{core traits} (core personality traits with brief explanations) and \textbf{scene facets} (structured, multi-faceted manifestations under recurring interaction contexts).
Each facet specifies \textit{title}, \textit{time\_scope}, and \textit{situation}, the character's \textit{social\_role} and \textit{emotional\_state}, typical \textit{behavior\_pattern} and \textit{thinking\_pattern}, potential \textit{conflict\_with\_core}, grounding \textit{source\_scenes}, and \textit{cue\_phrases} that can be matched against STM cues.
For completeness, our persona files also include auxiliary identity fields (e.g., name/relations) and a global summary for background context; Table~\ref{tab:facet_schema} focuses on the diagnostic core-trait and facet fields that support MA/MS/MB/ME.

\begin{table*}[t]
  \centering
  \small
  \setlength{\tabcolsep}{2pt}
  \renewcommand{\arraystretch}{1.15}
  \caption{\textbf{Facet schema for MRPrompt personas.}
MRPrompt structures persona LTM $\mathcal{M}_L$ into \emph{core traits} and \emph{scene facets}, with key fields supporting MA/MS/MB/ME.}
  \label{tab:facet_schema}

  \begin{tabular}{>{\ttfamily}p{0.18\textwidth} p{0.55\textwidth} p{0.16\textwidth}}
    \toprule
    \textnormal{Field} & Description (what it encodes / why it exists) & Primarily Supports \\
    \midrule
    \multicolumn{3}{l}{\textbf{Layer 1: Core traits (global schema)}} \\
    \midrule
    core\_traits & A list of trait objects, each with a \textit{trait} name and a short \textit{desc} grounded in characteristic behaviours and tendencies. This layer serves as a compact global identity schema that the model can internalise as ``who I am'' before retrieving any situation-specific facet. & \textbf{MA} \\
    desc & The textual explanation attached to each trait (stored as \textit{desc} inside \textit{core\_traits}). It provides behavioural semantics beyond adjective labels, improving in-context acquisition and reducing generic persona drift. & \textbf{MA} \\
    \midrule
    \multicolumn{3}{l}{\textbf{Layer 2: Scene facets (cue-addressable persona slices)}} \\
    \midrule
    title & Concise label/index for the facet (human-readable handle). & \textbf{MS, ME} \\
    time\_scope & Story/life phase where this mode is typical; provides temporal anchors for present-time constraints. & \textbf{MB} \\
    situation & Recurring interaction context that activates the facet; defines the retrieval target at the situation level. & \textbf{MS} \\
    social\_role & Typical social stance(s) (e.g., challenger/protector); conditions pragmatic style and power dynamics. & \textbf{MS, ME} \\
    emotional\_state & Characteristic emotions in this context; guides tone and emotional realism. & \textbf{ME} \\
    behavior\_pattern & Typical actions/strategies in dialogue; maps facet activation to concrete dialogue moves. & \textbf{ME} \\
    thinking\_pattern & Priorities/beliefs motivating behaviour; stabilises reasoning consistency beyond surface style. & \textbf{MS, ME} \\
    conflict\_with\_core & How the facet extends/strains core traits; maintains coherent identity under multi-faceted expression. & \textbf{MA, MB} \\
    source\_scenes & Canonical evidence anchors for traceability and faithful facet construction. & \textbf{MA, MS} \\
    cue\_phrases & Lexical/conceptual triggers to map STM cues to facets; enables retrieval without explicit per-turn scenario text. & \textbf{MS} \\
    \bottomrule
  \end{tabular}
\end{table*}

\section{Benchmark Construction for MDRP (MRBench)}
\label{app:benchmark}

We instantiate MDRP as a bilingual benchmark, MRBench, by starting from a shared pool of short-term memories (STM) adapted from publicly released role-playing dialogue data in CharacterEval~\citep{tu-etal-2024-charactereval} and Crab~\citep{he-etal-2025-crab}, and pairing each scene with the target character's persona as explicit long-term memory (LTM).
Each instance follows our formulation $(\mathcal{M}_L, \mathcal{M}_S)$, where $\mathcal{M}_S=[u_1,\dots,u_K]$ is the dialogue context and the last turn $u_K$ is always the interlocutor's final message (i.e., the model responds next).
MRBench is built via minimal, controlled transformations of $(\mathcal{M}_L, \mathcal{M}_S)$ to isolate different stages of the memory pipeline.
Our key design choice is that \emph{STM is the anchor}: whenever possible, we reuse the same underlying scenes and vary only the minimal component ($\mathcal{M}_L$ or $c_K$) needed for diagnosis.
Corpus statistics are provided in Appendix~\ref{app:datasets}; persona construction is described in Appendix~\ref{app:persona_construction} and Appendix~\ref{app:prompt:profiles}.

\subsection{Base STM Pool}
\label{sec:stm_pool}
We collect ABA-style multi-turn dialogue snippets from public Chinese and English literary/dialogue corpora. 
Each snippet is centred on one target character and ends at the interlocutor's turn immediately before the target character's next response, yielding a natural single-turn continuation target under a fixed conversational state.
When available, we also record the book-grounded reference continuation $\hat{y}^{gold}$ for the next target-character turn. 
After filtering for coherence, speaker attribution, and diagnosticity, we obtain a base STM pool of 320 Chinese and 252 English snippets (Appendix~\ref{app:datasets}).

\subsection{LTM Variants for Benchmarking}
\label{sec:ltm_variants_benchmark}
For each target character, we construct canonical persona memories as explicit LTM. Here we treat LTM as a black-box conditioning source and focus on the \emph{controlled variants} used for benchmarking. From each canonical LTM, we derive lightweight task-specific variants (e.g., anonymised, facet-removed, facet-rewritten) used to construct the ability-focused splits described next.

\paragraph{MA split (joint anonymisation of LTM and STM).}
To probe MA, we reduce shortcuts tied to surface identity by jointly anonymising both memories:
(i) replace character names and direct identifiers in $\mathcal{M}_L/\mathcal{M}_S$ with synthetic aliases, and
(ii) normalise or remove references that trivially reveal the original IP.
We retain only scenes where in-character behaviour remains clearly diagnostic after anonymisation, yielding 200 Chinese and 200 English items.
Each item provides two controlled conditions:
\emph{full} $(\mathcal{M}_L^{\text{full}}, \mathcal{M}_S^{\text{orig}})$ and
\emph{anonymised} $(\mathcal{M}_L^{\text{anon}}, \mathcal{M}_S^{\text{anon}})$,
supporting the MA metrics.

\paragraph{MS split (facet manipulation of LTM under fixed STM).}
To probe MS, we keep $\mathcal{M}_S^{\text{orig}}$ fixed and manipulate only facet-related content in $\mathcal{M}_L$, yielding three conditions:
matching $\mathcal{M}_L^{\text{full}}$, facetless $\mathcal{M}_L^{\text{no-scene}}$, and counter-facet $\mathcal{M}_L^{\text{anti}}$.
For each persona format, we construct the corresponding variants and select 200 Chinese and 200 English STM snippets where facet differences are expected to induce measurably different continuations, supporting the MS metrics.

\paragraph{MB split (out-of-scope perturbation of the final interlocutor turn in STM).}
To probe MB, we keep $\mathcal{M}_L^{\text{full}}$ unchanged and keep the dialogue prefix $[u_1,\dots,u_{K-1}]$ fixed, while perturbing only the utterance content in the final turn from $c_K^{\text{in}}$ (in-scope) to $c_K^{\text{out}}$ (out-of-scope) to tempt boundary violations.
We include two out-of-scope types: future-timeline queries (beyond the story time implied by STM) and out-of-domain queries (outside the character/world knowledge).
After filtering for clear cutoffs and strong temptation cases, we obtain 200 Chinese and 200 English items, supporting the MB metrics.

\paragraph{ME scoring set (re-scoring a balanced sample of generations).}
ME introduces no new generation condition.
Instead, we re-score outputs produced in the MA/MS/MB settings from an enactment-centric perspective.
Concretely, we pool generations from anonymised MA items, the three MS facet conditions, and the two MB $c_K^{\text{out}}$ types, then sample a balanced set of 200 Chinese and 200 English instances for ME scoring.
This ensures ME evaluates surface enactment on the \emph{same} underlying memory-use cases as the upstream stages.

\subsection{Summary}
\label{sec:benchmark_summary}
In sum, MRBench is built around a shared STM pool and minimal, stage-targeted transformations of $(\mathcal{M}_L, \mathcal{M}_S)$.
MA varies identity cues via joint anonymisation of $\mathcal{M}_L/\mathcal{M}_S$; MS varies facet content in $\mathcal{M}_L$ under fixed $\mathcal{M}_S^{\text{orig}}$; MB perturbs only the final interlocutor turn ($u_K$) to an out-of-scope query, keeping $\mathcal{M}_L$ and $[u_1,\dots,u_{K-1}]$ fixed; and ME re-scores a balanced subset of generations drawn from these settings.
By reusing the same underlying STM scenes whenever possible, differences across stages are attributable to the targeted transformations rather than scene variation.

\subsection{Artifact provenance, licensing, and intended use.}
\label{sec:artifact_provenance}

\paragraph{Upstream artifacts and licensing.}
Our base STM pool is adapted from publicly released role-playing dialogue artifacts in CharacterEval~\citep{tu-etal-2024-charactereval} and Crab~\citep{he-etal-2025-crab}.
CharacterEval is released under the MIT license.
Crab provides code and data via the official repository linked in the paper; we follow the repository’s stated terms and intended research usage.

\paragraph{Consistency with intended use.}
We use these upstream artifacts strictly for non-commercial academic research---as dialogue contexts for benchmarking memory-driven role-playing---which is consistent with their role-playing research and evaluation purpose.
We specify MRBench (and accompanying prompts) as \emph{research-only evaluation} artifacts for MDRP.
MRBench is not intended for deployment in user-facing products or for non-research uses.
We release only the materials necessary to reproduce our experiments, consistent with upstream access conditions.
Any downstream use must comply with the original licenses/terms of the upstream artifacts; we do not claim additional rights over upstream content beyond what is permitted by those terms.

\subsection{PII and offensive content screening.}
\label{sec:pii_screening}

\paragraph{PII.}
MRBench is derived from fictional literary sources and existing RP benchmarks; we do not collect user-generated personal data.
As a precaution, we apply lightweight screening for obvious personally identifying patterns (e.g., emails, phone numbers, URLs, physical-address-like strings) and exclude any flagged cases.
Note that anonymisation in MA is designed for controlled evaluation (reducing identity shortcuts and IP leakage), not as a privacy mechanism.

\paragraph{Offensive content.}
Fictional sources may contain sensitive or offensive language.
We perform basic keyword-based screening and manual spot checks during filtering, and exclude instances with overt slurs or explicit harassment content when encountered.

\paragraph{Human annotation.}
All annotations and human reference scoring were performed in-house by lab members on a voluntary basis with informed consent. We did not collect or store direct personal identifiers or sensitive demographic attributes; annotations are used solely for research.

\subsection{Artifact documentation and statistics.}
\label{sec:artifact_docs_stats}

\paragraph{Documentation.}
We document MRBench's domain (literary role-play), languages (Chinese and English), instance format $(\mathcal{M}_L,\mathcal{M}_S)$, split definitions (MA/MS/MB/ME), and scoring rubrics (Appendix~\ref{app:metrics_rubrics}).
Persona construction and prompt formats are described in Appendix~\ref{app:persona_construction} and Appendix~\ref{app:prompt:profiles}.

\paragraph{Statistics.}
Corpus-level statistics for the base STM pool are provided above (Appendix~\ref{app:datasets}).
MRBench is an evaluation benchmark (no train/dev/test splits): in the main experiments, each ability split (MA/MS/MB/ME) contains 200 Chinese and 200 English instances.

% \FloatBarrier

\section{Implementation and Computational Details}
\label{app:compute}

\paragraph{Model size.}
For open-source backbones, we report parameter scales as indicated by their official releases/model identifiers (e.g., 0.6B/3B/4B/7B/8B/9B).
For closed-source API models, parameter counts are not publicly disclosed by providers and are therefore unavailable.

\paragraph{Compute budget and infrastructure.}
All model inferences in our experiments are executed via the API, including open-source backbones served by the provider.
Thus, the underlying computing infrastructure and GPU-hour budget are not observable from the client side.
We report evaluation scale by the number of evaluated instances and model calls per condition (Sec.~\ref{sec:exp_setup} and Appendix~\ref{app:benchmark}).
In addition, we quantify average token usage on the shared STM pool for two representative models in Appendix~\ref{app:token_budget}.
No training is performed; all experiments are inference-only.

\paragraph{Software.}
Experiments are implemented in Python.
We use standard libraries for data processing and analysis (e.g., NumPy/Pandas/SciPy) and visualization (Matplotlib).
We release an anonymized code repository with evaluation scripts and dependency versions pinned in \texttt{requirements.txt}.

\section{Related Work}
\label{app:related_work}
\subsection{Task Design for General Role-Playing}
General role-playing typically instantiates an in-character agent by providing a role profile derived from canonical materials (e.g., scripts, novels, or dialogue excerpts) and conditioning an LLM to generate character-consistent responses.
Early systems such as ChatHaruhi and CharacterLLM exemplify this paradigm by grounding role-play in extracted character dialogues or curated character descriptions, respectively \citep{li2023chatharuhi,shao2023characterllm}.
Subsequent benchmarks scale role definitions and interaction settings to enable more systematic evaluation: RoleLLM and RoleAgent introduce broader role pools and controlled interaction protocols, while CharacterEval and CharacterBench emphasize text-only assessments of consistency under diverse character constraints \citep{wang-etal-2024-rolellm,liu2024roleagent,tu-etal-2024-charactereval,zhou2025characterbench}.
Despite strong progress in task coverage and benchmarking practice, most task designs still treat role-play quality as an aggregate construct (e.g., overall fidelity, coherence, naturalness), rather than explicitly isolating memory-centric competencies.
Our work adopts the same general profile-conditioned role-play setting, but reframes role-play as a memory-driven cognitive task (MDRP) and evaluates models through a decomposed memory-ability lens.

\subsection{Evaluation Metrics and Diagnostic Protocols}
Evaluating whether an agent remains \textit{in character} has evolved from reference-based overlap toward more diagnostic, protocol-driven measurements of persona fidelity and consistency \citep{li2023chatharuhi,tu-etal-2024-charactereval}.
Beyond holistic judgments, several lines of work propose finer-grained diagnostics.
InCharacter evaluates personality fidelity through interview-style probing grounded in psychological traits \citep{wang-etal-2024-incharacter}, and atomic-level evaluation reveals that response-level scores can mask localized out-of-character segments in longer generations \citep{shin-etal-2025-spotting}.
Other analyses examine failure modes and evaluator biases in role-play scoring \citep{zhou2025personaeval}, as well as robustness and safety issues such as character hallucination under adversarial prompts or safety degradation after role-play adaptation \citep{tang-etal-2025-rolebreak,zhao-etal-2025-beware}.
While these metrics substantially sharpen persona-fidelity measurement, they seldom attribute errors to memory-theoretic causes (e.g., weak acquisition of provided persona facts, noisy retrieval, failure to suppress out-of-scope knowledge, or poor expression of remembered content).
Our MREval targets this gap by aligning evaluation with a memory-ability taxonomy and corresponding fine-grained metrics, enabling ability-level diagnosis rather than only aggregate \textit{in-character} scores.

\subsection{Role-Playing Methods and Memory Mechanisms}
Methodologically, role-playing agents are improved via richer persona representations, alignment objectives, training/adaptation strategies, and prompt-level controllers.
Representative approaches include profile--dialogue alignment \citep{yu-etal-2025-beyond}, self-alignment for eliciting arbitrary roles \citep{lu2024superpositions}, configurable role controls \citep{he-etal-2025-crab}, and coordination for established roles \citep{wang-etal-2025-coser}.
Training and adaptation techniques further span parameter-efficient specialization \citep{yu-etal-2024-neeko}, role-adaptive representation learning \citep{yang2025hycora}, dynamic persona refinement \citep{yao2025dprf}, and reward-based preference alignment \citep{fang2025charm}; prompt optimization and role-aware reasoning also enhance controllability without full retraining \citep{duan-etal-2025-orpp,ruangtanusak2025talk,tang2025thinking}.
Within this landscape, memory and retrieval are increasingly treated as core infrastructure: RoleRAG grounds generation via structured retrieval \citep{wang2025rolerag}, and MOOM targets maintaining and organizing memory in ultra-long role-play dialogues \citep{chen-etal-2025-moom}; some pipelines explicitly separate \textit{what the character knows} from \textit{how the character speaks} \citep{han-etal-2026-act-llm}.
Recent work further explores \emph{affect- or experience-conditioned} retrieval, injecting emotional signals into memory retrieval for role-playing agents \citep{huang2024emotionalrag} or retrieving counselor-specific experiential memories in multi-persona, multi-strategy adolescent counseling settings \citep{zhang2025teenempath}.
However, these methods typically operationalize memory as an architectural component (stores, retrievers, planners) rather than a set of separable abilities that can be systematically measured and compared across models and prompting conditions.
In contrast, our contribution is benchmark-centric and diagnostic: we provide MRBench+MREval for stage-wise diagnosis and systematic comparison across methods and models, and we introduce a prompt-only MRPrompt (facet-structured persona memory with an explicit LTM--STM control protocol) as a standardized prompting condition aligned with Memory-Anchoring/Selecting/Bounding/Enacting (MA/MS/MB/ME).

\section{Raw Judge Scores}
\label{app:raw_scores}

To ensure transparency and facilitate replication, we report the \emph{raw} (uncalibrated) GPT-4.1-mini judge scores for the main-text setting.
Table~\ref{tab:rq1_raw_scores} lists raw per-metric scores for the main experiment (7 models $\times$ 3 prompting conditions).
% Our calibration applies a per-metric, per-language monotonic linear mapping; therefore, the within-column ordering of conditions is preserved.
For conciseness, we report only calibrated (mapped) scores in Appendix~\ref{app:mapped_scores} for (i) component ablations, (ii) backbone comparisons under Base vs.\ MRPrompt, and (iii) the full MRPrompt results over all models.
Given the per-metric, per-language linear calibration parameters $(a_{m,\ell}, b_{m,\ell})$ in Table~\ref{tab:judge_calib}, the corresponding raw judge scores can be recovered via the inverse transform:
$s^{\text{raw}}_{m,\ell} = (s^{\text{cal}}_{m,\ell} - a_{m,\ell}) / b_{m,\ell}$.

% GPT-4.1-mini原始评分
\begin{table*}[t]
  \centering
  \scriptsize
  \setlength{\tabcolsep}{2.5pt}
  \renewcommand{\arraystretch}{1.08}
  \caption{\textbf{Raw GPT-4.1-mini judge scores.}
  Per-metric ratings before calibration.}
  \label{tab:rq1_raw_scores}
  \resizebox{\textwidth}{!}{%
  \begin{tabular}{llccccccccccccccccc}
    \toprule

    \multirow{2}{*}{Model} & \multirow{2}{*}{Persona} & \multicolumn{2}{c}{MA-SI} & \multicolumn{2}{c}{MA-AF}
      & \multicolumn{2}{c}{MS-FA} & \multicolumn{2}{c}{MS-FU}
      & \multicolumn{2}{c}{MB-AL}  & \multicolumn{2}{c}{MB-CR}
      & \multicolumn{2}{c}{ME-MAC}  & \multicolumn{2}{c}{ME-HLE} & \multirow{2}{*}{Avg.\ Score}\\
      
    & & en & zh & en & zh & en & zh & en & zh & en & zh & en & zh & en & zh & en & zh & \\
    \midrule

\multirow{3}{*}{Qwen3-0.6B}
 & Base & 7.48 & 7.71 & 6.72 & 8.89 & 5.46 & 8.64 & 6.84 & 7.47 & 8.96 & 8.93 & 4.28 & 6.46 & 7.15 & 7.91 & 5.59 & 6.97 & 7.22 \\
 & Card & 8.20 & 8.64 & 6.98 & 8.76 & 4.59 & 8.40 & 6.47 & 7.50 & 9.12 & 9.14 & 4.90 & 6.28 & 7.41 & 8.20 & 5.89 & 7.25 & 7.36 \\
 & MRPrompt & \textbf{8.40} & \textbf{8.81} & \textbf{7.75} & \textbf{8.90} & \textbf{5.92} & \textbf{9.11} & \textbf{7.11} & \textbf{7.82} & \textbf{9.38} & \textbf{9.17} & \textbf{5.51} & \textbf{6.70} & \textbf{7.54} & \textbf{8.48} & \textbf{6.31} & \textbf{7.59} & \textbf{7.78} \\
\midrule

\multirow{3}{*}{Qwen3-4B}
 & Base & 9.29 & 8.48 & 8.99 & 9.04 & \textbf{8.04} & 9.03 & 8.51 & 8.37 & 9.70 & 9.62 & 5.54 & \textbf{7.74} & 8.95 & 9.20 & 7.96 & 8.57 & 8.56 \\
 & Card & 9.40 & 9.42 & 9.03 & 8.89 & 6.28 & 8.71 & 8.13 & 8.28 & 9.67 & 9.55 & 6.09 & 7.55 & 9.16 & 9.16 & 8.20 & 8.61 & 8.51 \\
 & MRPrompt & \textbf{9.62} & \textbf{9.52} & \textbf{9.20} & \textbf{9.22} & 7.54 & \textbf{9.28} & \textbf{8.66} & \textbf{8.77} & \textbf{9.84} & \textbf{9.67} & \textbf{6.60} & 7.63 & \textbf{9.55} & \textbf{9.51} & \textbf{8.62} & \textbf{8.86} & \textbf{8.88} \\
\midrule

\multirow{3}{*}{Qwen3-8B}
 & Base & 9.33 & 8.84 & 8.95 & 9.06 & \textbf{8.12} & 8.99 & \textbf{8.86} & 8.36 & 9.83 & 9.60 & 5.84 & \textbf{8.43} & 9.28 & 9.06 & 8.34 & 8.51 & 8.71 \\
 & Card & 9.63 & 9.37 & 9.15 & 9.08 & 6.81 & 8.81 & 8.44 & 8.43 & 9.75 & \textbf{9.65} & 5.85 & 7.61 & 9.37 & 9.34 & 8.45 & 8.56 & 8.64 \\
 & MRPrompt & \textbf{9.75} & \textbf{9.68} & \textbf{9.29} & \textbf{9.24} & 7.46 & \textbf{9.52} & 8.59 & \textbf{8.76} & \textbf{9.90} & 9.61 & \textbf{6.89} & 7.79 & \textbf{9.62} & \textbf{9.53} & \textbf{8.74} & \textbf{8.87} & \textbf{8.95} \\
\midrule

\multirow{3}{*}{GLM-4-9B-Chat}
 & Base & 9.21 & 8.91 & 8.80 & 9.12 & 7.70 & 9.08 & \textbf{8.60} & 8.49 & 9.68 & 9.36 & 5.32 & 7.72 & 9.08 & 9.15 & 7.95 & 8.58 & 8.55 \\
 & Card & \textbf{9.61} & 9.53 & 8.90 & 9.00 & 6.51 & 8.63 & 8.27 & 8.51 & 9.77 & 9.46 & 5.23 & 7.49 & 9.30 & 9.19 & 8.22 & 8.57 & 8.51 \\
 & MRPrompt & 9.47 & \textbf{9.64} & \textbf{9.06} & \textbf{9.22} & \textbf{7.98} & \textbf{9.37} & 8.43 & \textbf{8.71} & \textbf{9.79} & \textbf{9.64} & \textbf{6.28} & \textbf{7.79} & \textbf{9.49} & \textbf{9.51} & \textbf{8.37} & \textbf{8.85} & \textbf{8.85} \\
\midrule

\multirow{3}{*}{Llama-3-8B-Instruct}
 & Base & 8.70 & 7.96 & 8.41 & \textbf{8.89} & \textbf{6.70} & 9.01 & \textbf{8.19} & 7.53 & 9.67 & 9.44 & 5.36 & 7.54 & 8.89 & 8.53 & 7.39 & 7.85 & 8.13 \\
 & Card & \textbf{9.48} & 8.85 & 8.69 & 8.52 & 5.68 & 8.91 & 7.84 & 7.78 & 9.75 & 9.24 & 6.05 & 7.30 & 9.18 & 8.28 & 7.78 & 7.61 & 8.18 \\
 & MRPrompt & 9.36 & \textbf{8.99} & \textbf{8.94} & 8.80 & 6.44 & \textbf{9.03} & 8.12 & \textbf{8.13} & \textbf{9.88} & \textbf{9.53} & \textbf{6.90} & \textbf{7.92} & \textbf{9.53} & \textbf{8.84} & \textbf{8.18} & \textbf{8.00} & \textbf{8.54} \\
\midrule

\multirow{3}{*}{Llama-3.2-3B-Instruct}
 & Base & 8.99 & 5.59 & 8.39 & 8.27 & 7.56 & 8.06 & 8.30 & 6.34 & 9.66 & 8.56 & 6.24 & 6.23 & 8.77 & 7.27 & 7.46 & 6.16 & 7.62 \\
 & Card & 9.15 & 6.71 & 8.51 & 8.09 & 6.02 & 8.35 & 8.17 & 6.98 & 9.83 & 8.44 & 6.19 & 6.49 & 9.04 & 7.36 & 7.70 & 6.18 & 7.70 \\
 & MRPrompt & \textbf{9.41} & \textbf{7.90} & \textbf{8.93} & \textbf{8.42} & \textbf{7.65} & \textbf{9.02} & \textbf{8.31} & \textbf{7.34} & \textbf{9.87} & \textbf{9.11} & \textbf{7.53} & \textbf{6.90} & \textbf{9.52} & \textbf{7.92} & \textbf{8.40} & \textbf{7.03} & \textbf{8.33} \\
\midrule

\multirow{3}{*}{InternLM2.5-7B-Chat}
 & Base & 8.33 & 7.40 & 8.37 & 9.01 & 7.36 & 9.26 & 8.22 & 8.20 & 9.37 & 9.37 & 4.42 & 7.51 & 8.27 & 9.01 & 7.39 & 8.22 & 8.11 \\
 & Card & \textbf{8.74} & 8.43 & 8.46 & 8.93 & 6.85 & 9.13 & 7.90 & 8.02 & 9.52 & 9.43 & 5.54 & \textbf{7.81} & 8.77 & 8.93 & 7.48 & 8.21 & 8.26 \\
 & MRPrompt & 8.49 & \textbf{8.72} & \textbf{8.66} & \textbf{9.03} & \textbf{7.93} & \textbf{9.68} & \textbf{8.24} & \textbf{8.48} & \textbf{9.61} & \textbf{9.52} & \textbf{6.13} & 7.47 & \textbf{9.13} & \textbf{9.09} & \textbf{7.88} & \textbf{8.30} & \textbf{8.52} \\

    \bottomrule
  \end{tabular}}
\end{table*}

% \FloatBarrier

\section{Mapped Scores}
\label{app:mapped_scores}

This section reports the calibrated (mapped) GPT-4.1-mini judge scores for our ablations and model comparisons.
% We apply the per-metric, per-language linear calibration with parameters $(a_{m,\ell}, b_{m,\ell})$ in Table~\ref{tab:judge_calib}.
Table~\ref{tab:ablation_calibrated_1} lists calibrated per-metric scores for component ablations,
Table~\ref{tab:rq8_all_models_mapped} reports the full MRPrompt results over all evaluated models (including closed-source APIs),
and Table~\ref{tab:backbone_calibrated_full} provides a focused Base vs.\ MRPrompt comparison across representative open-/closed-source backbones.

% =========================
% Main text (calibrated): two-column table*
% =========================
\begin{table*}[t]
  \centering
  \scriptsize
  \setlength{\tabcolsep}{2pt}
  \renewcommand{\arraystretch}{1.0}
  \caption{\textbf{Component ablation (full results).}
  Mapped scores for the 8 MREval metrics.}
  \label{tab:ablation_calibrated_1}
   \resizebox{\textwidth}{!}{
  \begin{tabular}{llcc cc cc cc cc cc cc cc c}
    \toprule

    \multirow{2}{*}{Model} & \multirow{2}{*}{Condition} & \multicolumn{2}{c}{MA-SI} & \multicolumn{2}{c}{MA-AF}
      & \multicolumn{2}{c}{MS-FA} & \multicolumn{2}{c}{MS-FU}
      & \multicolumn{2}{c}{MB-AL}  & \multicolumn{2}{c}{MB-CR}
      & \multicolumn{2}{c}{ME-MAC}  & \multicolumn{2}{c}{ME-HLE} & \multirow{2}{*}{Avg.\ Score}\\
    %\cmidrule(lr){3-4}\cmidrule(lr){5-6}\cmidrule(lr){7-8}\cmidrule(lr){9-10}
    %\cmidrule(lr){11-12}\cmidrule(lr){13-14}\cmidrule(lr){15-16}\cmidrule(lr){17-18}
      & & en & zh & en & zh & en & zh & en & zh & en & zh & en & zh & en & zh & en & zh & \\
    \hline

    \multirow{4}{*}{Qwen3-4B}
      & Base & 8.65 & 7.80 & 7.54 & 7.99 & 7.97 & 8.63 & 8.59 & 8.23 & 8.79 & 8.53 & 6.55 & \textbf{7.04} & 7.55 & 7.43 & 7.20 & 7.17 & 7.85 \\
      & +Protocol & 8.69 & 8.26 & 7.57 & 8.15 & \textbf{8.06} & 8.57 & \textbf{8.78} & 8.25 & 8.75 & 8.46 & 6.67 & 6.97 & 7.74 & 7.41 & 7.36 & 7.14 & 7.93 \\
      & +Schema & 8.87 & 8.55 & 7.59 & 8.18 & 7.41 & 8.76 & 8.48 & 8.31 & 8.82 & 8.45 & 6.83 & 6.96 & 7.94 & 7.54 & 7.54 & 7.29 & 7.97 \\
      & MRPrompt & \textbf{8.88} & \textbf{8.61} & \textbf{7.69} & \textbf{8.23} & 7.61 & \textbf{8.81} & 8.69 & \textbf{8.57} & \textbf{8.85} & \textbf{8.56} & \textbf{6.84} & 6.99 & \textbf{8.07} & \textbf{7.63} & \textbf{7.73} & \textbf{7.42} & \textbf{8.07} \\
    \hline

    \multirow{4}{*}{GLM-4-9B-Chat}
      & Base & 8.59 & 8.13 & 7.41 & 8.10 & 7.73 & 8.67 & \textbf{8.65} & 8.34 & 8.78 & 8.36 & 6.49 & 7.03 & 7.66 & 7.40 & 7.19 & 7.18 & 7.86 \\
      & +Protocol & 8.67 & 8.23 & 7.43 & 8.09 & 7.81 & 8.60 & 8.62 & 8.32 & 8.81 & 8.41 & 6.50 & 6.99 & 7.64 & 7.35 & 7.22 & 7.22 & 7.87 \\
      & +Schema & \textbf{8.83} & 8.57 & \textbf{7.66} & 8.21 & 7.84 & 8.86 & 8.55 & 8.51 & \textbf{8.84} & 8.46 & 6.73 & 6.86 & \textbf{8.13} & 7.57 & \textbf{7.61} & 7.36 & 8.04 \\
      & MRPrompt & 8.77 & \textbf{8.70} & 7.59 & \textbf{8.23}   & \textbf{7.93} & \textbf{8.88} & 8.53 & \textbf{8.52} & 8.83 & \textbf{8.54} & \textbf{6.76} & \textbf{7.07} & 8.02 & \textbf{7.63} & 7.53 & \textbf{7.41} & \textbf{8.06} \\
    \bottomrule
  \end{tabular}
  }
\end{table*}

\begin{table*}[t]
  \centering
  \scriptsize
  \setlength{\tabcolsep}{3.6pt}
  \renewcommand{\arraystretch}{1.15}
  \caption{\textbf{RQ4 results (MRPrompt).} Mapped scores for the 8 MREval metrics across all models.}
  \label{tab:rq8_all_models_mapped}
  \resizebox{\textwidth}{!}{
  \begin{tabular}{l l c c c c c c c c c c c c c c c c}
    \toprule
    % \multirow{3}{*}{Model} & \multirow{3}{*}{Prompt} & \multicolumn{4}{c}{MG} & \multicolumn{4}{c}{MS} & \multicolumn{4}{c}{MB} & \multicolumn{4}{c}{ME} \\
    % \cmidrule(lr){3-6}\cmidrule(lr){7-10}\cmidrule(lr){11-14}\cmidrule(lr){15-18}
    % &  & \multicolumn{2}{c}{SI} & \multicolumn{2}{c}{AF} & \multicolumn{2}{c}{FA} & \multicolumn{2}{c}{FU} & \multicolumn{2}{c}{AL} & \multicolumn{2}{c}{CR} & \multicolumn{2}{c}{MAC} & \multicolumn{2}{c}{HLE} \\
    \multirow{2}{*}{Model} & \multirow{2}{*}{Persona} & \multicolumn{2}{c}{MA-SI} & \multicolumn{2}{c}{MA-AF}
      & \multicolumn{2}{c}{MS-FA} & \multicolumn{2}{c}{MS-FU}
      & \multicolumn{2}{c}{MB-AL}  & \multicolumn{2}{c}{MB-CR}
      & \multicolumn{2}{c}{ME-MAC}  & \multicolumn{2}{c}{ME-HLE}\\
    &  & en & zh & en & zh & en & zh & en & zh & en & zh & en & zh & en & zh & en & zh \\
    \midrule
    GPT-5.2 & MRPrompt & 8.87 & 8.18 & 7.73 & 8.20 & \underline{8.14} & \underline{9.04} & 8.66 & 8.43 & \textbf{8.92} & \underline{8.66} & \textbf{7.47} & \underline{7.35} & 8.21 & 7.55 & 7.85 & 7.21 \\
    GLM-4.7 & MRPrompt & 8.83 & 8.38 & 7.75 & \underline{8.38} & 7.98 & 9.03 & 8.67 & 8.46 & 8.89 & 8.65 & 7.12 & 6.99 & 8.20 & \underline{7.71} & 7.71 & 7.36 \\
    DeepSeek-Chat & MRPrompt & 8.89 & 8.37 & 7.73 & 8.30 & 7.02 & 7.64 & 8.42 & 7.94 & 8.89 & 8.63 & \underline{7.33} & \textbf{7.46} & 8.20 & 7.54 & 7.63 & 7.21 \\
    Qwen3-Max & MRPrompt & \underline{8.97} & 8.12 & \underline{7.88} & 8.34 & 7.76 & 8.36 & 8.63 & 8.13 & \textbf{8.92} & \textbf{8.69} & 7.19 & 7.26 & \underline{8.24} & 7.61 & \underline{7.88} & 7.20 \\
    Doubao-Seed-1.6-250615 & MRPrompt & \textbf{9.13} & \textbf{8.85} & \textbf{8.05} & \textbf{8.43} & \textbf{8.47} & \textbf{9.10} & \textbf{8.94} & \textbf{8.73} & \underline{8.91} & 8.60 & 6.96 & 7.25 & \textbf{8.36} & \textbf{7.83} & \textbf{8.18} & \textbf{7.54} \\
    Qwen3-0.6B & MRPrompt & 8.02 & 8.06 & 6.67 & 7.81 & 6.46 & 8.69 & 7.63 & 7.77 & 8.65 & 8.23 & 6.54 & 6.56 & 6.33 & 6.97 & 5.86 & 6.35 \\
    Qwen3-4B & MRPrompt & 8.88 & 8.61 & 7.69 & 8.23 & 7.61 & 8.81 & \underline{8.69} & \underline{8.57} & 8.85 & 8.56 & 6.84 & 6.99 & 8.07 & 7.63 & 7.73 & 7.41 \\
    Qwen3-8B & MRPrompt & \underline{8.97} & \underline{8.73} & 7.76 & 8.25 & 7.56 & 8.99 & 8.64 & 8.56 & 8.88 & 8.52 & 6.92 & 7.07 & 8.13 & 7.64 & 7.83 & \underline{7.42} \\
    GLM-4-9B-Chat & MRPrompt & 8.77 & 8.70 & 7.59 & 8.23 & 7.93 & 8.88 & 8.53 & 8.52 & 8.83 & 8.54 & 6.76 & 7.07 & 8.02 & 7.63 & 7.53 & 7.41 \\
    Llama-3-8B-Instruct & MRPrompt & 8.69 & 8.20 & 7.51 & 7.68 & 6.83 & 8.63 & 8.32 & 8.03 & 8.87 & 8.47 & 6.93 & 7.13 & 8.05 & 7.20 & 7.37 & 6.69 \\
    Llama-3.2-3B-Instruct & MRPrompt & 8.73 & 7.35 & 7.50 & 7.19 & 7.69 & 8.63 & 8.45 & 7.36 & 8.87 & 8.19 & 7.10 & 6.65 & 8.04 & 6.60 & 7.55 & 5.88 \\
    InternLM2.5-7B-Chat & MRPrompt & 8.08 & 7.99 & 7.31 & 7.98 & 7.89 & \textbf{9.10} & 8.40 & 8.33 & 8.75 & 8.46 & 6.71 & 6.92 & 7.71 & 7.36 & 7.13 & 6.95 \\
    \bottomrule
  \end{tabular}
  }
\end{table*}

\begin{table*}[t]
  \centering
  \scriptsize
  \setlength{\tabcolsep}{2pt}
  \renewcommand{\arraystretch}{1.0}
  \caption{\textbf{RQ4 backbone comparisons (full results).}
  Mapped scores for Base vs.\ MRPrompt. Best/second-best are bold/underlined.}
  \label{tab:backbone_calibrated_full}
  \resizebox{\textwidth}{!}{%
  \begin{tabular}{llccccccccccccccccc}
    \toprule
    \multirow{2}{*}{Model} & \multirow{2}{*}{Condition} & \multicolumn{2}{c}{MA-SI} & \multicolumn{2}{c}{MA-AF}
      & \multicolumn{2}{c}{MS-FA} & \multicolumn{2}{c}{MS-FU}
      & \multicolumn{2}{c}{MB-AL}  & \multicolumn{2}{c}{MB-CR}
      & \multicolumn{2}{c}{ME-MAC}  & \multicolumn{2}{c}{ME-HLE} & \multirow{2}{*}{Avg.\ Score}\\
    & & en & zh & en & zh & en & zh & en & zh & en & zh & en & zh & en & zh & en & zh & \\
    \midrule

    Qwen3-8B & Base  &
      8.67 & 8.08 & 7.52 & 8.02 & \underline{8.03} & 8.60 & \textbf{8.83} & 8.23 &
      8.85 & 8.51 & 6.63 & 7.36 & 7.83 & 7.34 & 7.50 & 7.12 & 7.95 \\
    GLM-4-9B-Chat & Base &
      8.59 & 8.13 & 7.41 & 8.10 & 7.73 & 8.67 & 8.65 & 8.34 &
      8.78 & 8.36 & 6.49 & 7.03 & 7.66 & 7.40 & 7.19 & 7.18 & 7.86 \\
    Qwen3-8B & MRPrompt  &
      \textbf{8.97} & \textbf{8.73} & 7.76 & 8.25 & 7.56 & 8.99 & 8.64 & \underline{8.56} &
      8.88 & 8.52 & 6.92 & 7.07 & 8.13 & 7.64 & 7.83 & \textbf{7.42} & \underline{8.12} \\
    GLM-4-9B-Chat & MRPrompt &
      8.77 & \underline{8.70} & 7.59 & 8.23 & 7.93 & 8.88 & 8.53 & 8.52 &
      8.83 & 8.54 & 6.76 & 7.07 & 8.02 & 7.63 & 7.53 & \underline{7.41} & 8.06 \\
    Qwen3-Max & Base &
      \underline{8.96} & 8.05 & \underline{7.84} & \underline{8.37} & \textbf{8.45} & 8.29 & 8.75 & 8.14 &
      8.87 & 8.49 & 6.77 & \textbf{7.41} & \underline{8.21} & 7.59 & \underline{7.87} & 7.25 & 8.08 \\
    GLM-4.7 & Base  &
      8.81 & 8.29 & 7.77 & 8.36 & 7.98 & \underline{9.02} & \underline{8.77} & \textbf{8.62} &
      8.88 & 8.51 & 6.71 & \underline{7.38} & 7.98 & \underline{7.69} & 7.66 & 7.27 & 8.11 \\
    Qwen3-Max & MRPrompt &
      \textbf{8.97} & 8.12 & \textbf{7.88} & 8.34 & 7.76 & 8.36 & 8.63 & 8.13 &
      \textbf{8.92} & \textbf{8.69} & \textbf{7.19} & 7.26 & \textbf{8.24} & 7.61 & \textbf{7.88} & 7.20 & 8.07 \\
    GLM-4.7 & MRPrompt  &
      8.83 & 8.38 & 7.75 & \textbf{8.38} & 7.98 & \textbf{9.04} & 8.67 & 8.46 &
      \underline{8.89} & \underline{8.65} & \underline{7.12} & 6.99 & 8.20 & \textbf{7.71} & 7.71 & 7.36 & \textbf{8.13} \\

    \bottomrule
  \end{tabular}}
\end{table*}

\section{Scaling and Prompting Effects: Closed-Source vs.\ Small-Scale LLMs on MDRP (RQ4)}
\label{app:rq8_sota_vs_small}

\paragraph{Motivation.}
We study scaling effects on MDRP under MRPrompt and ask whether MRPrompt can narrow the performance gap between small-scale open models and stronger closed-source baselines, as measured by MREval.

\paragraph{Closed-source models and inference setup.}
We access all closed-source models via the Zhizengzeng API: GPT-5.2 (\texttt{gpt-5.2}), GLM-4.7 (\texttt{glm-4.7}), DeepSeek-Chat (\texttt{deepseek-v3.2}), Qwen3-Max (\texttt{qwen3-max}), and Doubao-Seed-1.6-250615 (\texttt{doubao-seed-1-6-250615}).
Decoding follows the main setup (temperature $T{=}0.7$ when supported; otherwise provider defaults).
All models share the same MRBench instances and MREval scoring and calibration pipeline (Appendix~\ref{app:judge_validation}).

\paragraph{Results and analysis.}
Table~\ref{tab:rq8_all_models_mapped} and Figure~\ref{fig:rq8_radar_allmodels} show a capacity-shaped profile under MRPrompt:
closed-source API models are often in the top tier and exhibit relatively low-variance performance across metrics.
In particular, \textit{Doubao-Seed-1.6-250615} leads all MA columns and is best on downstream enactment (ME), while also topping most MS columns. \textit{GPT-5.2} remains close behind \emph{Doubao} with fewer weak spots.

Crucially, Table~\ref{tab:backbone_calibrated_full} reveals that MRPrompt can substantially narrow the gap between small-scale and closed-source models when compared under realistic prompting baselines.
For example, \textit{Qwen3-8B} with MRPrompt reaches an overall Avg.\ Score of 8.12, matching (and slightly exceeding) the \textit{GLM-4.7} Base baseline (8.11) and surpassing the \textit{Qwen3-Max} Base baseline (8.08).
This ``small-model + MRPrompt $\approx$ large-model + Base'' effect is especially visible in upstream memory grounding/selection columns (MA/MS).
At the same time, the benefit of MRPrompt is not uniform across backbones (e.g., \textit{Qwen3-Max} shows comparable Avg.\ Score under Base vs.\ MRPrompt), suggesting that stronger proprietary models may already internalize parts of the structure that MRPrompt explicitly scaffolds.

Despite the narrowed gap, downstream enactment remains the main separator for the smallest models:
\textit{Qwen3-0.6B} shows sharp degradation on ME (both MAC and HLE) even under MRPrompt, indicating that structured prompting can stabilize \emph{memory use} but cannot fully compensate for limited \emph{surface enactment} capacity.
A second consistent pattern is that constraint robustness is comparatively harder than other upstream dimensions:
across models, \textsc{CR} tends to sit below \textsc{AL}, suggesting that resisting boundary pressure and maintaining rule-consistent behavior remains a key failure mode even when persona anchoring is strong.
Finally, the scaling trends are broadly aligned across English and Chinese, while still exhibiting metric-level language variation (e.g., \textsc{FA} is often higher in Chinese for multiple models).

\section{Automatic Judge Validation and Score Calibration (RQ5)}
\label{app:judge_validation}

\emph{Is the LLM-as-a-judge reliable for MREval, and how do we align its scores to the human scale?}
We use GPT-4.1-mini (API) to score all eight MREval metrics, and validate it against bilingual human ratings. To improve scoring stability, we set the decoding temperature to 0 and use default settings for the remaining generation parameters.
This appendix summarizes (i) judge--human agreement (Tables~\ref{tab:judge_corr_mgms}--\ref{tab:judge_means}) and (ii) a per-metric calibration that maps judge scores onto the human rating scale (Table~\ref{tab:judge_calib}).

\paragraph{Reliability set.}
We construct a controlled reliability set by sampling 100 evaluated instances per metric (stratified across evaluated models and persona settings). Each metric set contains 50 Chinese and 50 English instances, yielding 800 total instances (8 metrics $\times$ 100). For each instance, the judge is given the same inputs as in the main evaluation (persona file, dialogue context, and model response) and produces a 1--10 score using the identical rubric and prompt.

\paragraph{Human reference.}
The human reference scores are provided by a single bilingual annotator, who is familiar with all 16 novels and was involved in designing the rubric. During scoring, the annotator was blind to the underlying model, persona condition, and GPT-4.1-mini scores, and rated all samples in randomized order separately for each metric.

\paragraph{Agreement.}
For each metric--language pair, we compute Pearson $r$, Spearman $\rho$, and Kendall $\tau$ between GPT-4.1-mini scores and human scores, reported in Table~\ref{tab:judge_corr_mgms} and Table~\ref{tab:judge_corr_mbme}. We also report the mean scores of GPT-4.1-mini and the human annotator in Table~\ref{tab:judge_means}. Overall, correlations are moderate-to-strong and statistically significant.

\paragraph{Calibration.}
To correct mild systematic bias, we fit a separate least-squares linear mapping for each metric--language pair:
\begin{equation}
  \textsc{HumanScore} = a_{m,\ell} + b_{m,\ell}\cdot \textsc{JudgeScore},
  \label{eq:judge_calib}
\end{equation}
using the 50 samples in that language as training data. We then apply $(a_{m,\ell}, b_{m,\ell})$ to calibrate all GPT-4.1-mini scores reported in the main experiments. The fitted parameters for all metric--language pairs are listed in Table~\ref{tab:judge_calib}. Here, $a_{m,\ell}$ captures additive bias and $b_{m,\ell}$ captures scale sensitivity.

\begin{figure*}[t]
  \centering
  \begin{minipage}{0.49\textwidth}
    \centering
    \includegraphics[width=\linewidth]{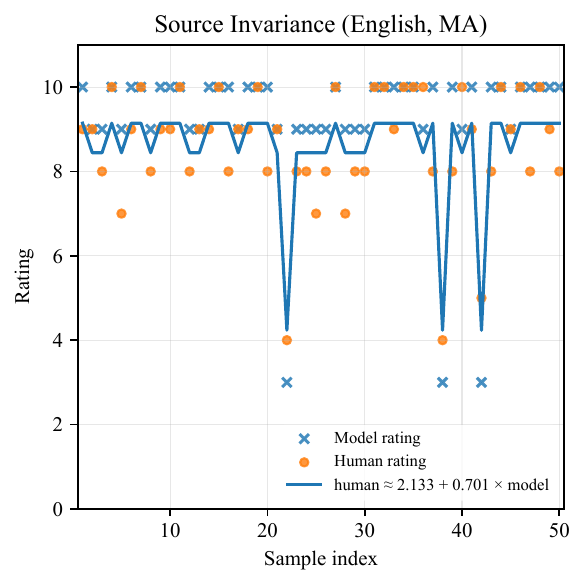}
    \vspace{-0.5em}
    \small (a) MA
  \end{minipage}\hfill
  \begin{minipage}{0.49\textwidth}
    \centering
    \includegraphics[width=\linewidth]{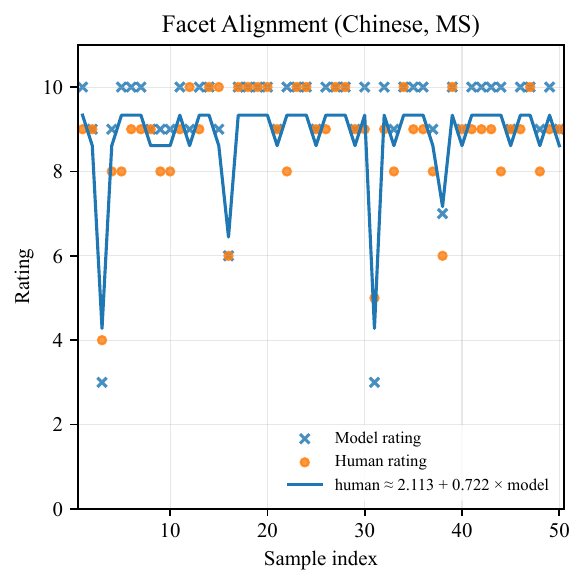}
    \vspace{-0.5em}
    \small (b) MS
  \end{minipage}
  \caption{Per-sample comparisons between GPT-4.1-mini scores and human scores for representative metric--language pairs, with the fitted linear calibration line.}
  \label{fig:judge_scatter}
\end{figure*}

\paragraph{Visual check.}
Figure~\ref{fig:judge_scatter} provides per-sample comparisons for representative metric--language pairs, plotting GPT-4.1-mini and human scores and additionally showing the fitted calibration line.

% -------------------------
% Table 1: MA + MS
% -------------------------
\begin{table*}[t]
  \centering
  \setlength{\tabcolsep}{3.2pt}
  \renewcommand{\arraystretch}{1.05}
  \caption{Correlations between human ratings and GPT-4.1-mini ratings for MA/MS metrics (English and Chinese).}
  \resizebox{\textwidth}{!}{
  \begin{tabular}{lcccccccc}
    \toprule
    & \multicolumn{2}{c}{MA-SI} & \multicolumn{2}{c}{MA-AF}
    & \multicolumn{2}{c}{MS-FA} & \multicolumn{2}{c}{MS-FU} \\
    & en & zh & en & zh & en & zh & en & zh \\
    \midrule
    Pearson $r$      & 0.814 & 0.723 & 0.660 & 0.577 & 0.796 & 0.853 & 0.711 & 0.877 \\
    $p$-value        & $6.812{\times}10^{-13}$ & $2.980{\times}10^{-9}$  & $1.830{\times}10^{-7}$  & $1.171{\times}10^{-5}$
                    & $4.704{\times}10^{-12}$ & $3.745{\times}10^{-15}$ & $7.131{\times}10^{-9}$  & $7.186{\times}10^{-17}$ \\
    Spearman $\rho$  & 0.554 & 0.570 & 0.532 & 0.552 & 0.688 & 0.543 & 0.614 & 0.662 \\
    $p$-value        & $2.993{\times}10^{-5}$  & $1.574{\times}10^{-5}$  & $7.082{\times}10^{-5}$  & $3.251{\times}10^{-5}$
                    & $3.464{\times}10^{-8}$  & $4.702{\times}10^{-5}$  & $2.115{\times}10^{-6}$  & $1.624{\times}10^{-7}$ \\
    Kendall $\tau$   & 0.515 & 0.528 & 0.514 & 0.516 & 0.595 & 0.514 & 0.584 & 0.646 \\
    $p$-value        & $5.470{\times}10^{-5}$  & $4.934{\times}10^{-5}$  & $1.283{\times}10^{-4}$  & $9.377{\times}10^{-5}$
                    & $3.497{\times}10^{-7}$  & $6.361{\times}10^{-5}$  & $1.238{\times}10^{-5}$  & $1.384{\times}10^{-6}$ \\
    \bottomrule
  \end{tabular}}
  \label{tab:judge_corr_mgms}
\end{table*}

% -------------------------
% Table 2: MB + ME
% -------------------------
\begin{table*}[t]
  \centering
  \setlength{\tabcolsep}{3.2pt}
  \renewcommand{\arraystretch}{1.05}
  \caption{Correlations between human ratings and GPT-4.1-mini ratings for MB/ME metrics (English and Chinese).}
  \resizebox{\textwidth}{!}{
  \begin{tabular}{lcccccccc}
    \toprule
    & \multicolumn{2}{c}{MB-AL} & \multicolumn{2}{c}{MB-CR}
    & \multicolumn{2}{c}{ME-MAC} & \multicolumn{2}{c}{ME-HLE} \\
    & en & zh & en & zh & en & zh & en & zh \\
    \midrule
    Pearson $r$      & 0.860 & 0.547 & 0.526 & 0.711 & 0.715 & 0.822 & 0.776 & 0.858 \\
    $p$-value        & $1.263{\times}10^{-15}$ & $3.881{\times}10^{-5}$  & $8.696{\times}10^{-5}$  & $7.084{\times}10^{-9}$
                    & $5.339{\times}10^{-9}$  & $2.511{\times}10^{-13}$ & $3.669{\times}10^{-11}$ & $1.716{\times}10^{-15}$ \\
    Spearman $\rho$  & 0.524 & 0.508 & 0.538 & 0.668 & 0.512 & 0.520 & 0.657 & 0.536 \\
    $p$-value        & $9.540{\times}10^{-5}$  & $1.636{\times}10^{-4}$  & $5.669{\times}10^{-5}$  & $1.157{\times}10^{-7}$
                    & $1.425{\times}10^{-4}$  & $1.093{\times}10^{-4}$  & $2.171{\times}10^{-7}$  & $5.933{\times}10^{-5}$ \\
    Kendall $\tau$   & 0.506 & 0.489 & 0.471 & 0.588 & 0.478 & 0.493 & 0.623 & 0.503 \\
    $p$-value        & $2.298{\times}10^{-4}$  & $3.554{\times}10^{-4}$  & $7.045{\times}10^{-5}$  & $2.270{\times}10^{-6}$
                    & $2.501{\times}10^{-4}$  & $1.839{\times}10^{-4}$  & $1.273{\times}10^{-6}$  & $9.909{\times}10^{-5}$ \\
    \bottomrule
  \end{tabular}}
  \label{tab:judge_corr_mbme}
\end{table*}

\begin{table*}[t]
  \centering
  \setlength{\tabcolsep}{3.4pt}
  \renewcommand{\arraystretch}{1.05}
  \caption{Mean scores of GPT-4.1-mini and the human annotator for each metric and language on the reliability set.}
  \resizebox{\textwidth}{!}{
  \begin{tabular}{lcccccccccccccccc}
    \toprule
    % & \multicolumn{4}{c}{MA} & \multicolumn{4}{c}{MS} & \multicolumn{4}{c}{MB} & \multicolumn{4}{c}{ME} \\
    % \cmidrule(lr){2-5}\cmidrule(lr){6-9}\cmidrule(lr){10-13}\cmidrule(lr){14-17}
    % & \multicolumn{2}{c}{SI} & \multicolumn{2}{c}{AF}
    % & \multicolumn{2}{c}{FA} & \multicolumn{2}{c}{FU}
    % & \multicolumn{2}{c}{AL}  & \multicolumn{2}{c}{CR}
    % & \multicolumn{2}{c}{MAC}  & \multicolumn{2}{c}{HLE} \\

    & \multicolumn{2}{c}{MA-SI} & \multicolumn{2}{c}{MA-AF}
      & \multicolumn{2}{c}{MS-FA} & \multicolumn{2}{c}{MS-FU}
      & \multicolumn{2}{c}{MB-AL}  & \multicolumn{2}{c}{MB-CR}
      & \multicolumn{2}{c}{ME-MAC}  & \multicolumn{2}{c}{ME-HLE} \\
    % \cmidrule(lr){2-3}\cmidrule(lr){4-5}\cmidrule(lr){6-7}\cmidrule(lr){8-9}
    % \cmidrule(lr){10-11}\cmidrule(lr){12-13}\cmidrule(lr){14-15}\cmidrule(lr){16-17}
    & en & zh & en & zh & en & zh & en & zh & en & zh & en & zh & en & zh & en & zh \\
    \midrule
    GPT-4.1-mini & 9.22 & 9.46 & 9.10 & 9.12 & 7.32 & 9.26 & 8.56 & 8.64 & 9.64 & 9.82 & 7.10 & 7.86 & 9.54 & 9.12 & 8.34 & 8.44 \\
    Human        & 8.60 & 8.56 & 7.62 & 8.10 & 7.46 & 8.80 & 8.62 & 8.46 & 8.76 & 8.66 & 6.98 & 7.10 & 8.06 & 7.38 & 7.50 & 7.06 \\
    \bottomrule
  \end{tabular}}
  \label{tab:judge_means}
\end{table*}

\begin{table*}[t]
  \centering
  \setlength{\tabcolsep}{3.0pt}
  \renewcommand{\arraystretch}{1.05}
  \caption{Linear calibration parameters $(a_{m,\ell}, b_{m,\ell})$ for each metric and language.}
  \resizebox{\textwidth}{!}{
  \begin{tabular}{lcccccccccccccccc}
    \toprule
    % & \multicolumn{4}{c}{MA} & \multicolumn{4}{c}{MS} & \multicolumn{4}{c}{MB} & \multicolumn{4}{c}{ME} \\
    % \cmidrule(lr){2-5}\cmidrule(lr){6-9}\cmidrule(lr){10-13}\cmidrule(lr){14-17}
    % & \multicolumn{2}{c}{SI} & \multicolumn{2}{c}{AF}
    % & \multicolumn{2}{c}{FA} & \multicolumn{2}{c}{FU}
    % & \multicolumn{2}{c}{AL}  & \multicolumn{2}{c}{CR}
    % & \multicolumn{2}{c}{MAC}  & \multicolumn{2}{c}{HLE} \\

    & \multicolumn{2}{c}{MA-SI} & \multicolumn{2}{c}{MA-AF}
      & \multicolumn{2}{c}{MS-FA} & \multicolumn{2}{c}{MS-FU}
      & \multicolumn{2}{c}{MB-AL}  & \multicolumn{2}{c}{MB-CR}
      & \multicolumn{2}{c}{ME-MAC}  & \multicolumn{2}{c}{ME-HLE} \\
    % \cmidrule(lr){2-3}\cmidrule(lr){4-5}\cmidrule(lr){6-7}\cmidrule(lr){8-9}
    % \cmidrule(lr){10-11}\cmidrule(lr){12-13}\cmidrule(lr){14-15}\cmidrule(lr){16-17}
    & en & zh & en & zh & en & zh & en & zh & en & zh & en & zh & en & zh & en & zh \\
    \midrule
    $a$ (intercept) & 2.133 & 1.238 & 1.189 & -3.676 & 2.245 & 2.113 & 2.793 & 1.136 & 4.464 & 2.226 & 5.016 & 3.474 & -0.183 & 1.488 & 0.757 & -0.018 \\
    $b$ (slope)     & 0.701 & 0.774 & 0.707 & 1.291 & 0.712 & 0.722 & 0.681 & 0.848 & 0.446 & 0.655 & 0.277 & 0.461 & 0.864 & 0.646 & 0.809 & 0.839 \\
    \bottomrule
  \end{tabular}}
  \label{tab:judge_calib}
\end{table*}

% \FloatBarrier

\section{Language-split Ability Correlations (RQ6)}
\label{app:rq5_language_corr}

This appendix provides language-split correlation analyses that complement the pooled results in \S\ref{sec:rq6}.
We compute ability-level scores by averaging the two metrics within each ability (MA/MS/MB/ME), and restrict to instances where all four abilities are available to ensure comparability.
We then compute Pearson correlations separately for English ($n{=}1470$) and Chinese ($n{=}1092$).

\paragraph{Pearson correlation matrices.}
Table~\ref{tab:rq5_corr_en} reports the English correlation matrix, and Table~\ref{tab:rq5_corr_zh} reports the Chinese matrix.
Both exhibit the same qualitative pattern as the pooled analysis: positive, weak-to-moderate coupling across abilities, with the strongest association typically involving ME.

\paragraph{Upstream--ME correlations.}
We additionally report language-split Pearson correlations between an aggregated upstream score (mean of MA/MS/MB) and ME:
$r_{\text{EN}}{=}0.34$ and $r_{\text{ZH}}{=}0.38$.

\begin{table}[t]
\centering
\small
\setlength{\tabcolsep}{6pt}
\renewcommand{\arraystretch}{1.12}
\caption{Pearson correlation matrix of ability scores on the English subset. ($n{=}1470$)}
\label{tab:rq5_corr_en}
\begin{tabular}{lcccc}
\toprule
 & MA & MS & MB & ME \\
\midrule
MA & 1.0000 & 0.1487 & 0.0993 & 0.3495 \\
MS & 0.1487 & 1.0000 & 0.0794 & 0.2286 \\
MB & 0.0993 & 0.0794 & 1.0000 & 0.1059 \\
ME & 0.3495 & 0.2286 & 0.1059 & 1.0000 \\
\bottomrule
\end{tabular}
\end{table}

\begin{table}[t]
\centering
\small
\setlength{\tabcolsep}{6pt}
\renewcommand{\arraystretch}{1.12}
\caption{Pearson correlation matrix of ability scores on the Chinese subset. ($n{=}1092$)}
\label{tab:rq5_corr_zh}
\begin{tabular}{lcccc}
\toprule
 & MA & MS & MB & ME \\
\midrule
MA & 1.0000 & 0.2902 & 0.0892 & 0.3940 \\
MS & 0.2902 & 1.0000 & 0.0365 & 0.2663 \\
MB & 0.0892 & 0.0365 & 1.0000 & 0.1421 \\
ME & 0.3940 & 0.2663 & 0.1421 & 1.0000 \\
\bottomrule
\end{tabular}
\end{table}

% \FloatBarrier

\section{Efficiency Analysis: Token Budget (RQ7)}
\label{app:token_budget}

To verify that MRPrompt gains are not driven by longer prompts, we measure token usage under the three persona conditions (Base, Card, MRPrompt) on the same shared STM pool.
Table~\ref{tab:token_budget} reports the average prompt/completion/total tokens (English/Chinese) under each condition, aggregated across instances.
Overall, MRPrompt does not substantially increase total tokens compared to the baselines, indicating that its gains come primarily from structured memory representation and protocol guidance rather than longer prompts.

\begin{table*}[t]
  \centering
  \footnotesize
  \setlength{\tabcolsep}{4pt}
  \renewcommand{\arraystretch}{1.15}
  \sisetup{
    table-number-alignment = center,
    round-mode = places,
    round-precision = 2
  }
  \caption{\textbf{Average token usage on the shared STM pool.} 
  Prompt ($\mathcal{M}_L$+$\mathcal{M}_S$+constraints), completion ($\hat{y}$), and total tokens per model--persona--language (en/zh).}
  \label{tab:token_budget}
  \begin{tabular}{
    l l
    S[table-format=4.2] S[table-format=4.2]
    S[table-format=2.2] S[table-format=2.2]
    S[table-format=4.2] S[table-format=4.2]
  }
    \toprule
    \multirow{2}{*}{Model} & \multirow{2}{*}{Persona}
      & \multicolumn{2}{c}{Prompt} & \multicolumn{2}{c}{Comp} & \multicolumn{2}{c}{Total} \\
    \cmidrule(lr){3-4}\cmidrule(lr){5-6}\cmidrule(lr){7-8}
      & & {en} & {zh} & {en} & {zh} & {en} & {zh} \\
    \midrule
    \multirow{3}{*}{Qwen3-4B}
      & Base & 3726.14 & 2873.16 & 65.39 & 53.02 & 3791.53 & 2926.19 \\
      & Card & 2866.16 & 2585.43 & 60.61 & 47.22 & 2926.77 & 2632.65 \\
      & MRPrompt & 3037.74 & 2825.49 & 65.40 & 56.79 & 3103.14 & 2882.28 \\
    \midrule
    \multirow{3}{*}{GLM-4-9B-Chat}
      & Base & 3673.54 & 2699.86 & 56.35 & 49.35 & 3729.89 & 2749.21 \\
      & Card & 2822.13 & 2455.76 & 58.46 & 48.02 & 2880.59 & 2503.78 \\
      & MRPrompt & 2994.50 & 2694.10 & 53.13 & 52.60 & 3047.63 & 2746.70 \\
    \bottomrule
  \end{tabular}
\end{table*}

\section{Additional Qualitative Case Studies (RQ8)}
\label{app:rq5_materials}

This appendix reports the qualitative materials for RQ8, which complement the main quantitative results by inspecting concrete MDRP behaviors under MREval.
We analyze two representative characters (Tom Sawyer; Charles Darnay) using \textit{GLM-4-9B-Chat}, comparing Base, Card, and MRPrompt.
For each case, we present the extended dialogue context ($\mathcal{M}_S$), the gold continuation $\hat{y}^{\text{gold}}$, and the three model outputs $\hat{y}$.

\paragraph{Tom Sawyer: facet-aligned success.}
Table~\ref{tab:rq5_tom_ext} illustrates the ``whitewashing'' episode, where Tom should theatrically concede only after Ben has fully taken the bait. Under Base, the model follows the plot but injects meta narration (e.g., ``breaking character'') and adopts an apologetic tone, diluting Tom's manipulative bravado. Card improves liveliness but remains generic, failing to clearly preserve the ``exclusive opportunity'' framing that motivates Ben's eagerness. In contrast, MRPrompt yields a short, in-character concession (``But don't say I didn't warn you'') that preserves Tom's playful dominance and aligns with the intended facet, qualitatively consistent with the MS/ME gains observed in RQ1--RQ3.

\paragraph{Charles Darnay: residual flattening.}
Table~\ref{tab:rq5_darnay_ext} examines Darnay's promise to withhold his present name. In the novel, this moment carries dramatic irony and latent moral tension; a faithful continuation should acknowledge the oddity and weight of the constraint while remaining deferential. All three conditions largely smooth this tension into a generic vow. Base is polite but non-committal about the unusual request; Card slightly misattributes the secrecy (``your secret''); MRPrompt best tracks the object of the promise (the \emph{name}) and ties compliance to Lucie's happiness, yet still under-expresses the implied inner unease. This case highlights a remaining failure mode: even when persona voice is grounded, models may under-realise \emph{implicit} narrative constraints (suppressed information and deferred revelation) as rhetorically salient dialogue.

\paragraph{Takeaways.}
Overall, the case studies support the main quantitative story. When success is primarily driven by \emph{scene-conditioned facet selection} and persona-faithful micro-decisions (Tom), MRPrompt produces more locally appropriate, in-character continuations than Base/Card. When success depends on \emph{subtle management of latent long-term information} (Darnay), even MRPrompt can converge to fluent but flattened outputs, consistent with RQ4: upstream gains tend to improve ME, but do not guarantee fine-grained realisation of suppressed knowledge and internal conflict.

\paragraph{Persona references.}
Brief persona descriptions for these two characters are provided in Appendix~\ref{app:persona_examples_tom} and Appendix~\ref{app:persona_examples_darnay}; full persona files are included in the supplementary material.

% 详细的case说明

\begin{table*}[t]
  \centering
  \scriptsize
  \setlength{\tabcolsep}{4.5pt}
  \renewcommand{\arraystretch}{1.08}
  \caption{\textbf{Tom Sawyer case.} Full context and continuations under three persona conditions.}
  \label{tab:rq5_tom_ext}
  \begin{tabularx}{\textwidth}{@{}p{0.16\textwidth}>{\raggedright\arraybackslash}X@{}}
    \toprule
    \textbf{Element} & \textbf{Text} \\
    \midrule
    \textbf{Full dialogue context} &
    \begin{minipage}[t]{\linewidth}\footnotesize
    \textbf{Ben Rogers}: (approaches Tom with a skip in his step) Hello, old chap, you got to work, hey?\newline
    \textbf{Tom Sawyer}: (pretends to be focused on his work, then looks up with feigned surprise) Why, it's you, Ben! I warn't noticing.\newline
    \textbf{Ben Rogers}: Say---I'm going in a-swimming, I am. Don't you wish you could? But of course you'd druther work---wouldn't you? Course you would!\newline
    \textbf{Tom Sawyer}: (looks thoughtful) What do you call work?\newline
    \textbf{Ben Rogers}: (looks confused) Why, ain't that work?\newline
    \textbf{Tom Sawyer}: (resumes whitewashing casually) Well, maybe it is, and maybe it ain't. All I know, is, it suits Tom Sawyer.\newline
    \textbf{Ben Rogers}: (looks skeptical) Oh come, now, you don't mean to let on that you like it?\newline
    \textbf{Tom Sawyer}: (continues painting with exaggerated care) Like it? Well, I don't see why I oughtn't to like it. Does a boy get a chance to whitewash a fence every day?\newline
    \textbf{Ben Rogers}: (watches Tom intently) Say, Tom, let me whitewash a little.\newline
    \textbf{Tom Sawyer}: (pretends to consider reluctantly) No---no---I reckon it wouldn't hardly do, Ben. You see, Aunt Polly's awful particular about this fence---right here on the street, you know---but if it was the back fence I wouldn't mind and she wouldn't. Yes, she's awful particular about this fence; it's got to be done very careful; I reckon there ain't one boy in a thousand, maybe two thousand, that can do it the way it's got to be done.\newline
    \textbf{Ben Rogers}: (eagerly) No---is that so? Oh come, now---lemme just try. Only just a little---I'd let you, if you was me, Tom.\newline
    \textbf{Tom Sawyer}: (hesitates dramatically) Ben, I'd like to, honest injun; but Aunt Polly---well, Jim wanted to do it, but she wouldn't let him; Sid wanted to do it, and she wouldn't let Sid. Now don't you see how I'm fixed? If you was to tackle this fence and anything was to happen to it---\newline
    \textbf{Ben Rogers}: (interrupts eagerly) Oh, shucks, I'll be just as careful. Now lemme try. Say---I'll give you the core of my apple.\newline
    \textbf{Tom Sawyer}: (pretends to be reluctant) Well, here---No, Ben, now don't. I'm afeard---\newline
    \textbf{Ben Rogers}: (desperately) I'll give you all of it!
    \end{minipage} \\
    \midrule
    \textbf{Gold continuation} &
    Tom Sawyer: [Success! Time to graciously concede.] (hands over the brush with feigned reluctance) Well, I suppose... if you're sure you want to, Ben. \\
    \midrule
    \textbf{Base output} &
    Tom Sawyer: (smiles, breaking character) OK, Ben. But just a little bit, and then you'll have to stop. I don't want to be accused of sharing if I didn't. \\
    \midrule
    \textbf{Card output} &
    Tom Sawyer: (playfully) Alright, Ben! I reckon you can handle a little fence painting. Just don't mess it up, or I'll have to whitewash the whole town next! \\
    \midrule
    \textbf{MRPrompt output} &
    Tom Sawyer: (smiles, gives in playfully) Alright, Ben! But don't say I didn't warn you. (starts to hand over the brush) \\
    \bottomrule
  \end{tabularx}
\end{table*}

\begin{table*}[t]
  \centering
  \scriptsize
  \setlength{\tabcolsep}{4.5pt}
  \renewcommand{\arraystretch}{1.08}
  \caption{\textbf{Charles Darnay case.} Full context and continuations under three persona conditions.}
  \label{tab:rq5_darnay_ext}
  \begin{tabularx}{\textwidth}{@{}p{0.16\textwidth}>{\raggedright\arraybackslash}X@{}}
    \toprule
    \textbf{Element} & \textbf{Text} \\
    \midrule
    \textbf{Full dialogue context} &
    \begin{minipage}[t]{\linewidth}\footnotesize
    \textbf{Charles Darnay}: Doctor Manette, I have come to speak to you about a matter of great importance to me.\newline
    \textbf{Doctor Manette}: (with evident constraint) Is Lucie the topic?\newline
    \textbf{Charles Darnay}: Yes, sir. I want to express that I love your daughter fondly, dearly, disinterestedly, devotedly.\newline
    \textbf{Doctor Manette}: (turning away) Not that, sir! Let that be! I adjure you, do not recall that!\newline
    \textbf{Charles Darnay}: Doctor Manette, I understand the unique bond you share with Lucie. I would never seek to come between you.\newline
    \textbf{Doctor Manette}: (after a pause) I do not doubt your loving Lucie; you may be satisfied of it.\newline
    \textbf{Charles Darnay}: Sir, I would never dream of separating Lucie from you. My hope is to join your family, not divide it.\newline
    \textbf{Doctor Manette}: Have you spoken to Lucie?\newline
    \textbf{Charles Darnay}: No, sir. Nor have I written to her. I wanted to speak with you first.\newline
    \textbf{Doctor Manette}: Your self-denial is to be referred to your consideration for her father. Her father thanks you.\newline
    \textbf{Charles Darnay}: Doctor Manette, I know the depth of Lucie's love for you. I only hope to add to that love, not diminish it in any way.\newline
    \textbf{Doctor Manette}: If she should ever tell me that you are essential to her perfect happiness, I will give her to you. But you must promise me something.\newline
    \textbf{Charles Darnay}: Of course, sir. What is it you wish me to promise?\newline
    \textbf{Doctor Manette}: You must not tell me your present name. If your suit should prosper, you shall tell me on your marriage morning. Do you promise?
    \end{minipage} \\
    \midrule
    \textbf{Gold continuation} &
    Charles Darnay: [This is an odd request, but I must respect it.] Willingly, Doctor Manette. You have my word. \\
    \midrule
    \textbf{Base output} &
    Charles Darnay: Yes, sir, I promise. And I assure you, it is with the greatest respect and honor that I make this promise. \\
    \midrule
    \textbf{Card output} &
    Charles Darnay: I promise, Doctor Manette. I will keep your secret until our marriage day, if it be as you wish. \\
    \midrule
    \textbf{MRPrompt output} &
    Charles Darnay: I promise, sir. My name will remain my own until the day we marry. It is but a small thing, in the grand scheme of your approval and Lucie's happiness. \\
    \bottomrule
  \end{tabularx}
\end{table*}
% \FloatBarrier

\section{MREval Metrics and Rubrics}
\label{app:metrics_rubrics}

Table~\ref{tab:metrics_rubrics} summarizes the eight MREval metrics and their corresponding scoring rubrics. 
Each metric is rated on a 1--10 Likert-style scale (higher is better), using 1/5/10 anchor descriptions.

\begin{table*}[t]
\centering
\small
\renewcommand{\arraystretch}{1.2}
\setlength{\tabcolsep}{6pt}
\caption{MREval metrics and scoring rubrics}
\begin{tabular}{p{0.18\textwidth} p{0.32\textwidth} p{0.46\textwidth}}
\hline
\textbf{Metrics} & \textbf{Explanations} & \textbf{Rubrics} \\
\hline
MA-SI (Source Invariance) &
Measures consistency between the response generated with the anonymized persona $\hat{y}^{\text{anon}}$ and the original one $\hat{y}$. A high score indicates grounding in persona semantics, not name priors. &
1: The two answers differ greatly in meaning, i.e., the model almost entirely relies on pretrained memory and clearly ignores the task-provided LTM/STM; 5: The two answers differ but both look reasonable, i.e., the model mixes task memory with pretrained memory, partially using LTM/STM but often dragged by old knowledge; 10: The two answers are almost semantically identical, i.e., the model mainly builds its answer on task LTM/STM with minimal interference from pretrained name-based memory. \\
\hline
MA-AF (Alias Fidelity) &
Assesses whether the behavior under an anonymized persona $\hat{y}^{\text{anon}}$ remains faithful to the original intended character, using the ground-truth response $\hat{y}^{\text{gold}}$ as an anchor. &
1: Once names and identity cues are removed, the answer's meaning and style drift drastically, indicating the model has not truly learned the task memory; 5: After de-identification, key information is partly preserved but tone or details become noticeably unstable; 10: After removing names, the answer keeps key content, tone, and reasoning path largely consistent, showing stable task memory learning under anonymization. \\
\hline
MS-FA (Facet Alignment) &
Quantifies the model's precision in selecting the correct scene facet by contrasting responses under the true $\mathcal{M}_L$ versus a counterfactual (inverted) LTM $\mathcal{M}_L^{\text{anti}}$. &
1: Outputs under different scene-facet configurations are almost the same, with no clear distinction between original/reversed facets; 5: There are some differences in tone or stance, but they are unstable and each output only weakly matches its intended facet; 10: Under the same STM, outputs for different facet settings are clearly separable and each is highly faithful to the expected personality and behaviour of its own facet. \\
\hline
MS-FU (Facet Utility) &
Measures the improvement gained by including scene-specific facets in the LTM, compared to a scene-ablated LTM $\mathcal{M}_L^{\text{no-scene}}$. &
1: With scene facets added, the answer is almost indistinguishable from, or even worse than, the no-facet version; 5: Some improvement in alignment is visible, but there are still many generic behaviours or misused scenes; 10: Under the same STM, adding scene facets makes the answer clearly more consistent with the persona, tone, and behavioural expectations, with role-playing quality significantly better than the no-facet setting. \\
% \hline
% \end{tabular}
% % \caption{MREval metrics and scoring rubrics}
% % \label{tab:metrics_rubrics}
% \end{table*}

% \begin{table*}[t]
% \centering
% \small
% \renewcommand{\arraystretch}{1.2}
% \setlength{\tabcolsep}{6pt}
% \begin{tabular}{p{0.18\textwidth} p{0.32\textwidth} p{0.46\textwidth}}
% \hline
% \textbf{Metrics} & \textbf{Explanations} & \textbf{Rubrics} \\
\hline
MB-AL (Answer Leakage) &
Scores the model's ability to avoid generating a forbidden reference answer $\hat{y}^{\text{out}}$ when presented with an out-of-scope prompt $c_K^{\text{out}}$ (e.g., a future plot spoiler). &
1: Frequently leaks future plot points or out-of-book information, effectively using an omniscient view instead of the character's current-time perspective; 5: Generally respects the timeline but occasionally inserts slightly ahead-of-time knowledge or mild spoilers; 10: Always answers strictly from the current time point, only using available memories and never revealing future or out-of-scope facts. \\
\hline
MB-CR (Controlled Response) &
Assesses the appropriateness of the model's response strategy to out-of-scope prompts, favoring expressions of uncertainty, refusal, or grounded speculation over confident fabrication. &
1: When faced with clearly out-of-memory/time-range questions, tends to fabricate confident answers with no self-restraint; 5: Sometimes expresses uncertainty, but the justification is vague or still mixed with speculation; 10: Clearly recognizes out-of-scope questions and responds with polite, explicit, memory-boundary-based refusal or uncertainty instead of hallucination. \\
\hline
ME-MAC (Memory-Aligned Coherence) &
Rates the logical and topical coherence of the response with respect to the activated memory and context. &
1: Despite having memory input, the answer is badly misaligned with context or persona, with logical jumps and serious structural confusion; 5: Overall related to the relevant memory, but with rough transitions, missing steps, or partially broken reasoning; 10: After memory alignment, the answer is well-structured, causally reasonable, and internally consistent, highly aligned with LTM/STM. \\
\hline
ME-HLE (Human-Like Enactment) &
Rates the naturalness, tonal appropriateness, and conversational fluency of the response, ensuring it embodies a human-like utterance consistent with the persona. &
1: Language is stiff and template-like, lacking emotional detail and conversational rhythm, overall feeling mechanical; 5: Basically natural, with some perceivable emotion and tone shifts, but still somewhat mechanical or flat; 10: Word choice, tone, and pacing are close to real human dialogue, showing rich yet controlled emotional expression while making good use of memory. \\
\hline
\end{tabular}
% \caption{MREval metrics and scoring rubrics}
\label{tab:metrics_rubrics}
\end{table*}
% \FloatBarrier 

\section{Prompts}
\label{app:prompt}

\subsection{Prompt Templates for Character Profiles}
\label{app:prompt:profiles}
Figures~\ref{fig:prompt_baseline_narrative_full}--\ref{fig:prompt_ours_facet_ltm} present the prompt templates used to construct (i) narrative character profiles, (ii) semi-structured persona cards, and (iii) our facet-structured long-term memory (LTM) personas.

\subsection{Prompt Templates for MRBench Split Construction (MA/MB)}
\label{app:prompt:tasks}
Figures~\ref{fig:prompt_mla_dataset} and~\ref{fig:prompt_mca_dataset} show the prompt templates used to construct the evaluation instances for the MA and MB splits in MRBench.

\subsection{Prompt Templates for Character Role-playing}
\label{app:prompt:roleplay}
Figures~\ref{fig:prompt_roleplay_original} and \ref{fig:prompt_roleplay_memory_aug} report the shared standard role-playing instruction and our specific LTM--STM control protocol.

\subsection{Prompt Templates for MREval Judging}
\label{app:prompt:eval_prompts}
Since the eight MREval metrics target different aspects of role-playing behavior, we require a set of metric-specific judging prompt templates.
Due to space limitations, we omit these templates from the appendix; the full prompt specifications are provided in the supplementary material.

\begin{figure*}[t]
  \centering

  \begin{minipage}[t]{0.49\textwidth}
    \centering
    \begin{ListingBoxTaggedFloat}{Prompt}{Chinese; abridged}
    \scriptsize
    \begin{CJK*}{UTF8}{gbsn}
    \begin{Code}
system_prompt_zh = (
  "你是一名专门为小说和影视角色撰写[人物长篇侧写]的专家，"
  "你擅长以叙事的方式从**原著中的关键情节**中呈现人物的性格和发展，而不是通过抽象的性格标签或总结性描述。"
  "你的任务是通过选取角色的**经典情节**，展现其在故事中的成长、转变和内心活动，"
  "将人物的经历呈现为**情节驱动的叙事**，而非单纯的性格概括。"
  "你的输出必须是合法的 JSON 字符串，不要输出任何解释、markdown 或额外文字。"
)

user_prompt_zh = (
  f"现在请你为小说《{novel}》中的角色「{character_name}」撰写一份[非结构化的人物长期记忆 summary]。\n\n"
  "重要要求：\n"
  "1. 最终输出必须是一个 JSON 对象，且**只能**包含一个顶层键：\"summary\"。\n"
  "2. summary 必须是多段落长文本，且**基于原著具体情节**（关键事件、行为言语、情感反应、场景细节），"
  "不得写成抽象标签或泛泛概括。\n"
  "3. 使用**固定标题格式**组织：先「[角色概览]」，后续至少 6 个「[场景X：xxx]」。\n"
  "4. [角色概览]：用 14--15 段话详细叙述生平与核心性格变化（用情节支撑，而非性格标签堆砌）。\n"
  "5. 每个[场景X]：必须对应原著具体情节，围绕时间/地点/背景、情绪、行为与说话风格、思考与决策过程展开；"
  "如与概览有反差需指出并解释；每个场景用 9--10 段话细写，避免泛泛概括。\n"
  "6. 输出必须为合法 JSON（仅一个顶层键；字符串用双引号；允许换行；JSON 外不输出任何文字）。\n"
)
    \end{Code}
    \end{CJK*}
    \end{ListingBoxTaggedFloat}
  \end{minipage}\hfill
  \begin{minipage}[t]{0.49\textwidth}
    \centering
    \begin{ListingBoxTaggedFloat}{Prompt}{English; abridged}
    \begin{Code}
system_prompt_en = (
  "You are an expert at writing long-form character dossiers for fictional characters in novels and films/TV. "
  "You specialize in revealing a character’s personality and development through **key canonical plot events**, "
  "not through abstract trait labels or generic summaries. "
  "Your job is to select **iconic scenes** and narrate how the character grows, changes, and thinks internally, "
  "presenting their life as **plot-driven storytelling** rather than a list of traits. "
  "Your output must be a valid JSON string. Do not output any explanations, markdown, or extra text."
)

user_prompt_en = (
  f"Now write an **unstructured long-term memory summary** for the character \"{character_name}\" "
  f"from the novel \"{novel}\".\n\n"
  "Important requirements:\n"
  "1. Output must be a JSON object with **only** one top-level key: \"summary\".\n"
  "2. The summary must be a long multi-paragraph text **grounded in specific canonical plot events** "
  "(major life events, actions/lines, emotions, concrete scenes), not abstract trait lists.\n"
  "3. Use a **fixed heading format**: start with \"[Overview]\", then at least 6 \"[Scene X: xxx]\" facets.\n"
  "4. [Overview]: use 8--9 paragraphs to cover life trajectory and core personality changes (scene-supported).\n"
  "5. Each [Scene X]: a concrete canonical episode, describing time/place/background, emotion, speaking/behavioural style, "
  "thinking/decision process; note and explain any contrast with [Overview]; write with rich detail.\n"
  "6. Output must be valid JSON only (double quotes, newlines allowed, no extra text outside JSON).\n"
)
    \end{Code}
    \end{ListingBoxTaggedFloat}
  \end{minipage}

  \caption{Prompt template for narrative character profiles (Base--LTM).}
  \label{fig:prompt_baseline_narrative_full}
\end{figure*}

\begin{figure*}[t]
  \centering

  \begin{minipage}[t]{0.49\textwidth}
    \centering
    \begin{ListingBoxTaggedFloat}{Prompt}{Chinese; abridged}
    \scriptsize
    \begin{CJK*}{UTF8}{gbsn}
    \begin{Code}
system_prompt_zh = (
  "你是一名为大语言模型编写[半结构化人物卡]的专家。"
  "现在你将看到某个角色的非结构化长期记忆 summary（包含[角色概览]和多个[场景X]段落），"
  "这些内容已经比较故事化，但仍是连续自然语言。"
  "你的目标不是做复杂的心理学建模，而是结合你的知识："
  "  1）在整体不改事实和人格走向的前提下，把信息压缩成一份简洁、易读的人物卡；"
  "  2）列出若干简短的性格标签（core_traits），方便下游模型快速抓住人设；"
  "  3）保留若干代表性的“关键场景条目”（scene_facets），每条对应一个或少数几个具体情节；"
  "并且：字段集合必须**有限且简单**，只允许使用 name / Nickname / Relationships / global_summary / "
  "Personality 这几个顶层键，"
  "在 Personality 内只允许 core_traits 和 scene_facets，且每个字段内部也要保持精简。"
  "请务必输出严格合法的 JSON，不能包含任何解释性文字或 Markdown。"
)

user_prompt_zh = f"""
角色「{name}」的原始 summary 如下：
{summary}

任务：在不改事实与宏观性格走向的前提下，把 summary 压缩为弱结构化人物卡：
- global_summary：7–8 段概括生平与性格变化；
- core_traits：4–8 个简短 trait 标签（仅 trait 字段）；
- scene_facets：8–10 个关键场景条目（贴近原[场景X]情节；保留时间/背景、情绪、典型行为；可合并相似场景）。

仅输出严格合法 JSON（不可增删顶层键），结构为：
{
"{{name}}": {
  "name": "...",
  "Nickname": "...",
  "Relationships": [{"name":"...","relationship":"..."}],
  "global_summary": "...",
  "Personality": {
    "core_traits": [{"trait":"..."}],
    "scene_facets": [{
      "title":"...",
      "situation":"...",
      "emotional_state":"...",
      "behavior_pattern":"..."
    }]
  }
}
}

规则：不杜撰与 summary 矛盾的新重大经历；
JSON 外不要输出任何文字。
"""
    \end{Code}
    \end{CJK*}
    \end{ListingBoxTaggedFloat}
  \end{minipage}\hfill
  \begin{minipage}[t]{0.49\textwidth}
    \centering
    \begin{ListingBoxTaggedFloat}{Prompt}{English; abridged}
    \begin{Code}
system_prompt_en = (
  "You are an expert in designing *semi-structured* persona cards for large language models. "
  "You will be given a long, episodic character summary with an [Overview] and several [Scene X] sections. "
  "These are already story-like. Your goal is NOT to perform complex psychological modeling, but to combine your knowledge to: "
  "  (1) compress the information into a concise and readable persona card; "
  "  (2) list a small set of short core trait labels (core_traits) for quick persona grasp; "
  "  (3) keep several representative key-scene entries (scene_facets) that stay close to concrete episodes "
  "      in terms of time/background, emotion, and behaviour. "
  "The allowed top-level keys are strictly limited to: "
  "  name / Nickname / Relationships / global_summary / Personality. "
  "Inside Personality, only core_traits and scene_facets are allowed, and each must use a simple internal structure. "
  "Always output valid JSON only, with no extra commentary."
)

user_prompt_en = f"""
Original summary for "{name}":
{summary}

Task: compress into a weakly structured persona card:
- global_summary: 6–7 paragraphs;
- core_traits: 4–8 short trait labels (trait only);
- scene_facets: 8–10 key-scene entries (close to original scenes; keep time/background, emotion, typical behaviour; may merge similar scenes).

Output ONLY valid JSON with this schema (do not add/remove keys):
{
"{{name}}": {
  "name":"...",
  "Nickname":"...",
  "Relationships":[{"name":"...","relationship":"..."}],
  "global_summary":"...",
  "Personality":{
    "core_traits":[{"trait":"..."}],
    "scene_facets":[{
      "title":"...",
      "situation":"...",
      "emotional_state":"...",
      "behavior_pattern":"..."
    }]
  }
}
}

Rules: do not invent contradicting major events;
output JSON only.
"""
    \end{Code}
    \end{ListingBoxTaggedFloat}
  \end{minipage}

  \caption{Prompt template for the semi-structured Card persona (Card--LTM).}
  \label{fig:prompt_sota_card}
\end{figure*}

\begin{figure*}[t]
  \centering

  \begin{minipage}[t]{0.49\textwidth}
    \centering
    \begin{ListingBoxTaggedFloat}{Prompt}{Chinese; abridged}
    \scriptsize
    \begin{CJK*}{UTF8}{gbsn}
    \begin{Code}
system_prompt_zh = (
  "你是一名人格与故事建模专家，擅长把长篇、情节化的人物描述抽象为可计算、可检索的结构化人格画像。"
  "你将看到某个角色的非结构化长期记忆 summary（包含[角色概览]和多个[场景X]段落），"
  "这些段落已经较为“故事化”，但仍然是自然语言长文本。"
  "你的任务不是简单地逐段重写，而是："
  "  1）识别其中的核心性格维度；"
  "  2）将多段相似情境抽象合并为若干“场景切面（scene facets）”；"
  "  3）输出一个结构化 JSON persona，便于下游模型按“场景切面 + 触发线索”来检索和调用。"
  "请务必输出严格合法的 JSON，不能包含任何解释性文字或 Markdown。"
)

user_prompt_zh = f"""
角色「{name}」原始 summary：
{summary}

任务：在不改变事实与宏观性格走向的前提下，将 summary 抽象为结构化 persona（跨场景归纳），
输出仅包含如下 JSON（不可增删顶层键）：

{
"{{name}}":{
  "name":"...",
  "Nickname":"...",
  "Relationships":"... (optional)",
  "global_summary":"... (1-2 paragraphs; abstract view)",
  "Personality":{
    "core_traits":[{"trait":"...","desc":"..."}],
    "scene_facets":[
      {
        "title":"...",
        "time_scope":[...],
        "situation":"...",
        "social_role":[...],
        "emotional_state":"...",
        "behavior_pattern":"...",
        "thinking_pattern":"...",
        "conflict_with_core":"...",
        "source_scenes":[...],
        "cue_phrases":[...]
      }
    ]
  }
}
}

约束：
1) 不编造与原 summary 矛盾的新经历；只做结构化、抽象与跨场景归纳；
2) core_traits 为来自多个情节的“向上抽象”，建议 4–8 个；
3) scene_facets 给 5–8 个：可合并相似场景，也可单列关键场景；覆盖主要情境类型；
4) 输出必须严格合法 JSON，JSON 外不输出任何文字。
"""
    \end{Code}
    \end{CJK*}
    \end{ListingBoxTaggedFloat}
  \end{minipage}\hfill
  \begin{minipage}[t]{0.49\textwidth}
    \centering
    \begin{ListingBoxTaggedFloat}{Prompt}{English; abridged}
    \begin{Code}
system_prompt_en = (
  "You are an expert in personality and narrative modeling. "
  "You will be given a long, episodic character summary that already contains an [Overview] and several [Scene X] paragraphs. "
  "These paragraphs are story-like and grounded in concrete events. "
  "Your job is **not** to paraphrase them, but to:\n"
  "  (1) identify the underlying stable traits,\n"
  "  (2) cluster similar situations into a small set of scene facets,\n"
  "  (3) output a structured JSON persona that makes those facets explicit for downstream retrieval."
  "Always output valid JSON only, with no extra commentary."
)

user_prompt_en = f"""
Original summary for "{name}":
{summary}

Task: reorganize into a structured persona with reusable scene facets (merge/summarize across similar scenes).
Output ONLY the following JSON schema (do not add/remove top-level keys):

{
"{{name}}":{
  "name":"...",
  "Nickname":"...",
  "Relationships":"... (optional)",
  "global_summary":"... (1-2 paragraphs; abstract view)",
  "Personality":{
    "core_traits":[{"trait":"...","desc":"..."}],
    "scene_facets":[
      {
        "title":"...",
        "time_scope":[...],
        "situation":"...",
        "social_role":[...],
        "emotional_state":"...",
        "behavior_pattern":"...",
        "thinking_pattern":"...",
        "conflict_with_core":"...",
        "source_scenes":[...],
        "cue_phrases":[...]
      }
    ]
  }
}
}

Rules: stay faithful to the original summary; core_traits should be true abstractions (aim ~4–8);
produce 5–8 scene_facets that cover major situational patterns (merge similar scenes; single out crucial ones);
output valid JSON only.
"""
    \end{Code}
    \end{ListingBoxTaggedFloat}
  \end{minipage}

  \caption{Prompt template for the facet-structured persona LTM used in MRPrompt (MRPrompt--LTM).}
  \label{fig:prompt_ours_facet_ltm}
\end{figure*}

\begin{figure*}[t]
  \centering

  \begin{minipage}[t]{0.49\textwidth}
    \centering
    \begin{ListingBoxTaggedFloat}{Prompt}{Chinese}
    \scriptsize
    \begin{CJK*}{UTF8}{gbsn}
    \begin{Code}
prompt_zh = f"""
你是一个对话文本重写专家。任务：把给定的对话上下文(context)、主题(topic)和参考回复(expected_output)重写为“只替换角色姓名为虚拟姓名且句子通顺”的版本，**不改变语义、情感或逻辑**。

说明：
1) 下面给出的是原始对话（context）、原始主题(topic)和期望回复（expected_output）。
2) 你需要将文中出现的以下原始姓名替换为对应的虚拟姓名（只替换这些映射中出现的名字）：
{mapping_lines}
3) 特别关注主角：原主角名为：{main_role}，对应虚拟名：{virtual_role}。虚拟人设如下（供文风参考），但请**不要改变原句含义**，仅用于确保用词风格一致：
{json.dumps(virt_profile, ensure_ascii=False, indent=2)}
4) 输出要求：**严格**只输出一个 JSON 对象（不要任何额外文本或解释），格式为：
{{"context": "<rewritten context>", "expected_output": "<rewritten expected_output>", "topic": "<rewritten topic>"}}
5) 要保证对话的说话者标签形式（例如 “姓名：内容”）保持正确，句子通顺自然，且不要留下任何原始名字残留。
6) 在“context”字段中，用**一个换行符（“\\n”）**分隔每个说话者的话语，并且不要插入多余的空行。

下面是原始文本：
<ORIGINAL_CONTEXT>
{orig_context}
</ORIGINAL_CONTEXT>

<ORIGINAL_EXPECTED>
{orig_expected}
</ORIGINAL_EXPECTED>

<ORIGINAL_TOPIC>
{orig_topic}
</ORIGINAL_TOPIC>

现在开始重写，并只输出 JSON。
"""
    \end{Code}
    \end{CJK*}
    \end{ListingBoxTaggedFloat}
  \end{minipage}\hfill
  \begin{minipage}[t]{0.49\textwidth}
    \centering
    \begin{ListingBoxTaggedFloat}{Prompt}{English}
    \begin{Code}
prompt_en = f"""
You are an expert in rewriting dialogue text. Task: rewrite the given dialogue context, topic, and expected output so that **only character names are replaced by the mapped virtual names** while preserving meaning, tone and logic.

Instructions:
1) The following mapping lists original names and their corresponding virtual names (replace only names present in this mapping):
{mapping_lines}
2) Main role: original name = {main_role}, virtual name = {virtual_role}. Here is the virtual character profile (for style/reference only):
{json.dumps(virt_profile, ensure_ascii=False, indent=2)}
3) Output requirement: **only** return a single JSON object, nothing else, exactly in this form:
{{"context": "<rewritten context>", "expected_output": "<rewritten expected_output>", "topic": "<rewritten topic>"}}
4) Ensure speaker labels (e.g., "Name: ...") remain correct; ensure fluent, natural sentences; leave no residual original names.
5) In the “context” field, separate each speaker’s utterance with **a single newline character (“\\n”)**, and do not insert extra blank lines.

Original context:
<ORIGINAL_CONTEXT>
{orig_context}
</ORIGINAL_CONTEXT>

Original expected output:
<ORIGINAL_EXPECTED>
{orig_expected}
</ORIGINAL_EXPECTED>

Original topic:
<ORIGINAL_TOPIC>
{orig_topic}
</ORIGINAL_TOPIC>

Begin rewriting now and output strict JSON.
"""
    \end{Code}
    \end{ListingBoxTaggedFloat}
  \end{minipage}

  \caption{Prompt template for constructing the MA evaluation dataset.}
  \label{fig:prompt_mla_dataset}
\end{figure*}

\begin{figure*}[t]
  \centering

  \begin{minipage}[t]{0.49\textwidth}
    \centering
    \begin{ListingBoxTaggedFloat}{Prompt}{Chinese}
    \scriptsize
    \begin{CJK*}{UTF8}{gbsn}
    \begin{Code}
prompt_zh = f"""
你是时点一致性与角色知识边界专家。
输入包含：history（对话历史）、query（当前由最后发言者提出的一句）和背景信息：role、scenario、novel_name。
请先基于 history + scenario + novel_name 判断 role 的当前时间点（简短说明），
然后严格按脚本指定类型生成问题：{desired_type}。

如果 desired_type 是 future_experience，
问题必须是 {role_play} 在当前时间点之后才可能经历的事件；
如果 desired_type 是 unknown_field，
问题必须超出 {role_play} 的知识/领域范围。

问题应当由 query 的发话人 {role_ask} 提问，
并且问题文本必须不出现在 history 中。

请只输出合法 JSON：
{{
  "question": "{role_ask}：...",
  "answer": "...",
  "meta": "简短时间点说明或额外信息"
}}

answer 里面的回答并不是角色扮演，
而是这个问题本身在书中/科学领域的标准答案；
并且不用借 {role_play} 之口说出，
而是直接回答问题的答案，不做其他解释。
"""
    \end{Code}
    \end{CJK*}
    \end{ListingBoxTaggedFloat}
  \end{minipage}\hfill
  \begin{minipage}[t]{0.49\textwidth}
    \centering
    \begin{ListingBoxTaggedFloat}{Prompt}{English}
    \begin{Code}
prompt_en = f"""
You are an expert in temporal consistency and character knowledge boundaries.
Input includes history, query, role, scenario, novel_name.
First briefly determine the role's current timeline based on history + scenario + novel_name,
then STRICTLY follow the script-specified desired_type: {desired_type}.

If desired_type is future_experience,
produce a question concerning an event only after the current timeline of {role_play};
If desired_type is unknown_field,
produce a question outside the {role_play}'s known domain.

Question must be asked by the 'from' speaker of query {role_ask}
and must NOT appear anywhere in history.

Output ONLY a valid JSON:
{{
  "question": "{role_ask}: ...",
  "answer": "...",
  "meta": "brief timeline note or extra info"
}}

The content in answer is not a role-play response,
but rather the standard answer to the question itself as presented in the book/scientific domain.
It should be given directly, without speaking through the voice of {role_play}
or doing any other explanation.
"""
    \end{Code}
    \end{ListingBoxTaggedFloat}
  \end{minipage}

  \caption{Prompt template for constructing the MB evaluation dataset.}
  \label{fig:prompt_mca_dataset}
\end{figure*}

\begin{figure*}[t]
  \centering

  \begin{minipage}[t]{0.49\textwidth}
    \centering
    \begin{ListingBoxTaggedFloat}{Prompt}{Chinese}
    \scriptsize
    \begin{CJK*}{UTF8}{gbsn}
    \begin{Code}
role_system_zh = f'''{role_information}

你现在是 {role}。你必须在角色扮演对话中**完全按照 {role} 的方式说话和行动**。  
不要提及你是一个 AI。无论在任何情况下，都不得脱离角色。

严格规则：
1. 你的回答**必须以 "{role}:" 开头**，且不能包含其他前缀。  
2. 在回答中，你可以使用以下标记：
- [ ] 表示 {role} 的心理活动（思考、内心独白）  
- ( ) 表示 {role} 的行为活动（动作、姿态、表情）  
请在合适的时候自然使用，而不是机械地每次都使用。  
3. 完全保持角色状态。使用 {role} 的语气、说话风格和性格特征。  
4. 只输出**一轮完整的对话内容**。不要生成额外的回合，也不要替其他角色发言。  
5. **绝不要解释你在做什么**。只需以 {role} 的身份作答。
6. 回答中文。

如果你的回答未严格遵守以上规则，必须立即纠正并重新生成符合要求的回复。
'''
    \end{Code}
    \end{CJK*}
    \end{ListingBoxTaggedFloat}
  \end{minipage}\hfill
  \begin{minipage}[t]{0.49\textwidth}
    \centering
    \begin{ListingBoxTaggedFloat}{Prompt}{English}
    \begin{Code}
role_system_en = f'''{role_information}

You are now {role}. You must act and speak **exactly** as {role} would in a role-playing conversation. 
Do not mention you are an AI. Do not break character under any circumstance.

STRICT RULES:
1. Your answer MUST start with "{role}:" and cannot contain any other prefix.  
2. In your reply, you may use:
- [ ] to represent {role}'s mental activity (thoughts, inner monologue)  
- ( ) to represent {role}'s behavioral activity (actions, gestures, expressions)  
Use them naturally when appropriate, not mechanically every time.  
3. Stay fully in character. Use {role}'s tone, speaking style, and personality traits.  
4. Only output ONE complete turn of dialogue. Do NOT create extra turns or speak for other roles.  
5. NEVER explain what you are doing. Just respond as {role}. 
6. Respond in English.

If your answer does not follow these rules exactly, you must immediately correct yourself and rewrite the response.
'''
    \end{Code}
    \end{ListingBoxTaggedFloat}
  \end{minipage}

  \caption{Original prompt template for character role-playing.}
  \label{fig:prompt_roleplay_original}
\end{figure*}

\begin{figure*}[t]
  \centering

  \begin{minipage}[t]{0.49\textwidth}
    \centering
    \begin{ListingBoxTaggedFloat}{Prompt}{Chinese}
    \scriptsize
    \begin{CJK*}{UTF8}{gbsn}
    \begin{Code}
role_system_zh = f"""
【角色长期记忆 / Long-Term Memory】
以下内容是关于角色「{role}」的一份长期记忆描述，包含其一生经历、核心人格特质以及在不同情境下的性格表现（包括可能的场景切面）：
{role_information}

你已经“记住”了上述长期记忆（LTM）。在回答时，你需要：

1. 以【角色长期记忆】中的信息作为人物设定的基础：
- 核心性格与价值观
- 重要人生经历与人际关系
- 在不同场景下的典型情绪、行为和说话风格（场景切面）

2. 把接下来给出的多轮对话视为【角色短期记忆 / Short-Term Memory】：
- 这些对话发生在当前的具体场景中
- 你需要根据对话中的内容，自行判断此刻「{role}」处于哪一种情境/气氛，并激活与之最匹配的性格切面（情绪、语气、行为风格）。
- 如果找不到最匹配的性格切面，则根据你对该角色的理解，选择一个合适的切面和性格进行回应。

【扮演与生成规则】
你现在就是「{role}」。在整个对话中你必须始终以 {role} 的身份说话和行动，不得以“模型”“AI”等任何第三人称出场。

严格规则：
1. 你的回答必须以「{role}：」开头。
2. 你可以使用：
- 「[ ]」表示 {role} 的心理活动（内心独白、瞬时想法）
- 「( )」表示 {role} 的动作、表情或身体行为
在自然合适的时候使用，而不是每句都用。
3. 只输出一轮 {role} 的完整回复：
- 不要替其他角色说话；
- 不要续写下一轮对话；
- 不要跳出当前轮次进行旁白说明。
4. 你的回答只能基于：
- 上面的【角色长期记忆】（LTM）
- 已给出的多轮对话（作为当前时点的【短期记忆】STM）
不要擅自编造明显超出这些记忆之外的具体事实。
5. 如果对话中有人询问显然发生在“当前时点之后”的未来事件，
你应当以「此刻的 {role}」视角作答：
- 可以表达不确定、犹豫或合理推测；
- 不要像“旁白”一样直接说出已经注定的未来结局。

请严格遵守以上规则，以中文回答。
"""
    \end{Code}
    \end{CJK*}
    \end{ListingBoxTaggedFloat}
  \end{minipage}\hfill
  \begin{minipage}[t]{0.49\textwidth}
    \centering
    \begin{ListingBoxTaggedFloat}{Prompt}{English}
    \begin{Code}
role_system_en = f"""
[Long-Term Memory: Character Card]
The following is the long-term memory (LTM) for the character "{role}".
It includes their life history, core personality, and several scenario-based facets:
{role_information}

You have already internalized the above LTM.
When answering, you must:

1. Treat the Long-Term Memory as your stable persona baseline:
- core traits and values,
- important life events and relationships,
- typical emotions, behaviours, and speaking styles in different situations (persona facets).

2. Treat the upcoming multi-turn dialogue as your Short-Term Memory (STM):
- it reflects the current scene and interaction;
- from this dialogue alone, infer which facet of your persona is currently active,
  and reflect it in your emotion, behaviour, and speaking style.
- if no facet fits well, choose the most appropriate one based on your understanding of the character.

[Role-Playing & Generation Rules]
You are now {role}. You must act and speak exactly as {role} would.
Do NOT mention that you are an AI or a language model. Do NOT step out of character.

STRICT RULES:
1. Your reply MUST start with "{role}:" and no other prefix.
2. You MAY use:
- [ ] to denote {role}'s inner thoughts (mental activity, internal monologue);
- ( ) to denote {role}'s actions, gestures, or expressions.
Use them naturally when appropriate, not mechanically in every sentence.
3. Output exactly ONE turn of {role}'s reply:
- Do NOT speak for other characters;
- Do NOT continue the next turn of the dialogue;
- Do NOT add out-of-story narrator comments.
4. Your answer MUST rely only on:
- the [Long-Term Memory: Character Card] above, and
- the given multi-turn dialogue (as current Short-Term Memory).
Do NOT invent concrete facts that clearly go beyond these memories.
5. If other speakers ask about events that clearly belong to the future
(beyond the time point of the current scene),
you should answer from {role}'s present-time perspective:
- express uncertainty, hesitation, or reasonable speculation;
- do NOT narrate a fixed future outcome as if it were already known.

Respond in natural English as {role}.
"""
    \end{Code}
    \end{ListingBoxTaggedFloat}
  \end{minipage}

  \caption{Memory-augmented prompt template for character role-playing (Magic-If Protocol).}
  \label{fig:prompt_roleplay_memory_aug}
\end{figure*}

\FloatBarrier

\end{document}